\documentclass{bmvc2k}
\usepackage{amssymb}
\usepackage{cleveref}
\usepackage{booktabs}
\usepackage{multirow}
\usepackage{wrapfig}
\usepackage{enumitem}

\usepackage{siunitx}
\usepackage{placeins}
\title{Approximating Continuous Convolutions for Deep Network Compression}
\addauthor{Theo W.\ Costain}{costain@robots.ox.ac.uk}{1}
\addauthor{Victor A.\ Prisacariu}{victor@robots.ox.ac.uk}{1}

\addinstitution{Active Vision Laboratory\\ University of Oxford\\ \textcolor{bmv@sectioncolor}{\bmvaUrl{https://code.active.vision}}}

\runninghead{T.\ W.\ Costain \& V.\ A.\ Prisacariu}{ApproxConv}

\def\eg{\emph{e.g}\bmvaOneDot}
\def\ie{\emph{i.e}\bmvaOneDot}

\def\vs{\emph{v.s}\bmvaOneDot}

\begin{document}

\maketitle

\begin{abstract}
    We present ApproxConv, a novel method for compressing the layers of a convolutional neural network.
    Reframing conventional discrete convolution as continuous convolution of parametrised functions over space, we use functional approximations to capture the essential structures of CNN filters with fewer parameters than conventional operations.
    Our method is able to reduce the size of trained CNN layers requiring only a small amount of fine-tuning.
    We show that our method is able to compress existing deep network models by half whilst losing only 1.86\% accuracy.
    Further, we demonstrate that our method is compatible with other compression methods like quantisation allowing for further reductions in model size.
\end{abstract}

\section{Introduction}
% \begin{figure}[t]
%     \centering
%     \includegraphics[width=0.3\linewidth]{example-image-1x1}
%     \includegraphics[width=0.3\linewidth]{example-image-1x1}
%     \includegraphics[width=0.3\linewidth]{example-image-1x1}
%     % \includegraphics[width=0.1\linewidth]{example-image-1x1}
%     % \includegraphics[width=0.1\linewidth]{example-image-1x1}
%     % \includegraphics[width=0.1\linewidth]{example-image-1x1}
%     % \includegraphics[width=0.1\linewidth]{example-image-1x1}
%     % \includegraphics[width=0.1\linewidth]{example-image-1x1}
%     \caption{Teaser Text}
% \end{figure}

\begin{wrapfigure}{r}{0.45\linewidth}
    \centering
    \vspace{-1.1em}
    \includegraphics[width=0.9\linewidth]{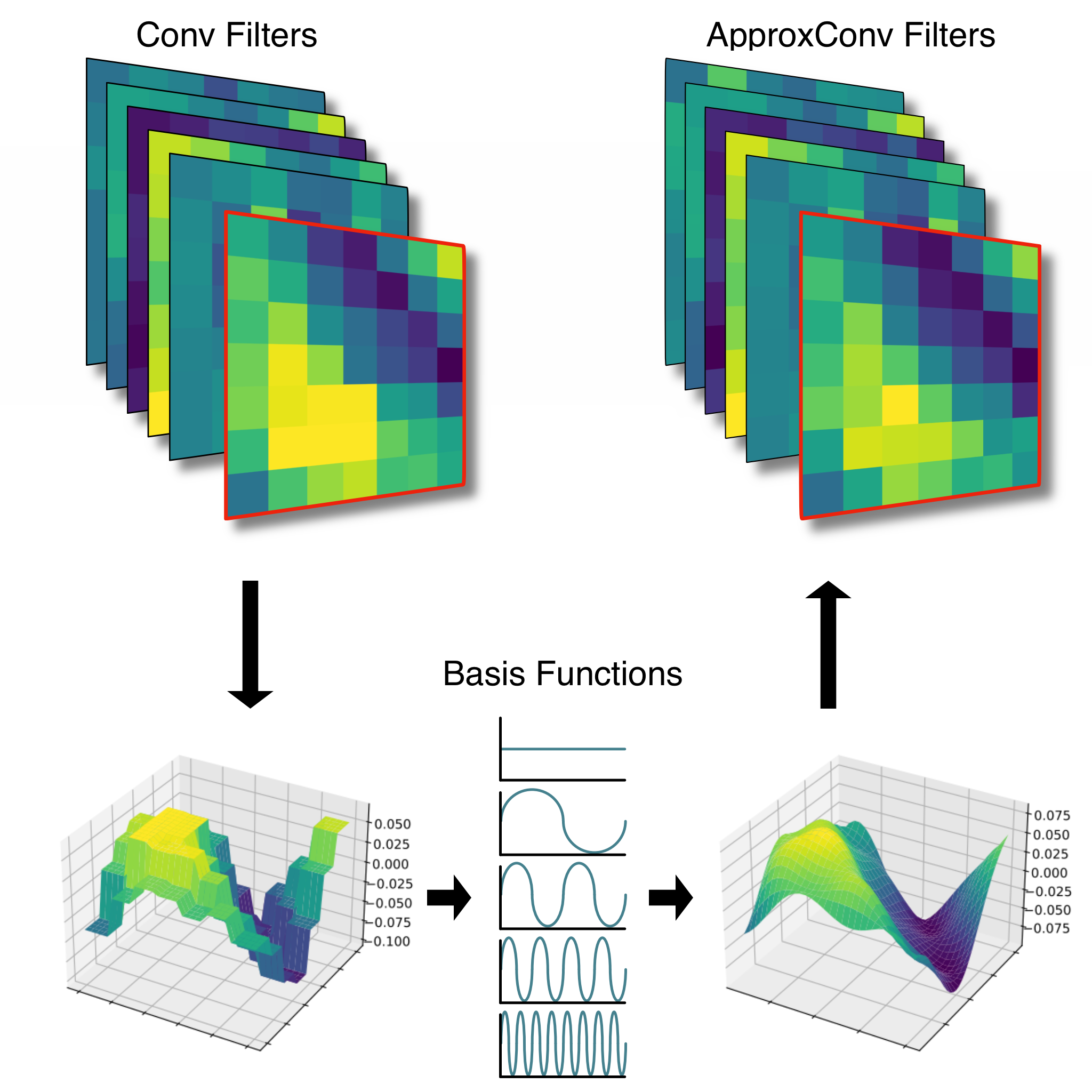}
    \caption{Our method compresses network filter weights, by approximating them using cosine and Chebyshev series.
    }
    \vspace{-1.1em}
\end{wrapfigure}

Deep vision models have demonstrated outstanding performance across a range of problems from semantic segmentation of microscopy images\cite{unet}, to object detection for nature conservation efforts\cite{conservation}.
In many cases, there is strong desire to apply these methods in compute constrained environments, \ie mobile devices, and as a result the desire for space and compute efficient CNN models is as strong as ever.

Despite this, recent works\cite{vit,dino} have seen explosive growth in parameter count for highly performant models, and there is no reason to expect this trend to abate soon\cite{gpt3}.

% On the other hand the trend in current literature for deep vision models is one of steady growth in both the size of models and, therefore, the number of parameters required.
% One of the surprising results of GPT-3\cite{gpt3}, suggests that this trend in parameter count growth is unlikely to abate soon.
% Worse still, recent developments in transformer based vision models have demonstrated outstanding performance on variety of tasks including classification\cite{vit} and segmentation\cite{dino} amongst others, but these models often require significantly more parameters and training data than comparable CNN models\cite{convnext}.
% Fortunately, recent work has shown it is far from clear that the capability of CNNs has been exhausted\cite{convnext}.

Approaches to reduce the size and compute requirements of deep models have been the focus of significant attention.
A number of works have proposed models designed explicitly for efficiency\cite{shufflenet,squeezenet,mobilenets}, and other efforts to improve the efficiency of deep models have included methods such as low-rank factorisation\cite{jaderberg2014speeding}, knowledge distillation\cite{hinton2015distil}, quantisation\cite{zhou2016dorefa}, and network pruning\cite{lecun1989optimal,frankle2018lottery}.

In this work, we propose to use functional approximations to reduce the number of parameters required to fully describe a CNN layer's filters.
We re-cast conventional discrete convolution on grids as discrete convolution on parametrised continuous functions of space.
In this way, we are able to replace the kernel weight tensor with an approximating function that is able to exploit the structures commonly found in CNN filters.
Specifically, we make use of cosine and Chebyshev series that are able to preserve the low frequency information that allows the networks to function, whilst minimising the effects of high frequency information that are often exploited by adversarial attacks\cite{zhou2021high}.
Our proposed method is particularly effective for larger kernel sizes, which despite falling out of favour in earlier deep CNN work\cite{resnet,vgg}, have recently seen a resurgence\cite{convnext}.

In our experiments, we show that our method is able to significantly reduce the number of parameters in a variety of models, and unlike comparable methods\cite{FracSRF,Fractional} is able to achieve this with only a limited amount of fine-tuning.
Further, we demonstrate that our approach is complementary to other network size reduction methods such as quantisation.
Complimenting our method with other approaches, we are able to increase the overall reduction in total model size with minimal further impact on accuracy.

In summary, the contributions of this work are as follows:
\begin{itemize}[noitemsep]
    \item A novel method to approximate kernel weights using functional approximations based on cosine and Chebyshev series.
    \item A framework that allows this method to be used without completely retraining models from scratch.
          % \item Demonstrations that our method is able to improve parameter efficiency of deep models across common benchmark datasets.
\end{itemize}

In the rest of the paper, we discuss works related and relevant to our method(\Cref{sec:rel}), our methodology (\Cref{sec:method}), our experiments and results (\Cref{sec:results}), before a final a discussion and the conclusion of the work (\Cref{sec:conclusion}).
\section{Related Work}
\label{sec:rel}
A number of works\cite{shufflenet,squeezenet,huang2017densely} have attempted to reduce the size of models by designing networks with efficiency in mind, such as MobileNets\cite{mobilenets} which introduced depthwise convolutions.

Quantisation methods\cite{nascimento2019dsconv,zhou2016dorefa,han2015deep_compression,song2018computation,drumond2018training} have seen great success in reducing the memory footprints of deep models.
Quantising 32-bit floating point numbers down to 8, 4 or even as low as 1 bit, these methods are able to drastically reduce the memory footprint of deep models.
Other works\cite{gennari2020finding,wang2019haq,elthakeb2020releq} demonstrated that non-uniform quantisation schemes, where different layers of the network can be quantised to different numbers of bits, can further reduce the size of the model without compromising accuracy.
These non-uniform schemes influence some of our experiments, where we employ a similar approach.

Knowledge distillation methods\cite{hinton2015distil,mirzadeh2020improved,zhang2019your,romero2015fitnets}, typically use larger (teacher) models to guide the training of smaller (student) ones, often allowing the student models to achieve performance parity despite their smaller size.

Network pruning methods\cite{lecun1989optimal,han2015learning,anwar2016coarse,luo2018thinet} attempt to discover redundant parameters in deep networks. Having identified these parameters, they can then be removed either individually, yielding sparse networks, or as a group, sometimes removing whole filters or channels.
Seminal works, like \cite{frankle2018lottery,malach2020proving}, demonstrate that it can be possible to prune networks before training, realising the size reduction benefits for training as well as inference.

Low rank factorisation methods\cite{sainath2013low,jaderberg2014speeding,denton2014exploiting,zhang2015accelerating} reduce both the total number of parameters in CNN layers, as well as the number of FLOPs required to compute the outputs of the layer.
Using a variety of methods, these approaches decompose the weight tensors of layers into a series of smaller tensors. These smaller tensors can then be convolved with the inputs sequentially (in a fashion similar to separable filters from classical methods), yielding an approximation of the original output.

% Finally, there are two methods that are most similar to our proposed method\cite{FracSRF, Fractional}.
Both methods learn parametrised functions over space, however rather than generalised approximations, they make use of gaussian and gaussian derivative functions.
\citet{Fractional} learns only simple classical image filters (\eg Sobel filter) and \citet{FracSRF} attempt to replicate \citet{SRF} using fewer parameters.
Using fractional calculus, they are able to efficiently parametrise a family of gaussian and gaussian-derivative functions using only 3\cite{FracSRF} and 6\cite{Fractional} parameters per filter.
While this approach allows for significant parameter savings, in comparison to our proposed method, the shape of kernels that can be learnt is substantially limited.
As a result, whereas their methods require training from scratch, our approach can be applied to pre-trained models, .

In 3D settings, dense regular grid structures are too memory intensive and so receive little attention.
Instead, much of the effort focuses on irregular data structures like point-clouds or meshes.
In these settings it becomes desirable to have a kernel function that can vary continuously over 3D space.
Similar to our method, SplineCNN\cite{splinecnn} uses b-Splines to generate a continuous function over space.
% Their approach uses a different "spline" for each kernel location, preventing their method from being used to compress a network in the same manner as ours.
FlexConv\cite{flexconv} learns a kernel weight function based on the distance between sampled point locations and a hyperplane.
% Whilst effective, like other works \cite{Fractional,FracSRF}, these kernel functions are not general approximators, and as such are not reliably able to replicate arbitrary kernels.
\citet{edgeconv} use shared MLPs with edge information as input to define the kernel weight functions.
% For regular grids, like \cite{splinecnn}, this approach requires the same number of parameters as conventional convolution.

% Moving to the 2D setting this paper is concerned with, there are a vast array of approaches to reducing the number of parameters of deep networks, and in the case of some methods the number of operations.

\section{Method}
\label{sec:method}
We begin by covering the general form of continuous convolution and how this formulation can allow for network compression.
Next we outline how we come to our choice of approximation functions, before finally discussing how we apply our method to pre-trained networks.

\subsection{Continuous Convolution}
In general, discrete $n$-dimensional convolution can be expressed in the form
\begin{equation}
    x^\prime = \sum_{x_i\in\mathcal{N}(x)}x_i \times w_i
\end{equation}
where $\mathcal{N}(x)$ represents the set of grid locations neighbouring $x$, and $w$, typically referred to as the kernel, contains an entry for every neighbouring location.
If we replace the discrete grid in the above equation with set of locations sampled from a continuous function over space, we can re-write the above equation as
\begin{equation}
    \label{eq:contconv}
    x^\prime(\mathbf{p}) = \sum_{p_i\in\mathcal{N}(x)} x(\mathbf{p}_i) \times w(\mathbf{p}_i)
\end{equation}
where $\mathbf{p}\in\mathbb{R}^n$, $x:\mathbb{R}^n\mapsto\mathbb{R}^d$, and $w:\mathbb{R}^n\mapsto \mathbb{R}^d$. To recover simple 2D convolution with a $3\times3$ kernel, we can simply let $p$ be the set of 2-vectors $[(-1,-1),(-1,0),(-1,1),\cdots]$.

This formulation has two key advantages:
i) The same convolutional operation can be used on both regular $n$D grids, and on arbitrary $n$D structures (\eg point-clouds or meshes) over which we can define a neighbourhood function $\mathcal{N}$;
ii) Replacing the tensor $w$ with a carefully chosen function $w(\cdot)$, we can reduce the memory requirements of 2D convolutional layers by exploiting the structures present in the tensors.

\subsection{Network Compression}
Our work is concerned with the second of the two advantages above: that a judicious choice of $w(\cdot)$ can permit parameter savings.
As has been shown by other works\cite{lecun1989optimal,malach2020proving} deep networks often contain significant amounts of redundant information, and often filters that contain low-frequency information.
Accordingly, it is often possible to substantially reduce the size of a model by removing redundant parameters\cite{han2015learning,he2017channel} or preserving only \emph{relatively} low-frequency structures\cite{FracSRF}.

The choice of function $w(\cdot)$ is primarily constrained by the number of parameters it requires.
As discussed previously, methods like FracSRF\cite{FracSRF} and Fractional Filters\cite{Fractional} choose efficiently parametrised gaussian and gaussian-derivatives as $w(\cdot)$.
But these functions are not able to approximate arbitrary functions.

Separate to these approaches, the field of functional approximation has been of interest for over a century\cite{TrefethenLloydN2019Ataa}, motivated by applications ranging from solving PDEs to numerical integration.
As a result a significant body of work is concerned with using various orthogonal basis functions for approximations.
From this corpus, we use two common families of functions to allow our method to learn arbitrary kernel functions: cosine and Chebyshev series.

\subsection{Cosine and Chebyshev Series}
Cosine and Chebyshev series are able to approximate any periodic function that satisfies specific boundary conditions and the Dirichlet conditions.
These conditions require that the function: i) is absolutely integrable; ii) is of bounded variation; iii) has a finite number of non-infinite discontinuities.
As any discrete weight kernel can be fully described by a piecewise linear function, it is trivial to demonstrate\footnote{We leave a complete proof as an exercise for the reader} that this piecewise function satisfies the above conditions.
Further, the requirement that the function be periodic (in the case of the cosine series) can be trivially satisfied by repeating the function at the boundaries, although careful choice of boundary conditions and approximation interval is necessary.

% Although the Fourier series is more commonly used in numerical approximation, 
% This choice is motivated by a desire for simplicity of implementation and extension.
Although the Fourier series is generally more well known than the cosine series, extensions of the Fourier series to higher dimensions introduces a number of "cross terms" that grow in complexity as the number of dimensions increases.
An alternative is to use the sine or cosine series, which are equivalent to the fourier series under certain conditions.
Specifically, for an \emph{even} function, the coefficients of the sine terms go to zero leaving only cosine terms, and vice versa for \emph{odd} functions.
By shifting the center of our approximation interval, it is possible to coerce the function to behave as an even or odd function.
The advantages of the sine and cosine forms is that their extensions into higher dimensions are trivial and, for $\mathbf{x}\in\mathbb{R}^n$ with $N$ harmonics, take the following (for cosine) form
\begin{equation}
    \label{eq:kfcos}
    \hat{w}(\mathbf{x}) = \sum_{i_0=0}^{N}\cdots\sum_{i_n=0}^{N} a_{i_0\cdots i_n}\cos(i_0x)\cdots\cos(i_ny)
\end{equation}

The choice of cosine over sine is motivated by concerns over boundary conditions.
Specifically, for cosine series the boundary conditions require/enforce symmetric periodic repetition which minimises the potential for discontinuity at the boundary.

In our experiments, we also make use of the Chebyshev series as it is closely related to the cosine series through the definition that $T_n(\cos(x))=\cos(nx)$, which conveniently yields a very similar expression to \cref{eq:kfcos}
\begin{equation}
    \hat{w}(\mathbf{x}) = \sum_{i_0=0}^{N}\cdots\sum_{i_n=0}^{N} a_{i_0\cdots i_n}T_{i_0}(x_0)\cdots T_{i_n}(x_M)
\end{equation}
where $T_n$ is the $n$\textsuperscript{th} Chebyshev polynomial of the first kind, and $\mathbf{x}\in\mathbb{R}^{M}$.
The Chebyshev series can have faster "convergence" (\ie better approximation with fewer terms) than the trigonometric series in \cref{eq:kfcos}.

To achieve compression for a $C_\text{out},C_\text{in},K\times K$ kernel, we choose $\hat{w}$ to be a 2D cosine or Chebyshev series with $N$ ``harmonics''or ``orders'', where $N\leq K$.
Replacing a $3\times 3$ kernel, with a series 2 harmonics leads to an over 50\% reduction in parameters.
Our experiment show however, that 2 harmonics \emph{can} be insufficient to properly approximate a $3\times 3$ kernel.
However, for $7\times 7$ kernels, which have seen a resurgence in recent work\cite{convnext}, the increased number of harmonics permitted whilst still maintaining a reduction in parameters, can allow for much better compression depending on the complexity of the filters in question.
In our experiments, we refer to replacing the kernel weights with a cosine series as \textbf{CosConv}, and \textbf{ChebConv} when using a Chebyshev series.

Whilst it is not necessary to use the same number of harmonics in the $x$ and $y$ directions, doing so significantly increases the hyperparameter space to be evaluated for our method.
As we are not aware of any popular or common methods that make use of anisotropic kernels, we do not investigate differing spatial harmonics in this work.
However, with an appropriate search method, anisotropic harmonics might permit further increases in memory savings.

Also not investigated is another possibility for reducing parameter count using our method, through variable harmonics for each filter.
Although, as above, this is essentially a problem of hyperparameter/architecture search and is outside the scope of this work.

\subsection{Finding Weights}\label{sec:initapprox}
Depending on the number of harmonics used in the approximating function, there may exist a closed form solution to find the approximation.
In the case of the cosine series, this takes the form of the discrete cosine transform.
This closed form solution, however, is only possible when choosing the same number of parameters for the approximation as for the original weight kernel.
Accordingly, when approximating the kernel weights for compression, we use a simple iterative gradient descent optimisation to minimise the mean squared error loss

\begin{equation}
    \mathcal{L} = \frac{1}{K^2}\sum_{i,j=0}^{K} (w_{ij} - \hat{w}(\mathbf{p}_{ij}))^2
\end{equation}

In conventional numerical approximation, the choice of ``sample points'' $\mathbf{p}_{ij}$ is extremely important\cite{TrefethenLloydN2019Ataa}.
Normally, performing approximation using Chebyshev polynomials requires careful choice of sample points\footnote{For Chebyshev series, the Chebyshev-Gauss-Lobatto points. For cosine series, equispaced points.} to avoid Runge's phenomenon, where approximation error at the ends of the interval increases with the number of harmonics\cite{TrefethenLloydN2019Ataa}.
However, in our case, as we only care about the value of the approximation at the individual sample points, we can safely ignore this phenomenon.
% If we were to change our sample points after the initial approximation, it would then become necessary to choose how we sample the kernel.
% TODO: CosConv sample ranges and EdgeConstant

In our experiments, we found that the Chebyshev series was significantly more sensitive to the initialisation of the weights before the initial approximation.
For ChebConv, we found the best results were achieved by setting the weights of the ``DC'' harmonics to the mean of the kernel weights for each filter, and the weights of all higher harmonics are set to 0.
Conversely, we found CosConv is less sensitive to the initial choice of weights, and that sampling weights from $\mathcal{N}(0,\frac{1}{C_\text{in}K^2})$ was sufficient.
In both cases, we suspect that there is likely a better motivated initialisation scheme, but we leave this for future work.
However for CosConv, we do not expect a better initialisation will improve performance.

% \subsection{Fine Tuning}
% Following this initial approximation, we then fine-tune the network for a small number of epochs.
% This allows the network to correct for any small approximation errors that the initial learning of weights might have caused.
% Because our approximation is continuous and differentiable, we are able to back-propagate through our approximations without any complex machinery.

% Unlike other methods discussed previously, our approach does not require the training of the entire network. Whilst this does come at the cost of increased memory compared to other compression methods (SRF, Adaptive Kernels), the reduced training requirements may allow our approach to be more readily deployed in certain contexts.
% \section{Experiments}
% \label{sec:expr}
% \input{sections/4_experiments.tex}
\section{Experiments \& Results}
\label{sec:results}
Following this initial approximation, we fine-tune the network for a small number of epochs.
This allows the network to correct for any small approximation errors that the initial approximation of weights might have caused.
In all our experiments, we fine-tune the networks for 5 epochs.
Specific details of training schedules, and the datasets used, are presented in the supplementary material.

% We use two common datasets in our experiments: CIFAR-10\cite{cifar} and ImageNet\cite{imagenet}.
% CIFAR-10 consists 50,000 training images and 10,000 validation images, distributed across 10 categories.
% During retraining/fine-tuning, images are randomly cropped to $32\times32$, randomly horizontally flipped and normalised. During validation/testing images are normalised identically to training.
% The much larger ImageNet dataset consists of 1.28M training images and 50,000 validation images across 1000 categories.
% Following other works, we report our results for both datasets using the validation split.

\subsection{CIFAR}
\begin{table}[htb]
    \centering
    \resizebox{0.99\linewidth}{!}{

        \begin{tabular}{cccccccccc}
            \toprule
                                                                &                             & \multicolumn{3}{c}{ResNet-20} & \multicolumn{3}{c}{ResNet-32}                                                                          \\
            \cmidrule(lr){3-5}\cmidrule(lr){6-8}
                                                                & Layer                       & Pre Top 1                     & Post Top 1                    & $\Delta$ & Pre Top 1 & Post Top 1 & $\Delta$ & \# Params & $\Delta$ \% \\
            \midrule
                                                                & Conv2D                      & -                             & 91.73                         & -        & -         & 92.78      & -        & 0.46M     & 0.0         \\
            [0.5em]
                                                                & Fractional\cite{Fractional} & 91.25                         & 91.29                         & 0.04     & -         & -          & -        & -         & -           \\
            % & MBS            & 91.86                         & 91.19                         & -0.67    & 0.19M     & 70.37       & 92.24     & 91.82      & -0.42    & 0.33M     & 46.81       \\
                                                                & FracSRF\cite{FracSRF}       & -                             & -                             & -        & 92.28     & 91.60      & -1.18    & 0.16M     & 34.00       \\
                                                                & FracSRF$\dagger$            & 10.00                         & 38.08                         & -53.65   & 9.82      & 37.91      & -54.87   & 0.15M     & 33.43       \\
            [1em]
            %   & 3,3,3,3                     & 91.73                         & 91.68                         & -0.05    & 92.63     & 92.56      & -0.22    & 0.46M     & 100.00      \\
            \multirow{5}{*}{\rotatebox[origin=c]{90}{CosConv}}  & 3,3,3,2                     & 87.22                         & 90.75                         & -0.98    & 87.74     & 91.51      & -1.27    & 0.27M     & 57.74       \\
                                                                & 3,3,2,2                     & 51.43                         & 89.55                         & -2.18    & 75.62     & 90.99      & -1.79    & 0.22M     & 47.35       \\
                                                                & 3,2,2,2                     & 21.16                         & 86.99                         & -4.74    & 16.52     & 88.4       & -4.38    & 0.21M     & 44.57       \\
                                                                & 2,2,2,3                     & 39.63                         & 86.87                         & -4.86    & 20.92     & 88.64      & -4.14    & 0.40M     & 86.65       \\
                                                                & 2,2,2,2                     & 17.55                         & 85.98                         & -5.75    & 15.39     & 87.87      & -4.91    & 0.21M     & 44.48       \\
            [0.5em]
            %  & 3,3,3,3                     & 91.73                         & 91.61                         & -0.12    & 92.63     & 92.55      & -0.23    & 0.46M     & 100.00      \\
            \multirow{5}{*}{\rotatebox[origin=c]{90}{ChebConv}} & 3,3,3,2                     & 87.22                         & 90.76                         & -0.97    & 87.74     & 91.48      & -1.30    & 0.27M     & 57.74       \\
                                                                & 3,3,2,2                     & 51.43                         & 89.83                         & -1.90    & 75.62     & 91.16      & -1.62    & 0.22M     & 47.35       \\
                                                                & 3,2,2,2                     & 21.16                         & 87.79                         & -3.94    & 16.52     & 88.48      & -4.30    & 0.21M     & 44.57       \\
                                                                & 2,2,2,3                     & 23.63                         & 87.95                         & -3.78    & 20.92     & 89.26      & -3.52    & 0.40M     & 86.65       \\
                                                                & 2,2,2,2                     & 17.55                         & 86.66                         & -5.07    & 15.39     & 88.16      & -4.62    & 0.21M     & 44.48       \\
            \bottomrule
        \end{tabular}

    }
    \vspace{0.3em}
    \caption{Results showing the application of our method to the ResNet-20 and ResNet-32 models on CIFAR10.
        The sequence of numbers for the CosConv and  ChebConv rows denote the number of ``harmonics'' for each of the blocks in the network.
        $\dagger$ indicates using our regression and retraining approach.
    }
    \label{tab:cifar}
\end{table}
We first conduct experiments on CIFAR-10\cite{cifar} to validate our method.
We use the smaller ResNet-20 and ResNet-32 networks from \citet{resnet}.
These models consist of 4 ``blocks'' of filters, with different numbers of channels for each block.
The results of our experiments are presented in \cref{tab:cifar}.
The sequence of numbers in the layer column next to the CosConv and ChebConv labels represent the number of harmonics used for the layers in each block.
The columns show the accuracy after the initial approximation (Pre Top 1), and the accuracy after 5 epochs of fine tuning.
The $\Delta$ columns show: i) the difference in Top 1 accuracy between the method/configuration and the baseline 2D convolution; ii) the percentage change in number of parameters.
Our method is able to reduce the size of ResNet-20 by 42\% losing only 1\% accuracy.
For both the 20 and 32 variants, the final block contains the majority of the parameters because of the larger channel dimension.
Accordingly, reducing the number of parameters of this final block alone has more of an effect on the overall number of parameters than all of the other blocks combined (see $\Delta \%$ for 2,2,2,2 \vs 2,2,2,3).
Whilst not designed to approximate previously learned filters, we investigate applying our ``regression and re-training'' approach to both Fractional Filters\cite{Fractional} as well as FracSRF\cite{FracSRF}.
FracSRF was not able to approximate the kernels initially, but was able to recover some of the accuracy during fine-tuning.
We found that Fractional Filters were not able to approximate the kernels to any degree even after fine tuning, and the results were no better than random guessing (10\% Top 1).
% We do not wish to suggest that this comparison is \emph{fair}, as neither of the authors claimed their methods would be able to approximate already learned filters.
\begin{figure}[tb]
    \centering
    \includegraphics[width=0.18\linewidth]{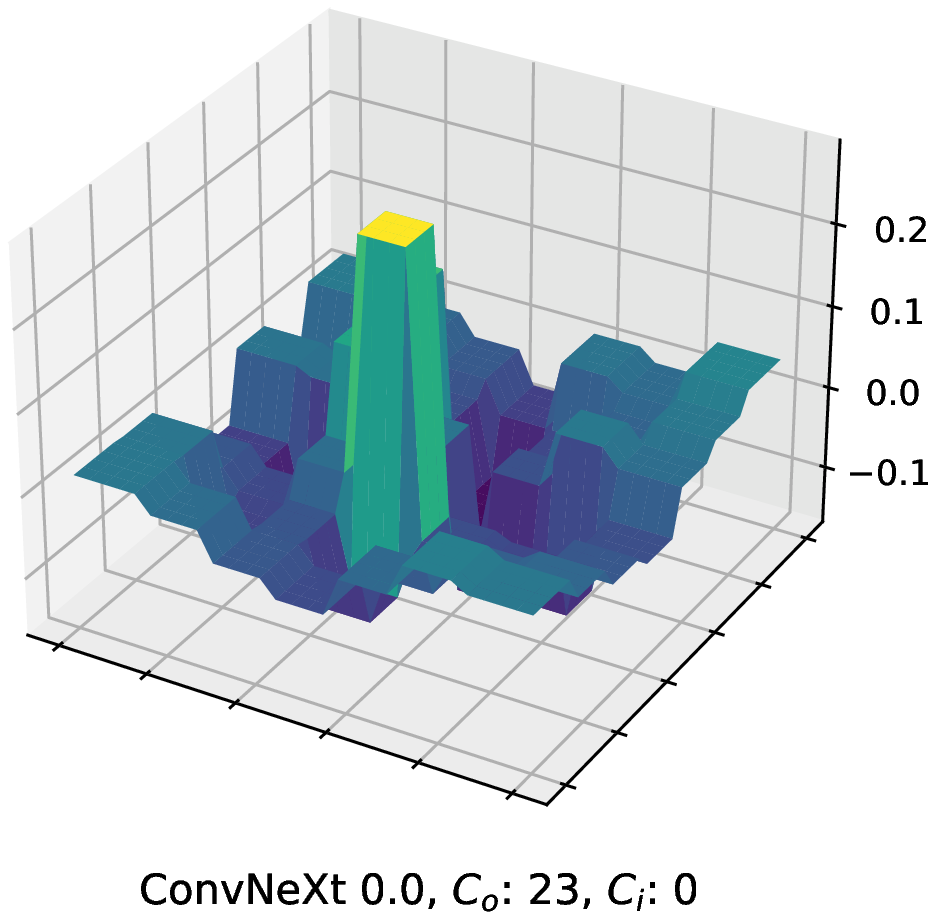}
    \includegraphics[width=0.18\linewidth]{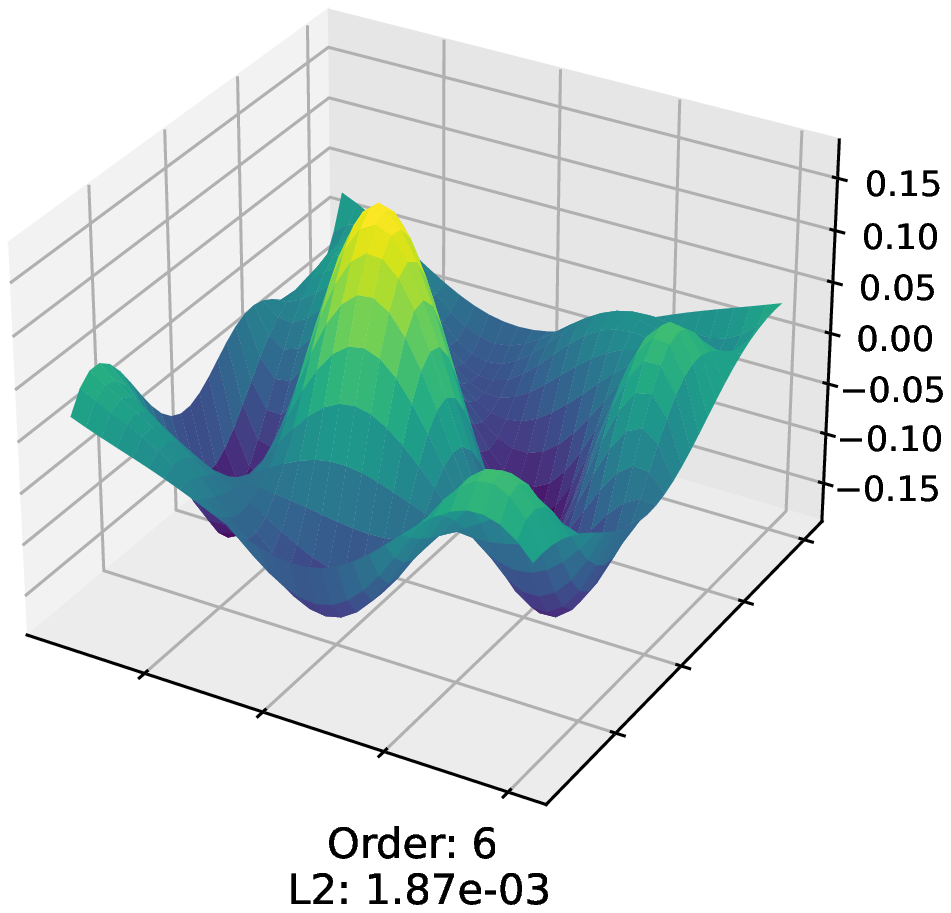}
    \includegraphics[width=0.18\linewidth]{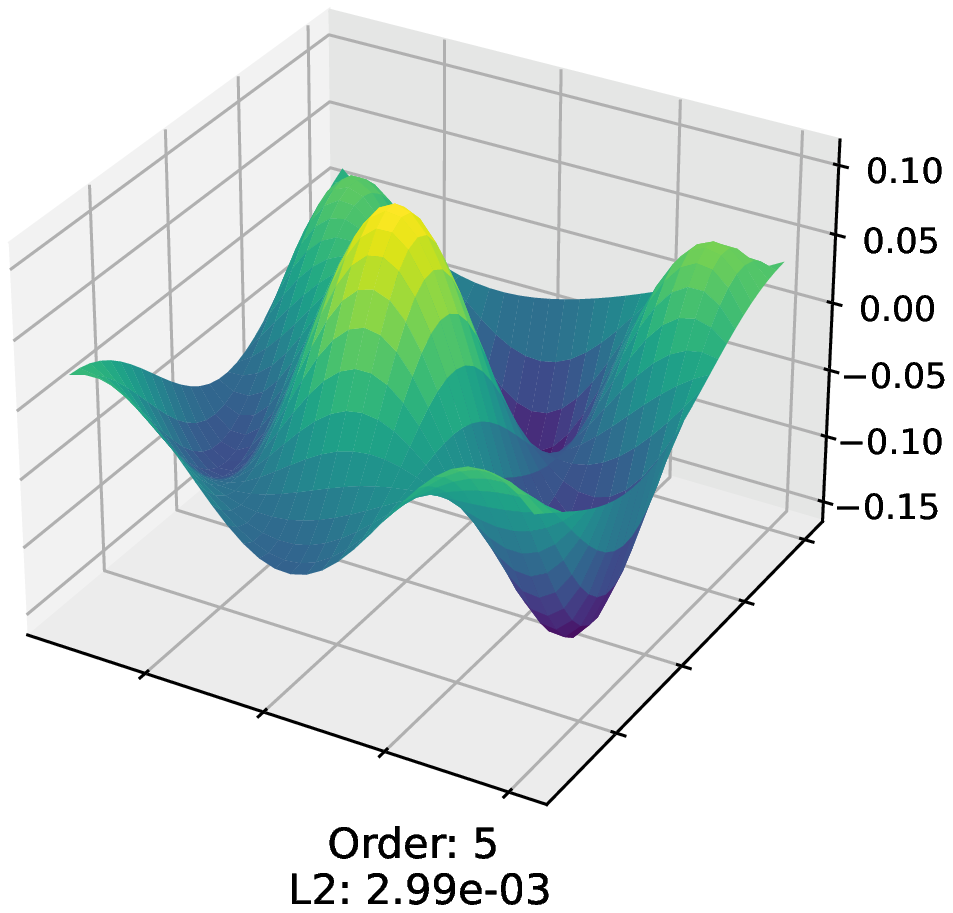}
    \includegraphics[width=0.18\linewidth]{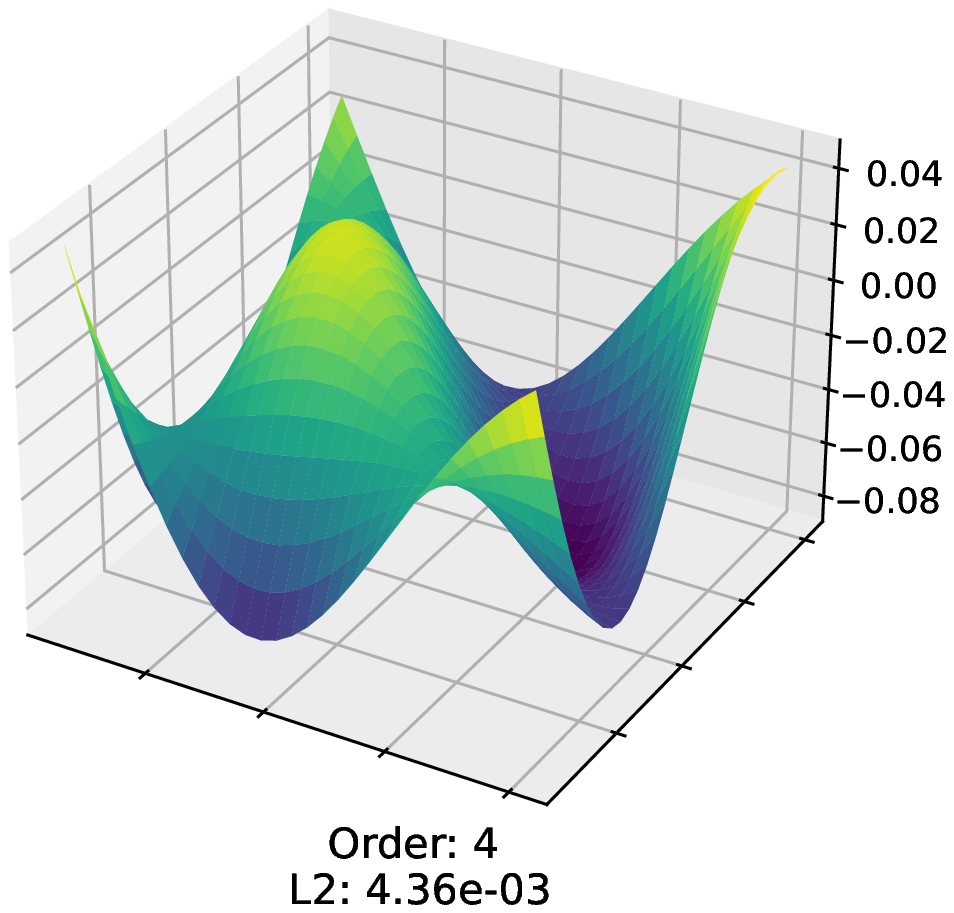}

    \includegraphics[width=0.18\linewidth]{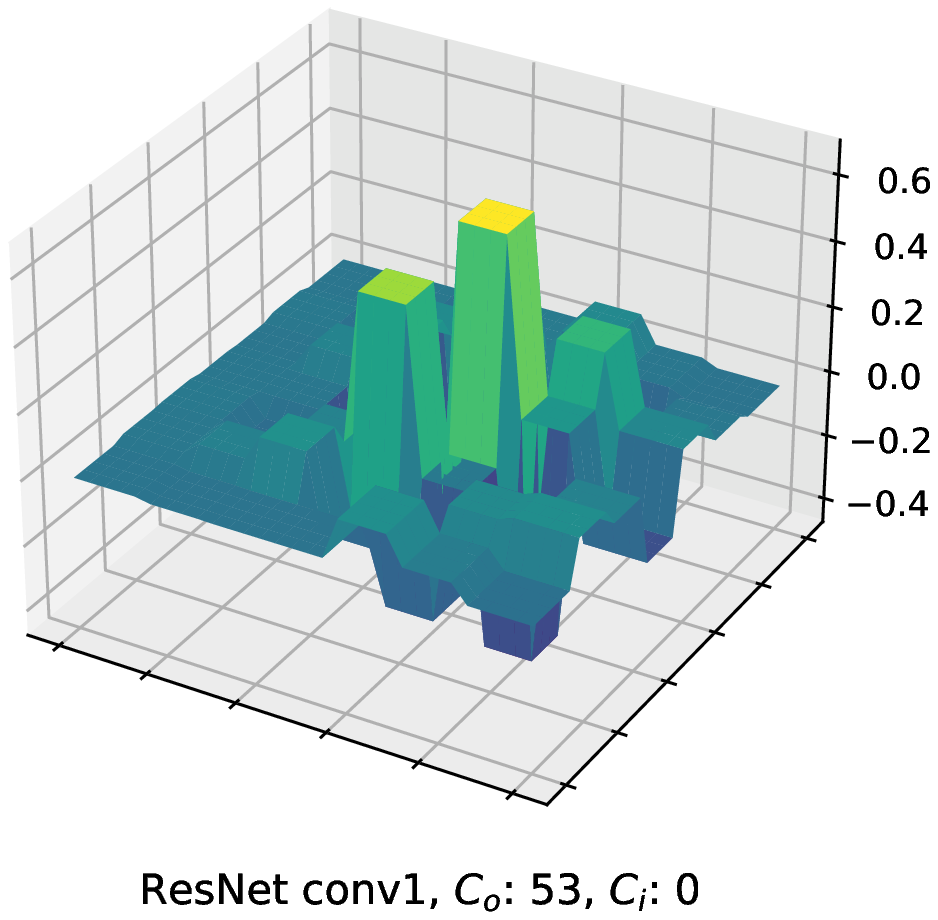}
    \includegraphics[width=0.18\linewidth]{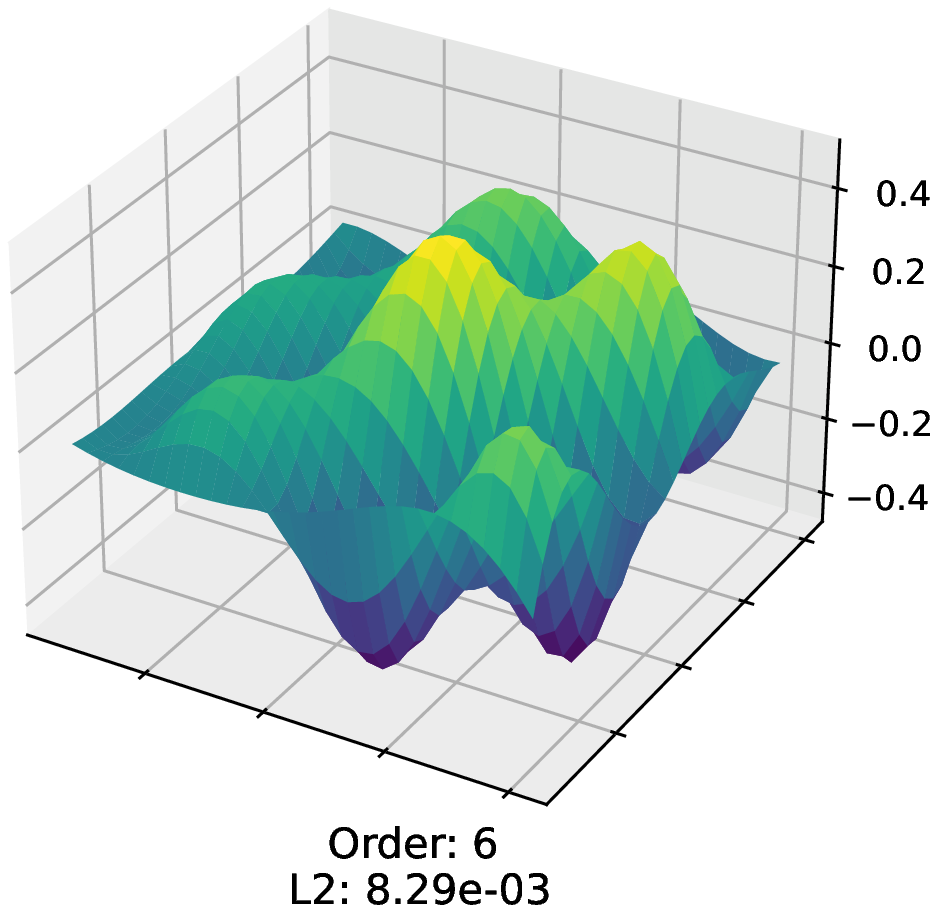}
    \includegraphics[width=0.18\linewidth]{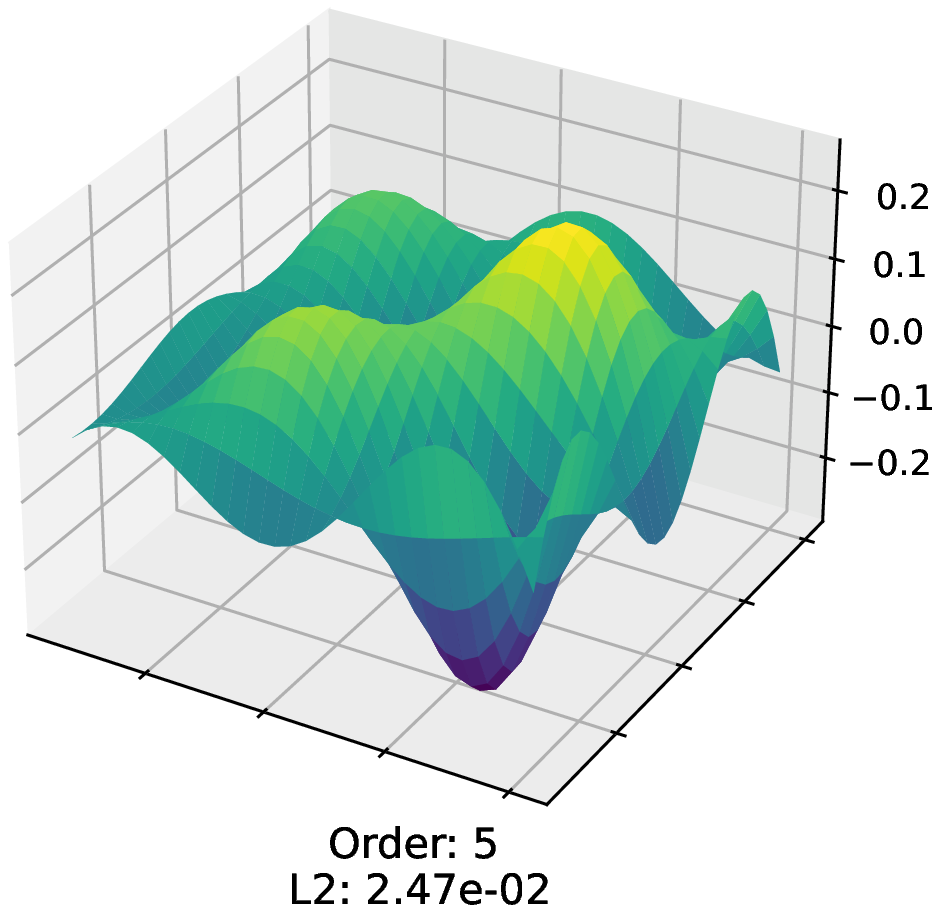}
    \includegraphics[width=0.18\linewidth]{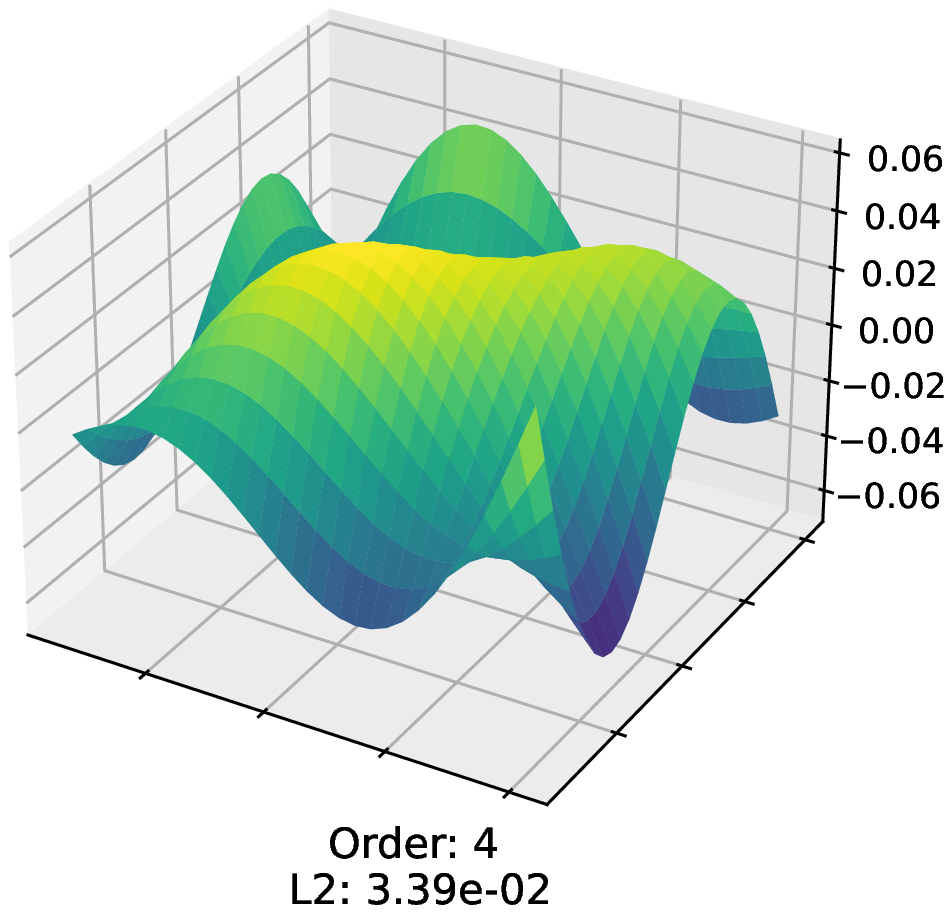}

    \includegraphics[width=0.18\linewidth]{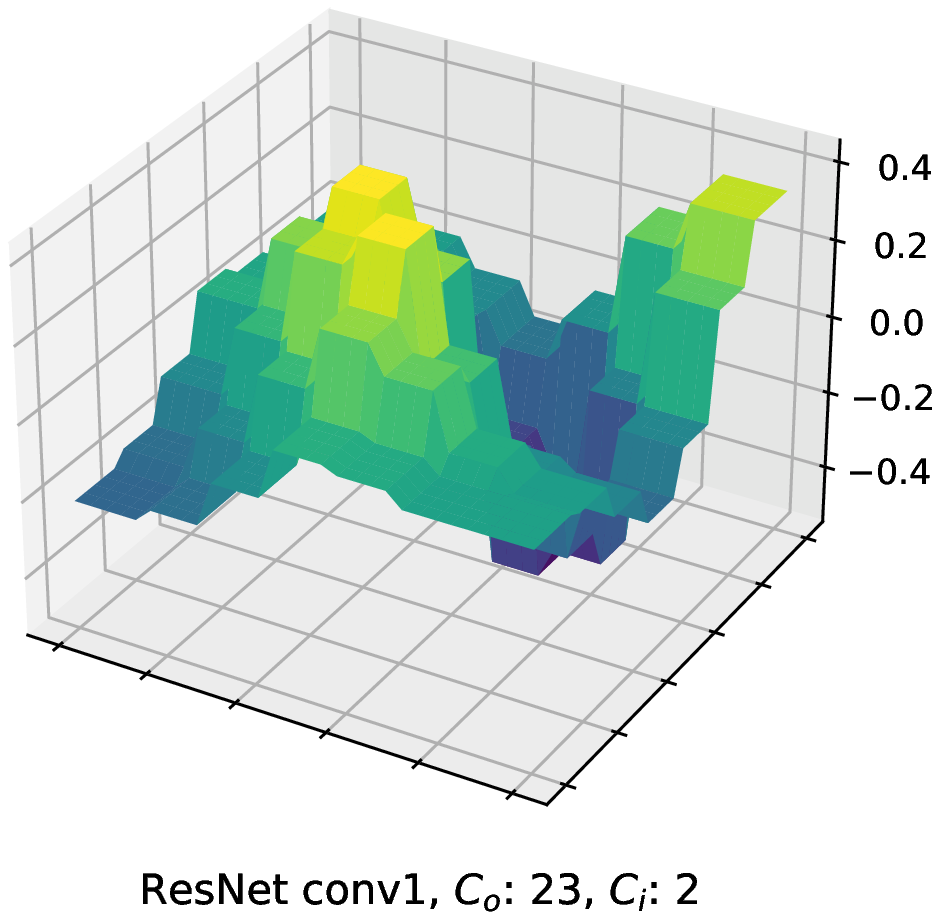}
    \includegraphics[width=0.18\linewidth]{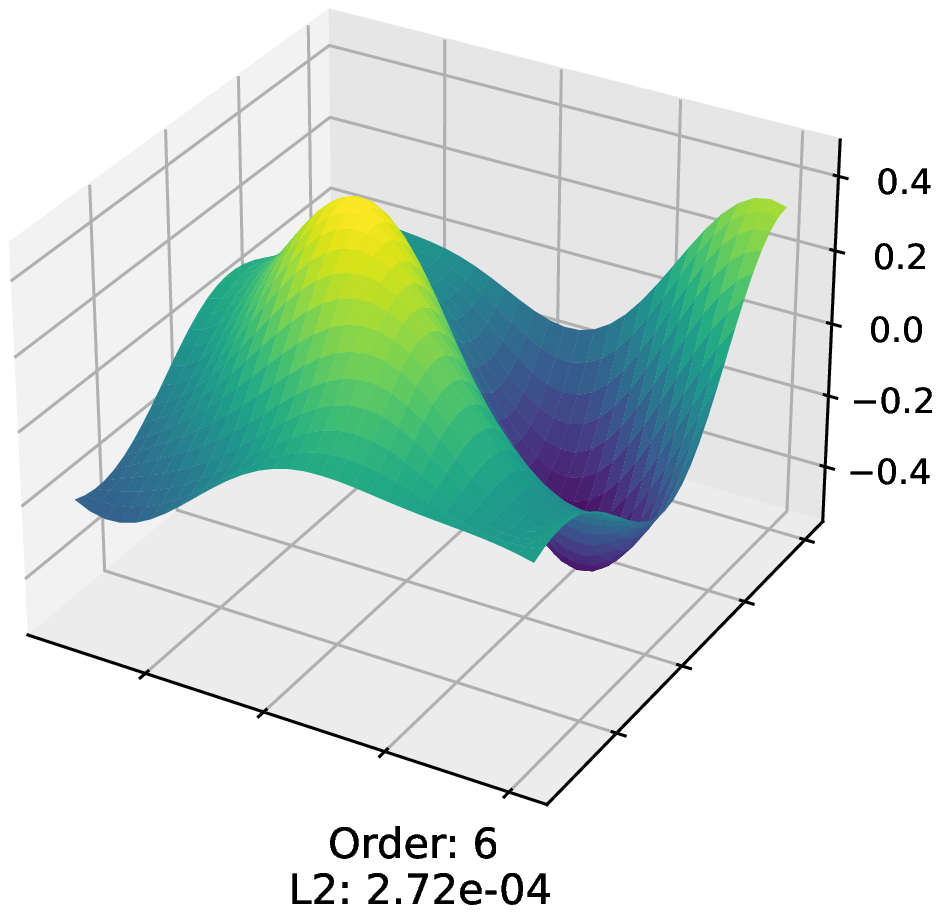}
    \includegraphics[width=0.18\linewidth]{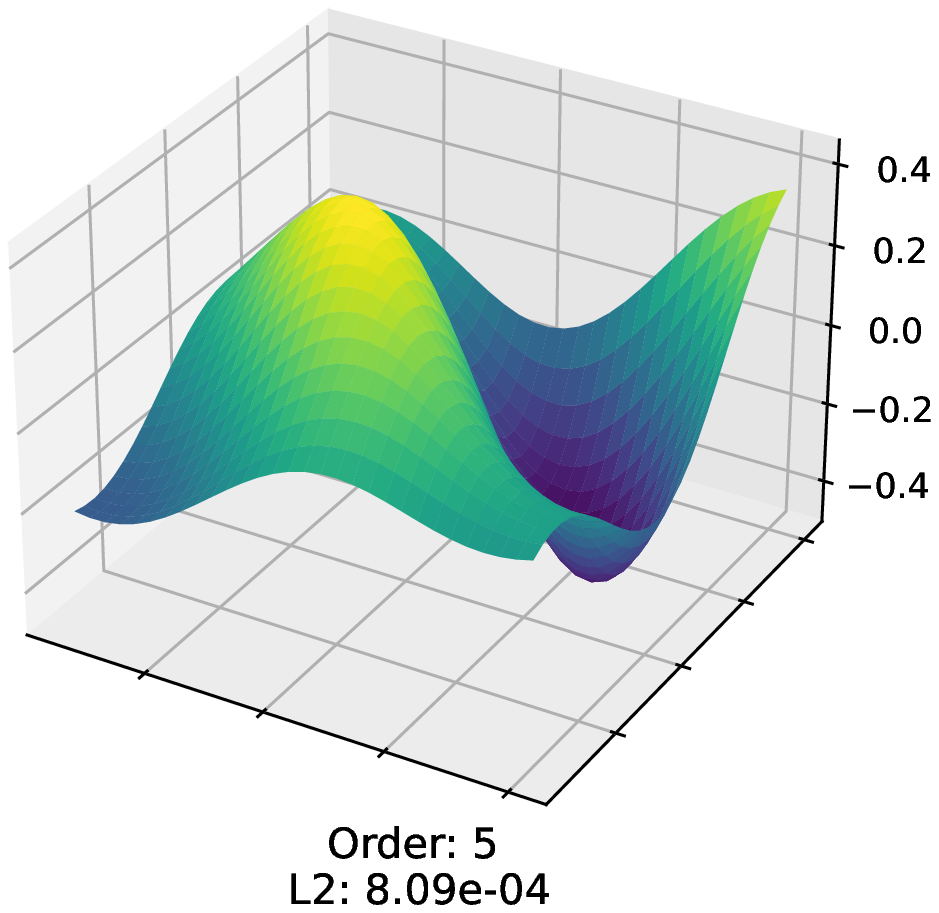}
    \includegraphics[width=0.18\linewidth]{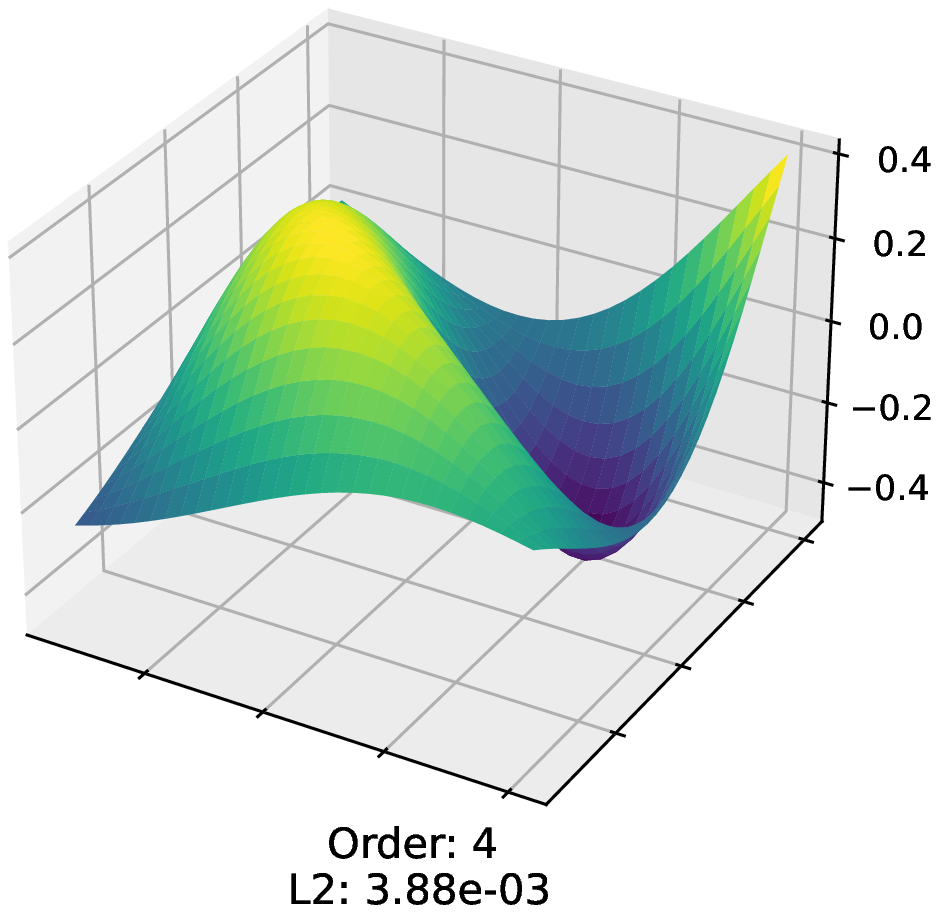}

    \caption{Filters from the ConvNeXt and ResNet networks compressed using our method.
        Left: Original Kernel, Right (in columns): Compressed version of this kernel using CosConv with 6, 5, and 4 orders/harmonics respectively.}
    \label{fig:approx}
\end{figure}

\begin{table}[p]
    \centering
    \resizebox{0.99\linewidth}{!}{
        \begin{tabular}{cccccccccc}
            \toprule
                      & \multicolumn{3}{c}{CosConv} & \multicolumn{3}{c}{ChebConv}                                                                                 \\
            \cmidrule(lr){2-4}\cmidrule(lr){5-7}
            % Config    & Pre Top 1 & Post Top 1 & $\Delta$ \% & \# Params. & $\Delta$ \% & \# Comp. Params. & $\Delta$ \%                                                                             \\
            Config    & Pre Top 1                   & Post Top 1                   & $\Delta$ \% & Pre Top 1 & Post Top 1 & $\Delta$ \% & \# Params. & $\Delta$ \% \\
            \midrule
            Conv2D    & -                           & 69.76                        & 0.0         & -         & 69.76      & 0.0         & 10.99M     & 0.0         \\
            % FracSRF (R\&R) & -         &            &          & 4.35M     & 37.22           & 3.66M          & 33.31                \\
            [1em]
            6,3,3,3,3 & 69.46                       & 70.19                        & 0.43        & 69.32     & 70.11      & 0.35        & 11.68M     & 99.98       \\
            6,3,3,3,2 & 64.33                       & 68.97                        & -0.79       & 64.11     & 68.62      & -1.14       & 7.09M      & 60.70       \\
            6,3,3,2,2 & 56.59                       & 67.90                        & -1.86       & 56.35     & 67.23      & -2.53       & 5.94M      & 50.88       \\
            6,3,2,2,2 & 14.56                       & 66.89                        & -2.87       & 14.68     & 65.63      & -4.13       & 5.66M      & 48.43       \\
            5,3,3,3,3 & 66.71                       & 69.80                        & 0.04        & 65.79     & 69.74      & -0.02       & 10.99M     & 94.10       \\
            5,3,3,3,2 & 60.81                       & 68.53                        & -1.23       & 59.71     & 68.15      & -1.61       & 7.09M      & 60.68       \\
            5,3,3,2,2 & 51.92                       & 67.48                        & -2.28       & 13.58     & 65.44      & -4.32       & 5.94M      & 50.86       \\
            4,3,3,3,2 & 56.21                       & 68.05                        & -1.71       & 54.37     & 67.34      & -2.42       & 7.09M      & 60.67       \\
            4,3,3,2,2 & 45.99                       & 66.84                        & -2.92       & 11.35     & 64.66      & -5.1        & 5.94M      & 50.85       \\

            \bottomrule
        \end{tabular}
    }
    \vspace{0.3em}
    \caption{Results showing the application of our method to ResNet-18 on the ImageNet dataset.
        The sequence of numbers for the CosConv and  ChebConv rows denote the number of ``harmonics'' for each of the blocks in the network.}
    \label{tab:imagenet}
\end{table}
\begin{table}[p]
    \centering
    \resizebox{0.8\linewidth}{!}{
        \begin{tabular}{cccccccc}
            \toprule
                                                                                      & Pre Top 1 & Post Top 1 & $\Delta$ \% & \# Comp. Params. & $\Delta$ \% \\
            \midrule
            Conv2D                                                                    & -         & 82.10      &             & 0.33M            & 0.0         \\
            [0.5em]
            $7_{\times3}$, $7_{\times3}$, $7_{\times9}$, $6_{\times3}$,               & 80.27     & 81.19      & -0.91       & 0.30M            & 90.90       \\
            $7_{\times3}$, $7_{\times3}$, $7_{\times4}$, $6_{\times5}$, $6_{\times3}$ & 79.10     & 80.87      & -1.23       & 0.27M            & 83.32       \\
            $7_{\times3}$, $7_{\times3}$, $6_{\times9}$, $6_{\times3}$                & 70.10     & 79.94      & -2.16       & 0.25M            & 77.25       \\
            $7_{\times3}$, $6_{\times3}$, $6_{\times9}$, $6_{\times3}$                & 49.60     & 79.44      & -2.66       & 0.25M            & 74.98       \\
            $7_{\times3}$, $7_{\times3}$, $5_{\times9}$, $5_{\times3}$                & 62.40     & 79.28      & -2.82       & 0.19M            & 58.01       \\
            $7_{\times3}$, $7_{\times3}$, $5_{\times9}$, $4_{\times3}$                & 57.67     & 78.39      & -3.71       & 0.17M            & 51.71       \\
            $7_{\times3}$, $7_{\times3}$, $4_{\times9}$, $4_{\times3}$                & 20.30     & 75.94      & -6.16       & 0.14M            & 42.26       \\
            $6_{\times3}$, $6_{\times3}$, $6_{\times9}$, $6_{\times3}$                & 24.75     & 76.26      & -5.84       & 0.24M            & 73.84       \\
            $5_{\times3}$, $5_{\times3}$, $5_{\times9}$, $5_{\times3}$                & 14.94     & 77.29      & -4.81       & 0.17M            & 51.71       \\
            \bottomrule
        \end{tabular}
    }
    \vspace{0.3em}
    \caption{Results showing our method (CosConv) used on the $7\times 7$ depth-wise filters from ConvNeXt, showing the accuracy  after the initial approximation and then after fine tuning for 5 epochs. Comp. Params.
        (Compressible Parameters) shows the number of parameters making up the $7\times 7$ convolutions in the network.
        % The linear and downsampling operations that make up the bulk of the layers in the network cannot be compressed by ours or similar methods, requiring more conventional pruning approaches.
    }
    \label{tab:convnext}
\end{table}
\begin{table}[p]
    \centering
    \resizebox{0.9\linewidth}{!}{
        \begin{tabular}{cccccccccc}
            \toprule
                 &           & \multicolumn{3}{c}{CosConv} & \multicolumn{3}{c}{ChebConv}                                                                     \\
            \cmidrule(lr){3-5}\cmidrule(lr){6-8}
            bits & Config    & Top1                        & Quant. Top1                  & Delta & Top1  & Quant. Top1 & Delta & \# Params & Model size (MB) \\
            \midrule
            32   & Conv2D    & 69.76                       & 69.76                        & 0.0   & 69.76 & 69.76       & 0.0   & 11.68M    & 46.7            \\
            [0.5em]
            8    & Conv2D    & 69.76                       & 69.74                        & -0.02 & 69.76 & 69.74       & -0.02 & 11.68M    & 23.4            \\
            8    & 6,3,3,3,2 & 68.97                       & 68.12                        & -0.85 & 67.23 & 67.39       & 0.16  & 7.09M     & 14.2            \\
            8    & 6,3,3,2,2 & 67.90                       & 65.23                        & -2.67 & 65.63 & 65.20       & -0.43 & 5.94M     & 11.9            \\
            8    & 5,3,3,3,2 & 68.53                       & 67.56                        & -0.97 & 68.15 & 66.80       & -1.35 & 7.09M     & 14.2            \\
            8    & 5,3,3,2,2 & 67.48                       & 64.69                        & -2.79 & 65.44 & 64.47       & -0.97 & 5.94M     & 11.9            \\
            [0.5em]
            4    & Conv2D    & 69.76                       & 69.28                        & -0.48 & 69.76 & 69.28       & -0.48 & 11.68M    & 17.5            \\
            4    & 6,3,3,3,2 & 68.97                       & 66.70                        & -2.27 & 67.23 & 66.50       & -0.73 & 7.09M     & 10.6            \\
            4    & 6,3,3,2,2 & 67.90                       & 64.10                        & -3.80 & 65.63 & 63.88       & -1.75 & 5.94M     & 8.9             \\
            4    & 5,3,3,3,2 & 68.53                       & 66.38                        & -2.15 & 68.15 & 65.30       & -2.85 & 7.09M     & 10.6            \\
            4    & 5,3,3,2,2 & 67.48                       & 62.76                        & -4.72 & 65.44 & 63.10       & -2.34 & 5.94M     & 8.9             \\

            \bottomrule
        \end{tabular}
    }
    \vspace{0.3em}
    \caption{
        Results showing how our method applied to ResNet-18 on the ImageNet dataset performs under quantisation.
    }
    \label{tab:quant}
\end{table}

\subsection{ImageNet}
We conduct further experiments on the ImageNet\cite{imagenet} dataset.
These experiments use both the ResNet-18\cite{resnet} and ConvNeXt-T\cite{convnext} models.
The results of our experiments on ResNet-18 are presented in \cref{tab:imagenet}.
We see that our method is able to reduce the size of the model, with a modest cost to accuracy.
Again, like ResNet-20 and ResNet-32, the structure of ResNet-18 has increasing parameter counts through the blocks, and that compressing the final layers of the network provides outsize reduction in parameter count.
Although the top configuration reduces the number of parameters in the first layer by 30\% with only a loss of 0.8 in Top 1 accuracy, the distribution of parameters in the network mean this has a barely noticeable effect on the overall parameter count (0.02\%).
However, decreasing the order further (config 5,3,3,3,3), reduces the number of parameters in the first layer by $\approx$50\% with no loss to Top 1 accuracy, but still only resulting in a 6\% reduction in model size.
The deficiencies in the initialization of the ChebConv layers is more apparent here, where the same configuration loses more accuracy than an equivalent CosConv.

\Cref{fig:approx} shows two filters from the first layer of ResNet-18 and one from the ConvNeXt-T model. We show CosConv approximations of various orders, as well as the L2 error between the approximation and the original kernel.
Whilst our method is less able to recreate the strong discontinuities found in the depth-wise ConvNeXt kernels than the less discontinuous ResNet filters, our method is able to preserve much of the filter's structure.

To better investigate the capability of our method to reduce the parameter counts of layers with a larger kernel size, we evaluate our method on the $7\times 7$ depthwise kernels that are present throughout the network, and present these results in \cref{tab:convnext}.
Because the \emph{vast} majority of the parameters in ConvNeXt pertain to operations are either pointwise/linear operations, or $2\times 2$ downsampling kernels for which our method cannot be meaningfully applied, we consider only the ``compressible parameters'' which are the parameters of the $7\times 7$ depth wise convolutions.
The results show that our method is able to compress the applicable layers of the network by 42\% with a 2.8\% loss in Top 1 accuracy.
We suspect that the impact of our method on ConvNeXt is not as strong as on ResNet-18, because ConvNeXt contains significantly more discontinuous kernels.

\subsection{Quantization}
To show that our method can complement other existing methods for reducing the size of deep models, we perform experiments to show the application of our method in tandem with quantisation.
We use a basic quantisation scheme and apply it on top of our method, quantizing the parameters of our kernel functions, as well as the activations of the network.
Further details of the setup for these experiments are presented in the supplementary material.
\Cref{tab:quant} shows our experiments quantising ResNet-18 as well as applying our method.
The `Top 1' column shows the performance of the unquantised configuration, and `Quant. Top1' shows the results after fine-tuning, applying both our method and quantisation.
The results show that a further 3-4$\times$ reduction in model size, compared to unquantised models (40\% reduction compared to quantised conventional convolution), is achieved with a 0.37\% reduction in accuracy.

\section{Conclusion}
\label{sec:conclusion}
In summary, we have presented a novel method to reduce the number of parameters required to represent a conventional 2D convolution.
Using functional approximations in the form of cosine and Chebyshev series, we are able to represent the weights of a kernel using fewer parameters.
We outline an approach to learn an initial set of weights based on pre-trained models that can be further refined through a small amount of fine-tuning.
Through experiments, we demonstrated that our method is able to reduce the size of common deep vision models by as much as $\sim50$\% with a minor reduction in accuracy, as well as the compatibility between our method and quantisation.

In future work, we hope to examine how anisotropic and variable numbers of harmonics per filter might improve the compression.
% As well, we hope to improve the motivation of the initialization of weights for both cosine and Chebyshev series.
\section*{Acknowledgements}
This work was supported by the Engineering and Physical Sciences Research Council \newline [EP/R513295/1].
We would also like to thank Henry Howard-Jenkins for constructive discussions regarding this work.

\bibliography{bib}

\begin{thebibliography}{46}
\providecommand{\natexlab}[1]{#1}
\providecommand{\url}[1]{\texttt{#1}}
\expandafter\ifx\csname urlstyle\endcsname\relax
  \providecommand{\doi}[1]{doi: #1}\else
  \providecommand{\doi}{doi: \begingroup \urlstyle{rm}\Url}\fi

\bibitem[Anwar and Sung(2016)]{anwar2016coarse}
Sajid Anwar and Wonyong Sung.
\newblock Coarse pruning of convolutional neural networks with random masks.
\newblock 2016.
\newblock URL \url{https://openreview.net/forum?id=HkvS3Mqxe}.

\bibitem[Bengio et~al.(2013)Bengio, L{\'e}onard, and
  Courville]{bengio2013estimating}
Yoshua Bengio, Nicholas L{\'e}onard, and Aaron Courville.
\newblock Estimating or propagating gradients through stochastic neurons for
  conditional computation.
\newblock \emph{arXiv preprint arXiv:1308.3432}, 2013.

\bibitem[Brown et~al.(2020)Brown, Mann, Ryder, Subbiah, Kaplan, Dhariwal,
  Neelakantan, Shyam, Sastry, Askell, Agarwal, Herbert-Voss, Krueger, Henighan,
  Child, Ramesh, Ziegler, Wu, Winter, Hesse, Chen, Sigler, Litwin, Gray, Chess,
  Clark, Berner, McCandlish, Radford, Sutskever, and Amodei]{gpt3}
Tom Brown, Benjamin Mann, Nick Ryder, Melanie Subbiah, Jared~D Kaplan, Prafulla
  Dhariwal, Arvind Neelakantan, Pranav Shyam, Girish Sastry, Amanda Askell,
  Sandhini Agarwal, Ariel Herbert-Voss, Gretchen Krueger, Tom Henighan, Rewon
  Child, Aditya Ramesh, Daniel Ziegler, Jeffrey Wu, Clemens Winter, Chris
  Hesse, Mark Chen, Eric Sigler, Mateusz Litwin, Scott Gray, Benjamin Chess,
  Jack Clark, Christopher Berner, Sam McCandlish, Alec Radford, Ilya Sutskever,
  and Dario Amodei.
\newblock Language models are few-shot learners.
\newblock In \emph{Advances in Neural Information Processing Systems},
  volume~33, pages 1877--1901. Curran Associates, Inc., 2020.

\bibitem[Caron et~al.(2021)Caron, Touvron, Misra, J\'egou, Mairal, Bojanowski,
  and Joulin]{dino}
Mathilde Caron, Hugo Touvron, Ishan Misra, Herv\'e J\'egou, Julien Mairal,
  Piotr Bojanowski, and Armand Joulin.
\newblock Emerging properties in self-supervised vision transformers.
\newblock In \emph{Proceedings of the IEEE/CVF International Conference on
  Computer Vision (ICCV)}, pages 9650--9660, October 2021.

\bibitem[Deng et~al.(2009)Deng, Dong, Socher, Li, Li, and Fei-Fei]{imagenet}
Jia Deng, Wei Dong, Richard Socher, Li-Jia Li, Kai Li, and Li~Fei-Fei.
\newblock Imagenet: A large-scale hierarchical image database.
\newblock In \emph{2009 IEEE Conference on Computer Vision and Pattern
  Recognition}, pages 248--255, 2009.
\newblock \doi{10.1109/CVPR.2009.5206848}.

\bibitem[Denton et~al.(2014)Denton, Zaremba, Bruna, LeCun, and
  Fergus]{denton2014exploiting}
Emily~L Denton, Wojciech Zaremba, Joan Bruna, Yann LeCun, and Rob Fergus.
\newblock Exploiting linear structure within convolutional networks for
  efficient evaluation.
\newblock \emph{Advances in neural information processing systems}, 27, 2014.

\bibitem[Dosovitskiy et~al.(2021)Dosovitskiy, Beyer, Kolesnikov, Weissenborn,
  Zhai, Unterthiner, Dehghani, Minderer, Heigold, Gelly, Uszkoreit, and
  Houlsby]{vit}
Alexey Dosovitskiy, Lucas Beyer, Alexander Kolesnikov, Dirk Weissenborn,
  Xiaohua Zhai, Thomas Unterthiner, Mostafa Dehghani, Matthias Minderer, Georg
  Heigold, Sylvain Gelly, Jakob Uszkoreit, and Neil Houlsby.
\newblock An image is worth 16x16 words: Transformers for image recognition at
  scale.
\newblock In \emph{International Conference on Learning Representations}, 2021.

\bibitem[Drumond et~al.(2018)Drumond, Lin, Jaggi, and
  Falsafi]{drumond2018training}
Mario Drumond, Tao Lin, Martin Jaggi, and Babak Falsafi.
\newblock Training dnns with hybrid block floating point.
\newblock \emph{Advances in Neural Information Processing Systems}, 31, 2018.

\bibitem[Elthakeb et~al.(2020)Elthakeb, Pilligundla, Mireshghallah,
  Yazdanbakhsh, and Esmaeilzadeh]{elthakeb2020releq}
Ahmed~T Elthakeb, Prannoy Pilligundla, Fatemehsadat Mireshghallah, Amir
  Yazdanbakhsh, and Hadi Esmaeilzadeh.
\newblock Releq: A reinforcement learning approach for automatic deep
  quantization of neural networks.
\newblock \emph{IEEE micro}, 40\penalty0 (5):\penalty0 37--45, 2020.

\bibitem[Fey et~al.(2018)Fey, Lenssen, Weichert, and M{\"u}ller]{splinecnn}
Matthias Fey, Jan~Eric Lenssen, Frank Weichert, and Heinrich M{\"u}ller.
\newblock Splinecnn: Fast geometric deep learning with continuous b-spline
  kernels.
\newblock In \emph{Proceedings of the IEEE conference on computer vision and
  pattern recognition}, pages 869--877, 2018.

\bibitem[Frankle and Carbin(2018)]{frankle2018lottery}
Jonathan Frankle and Michael Carbin.
\newblock The lottery ticket hypothesis: Finding sparse, trainable neural
  networks.
\newblock In \emph{International Conference on Learning Representations}, 2018.

\bibitem[Gennari~do Nascimento et~al.(2020)Gennari~do Nascimento, Costain, and
  Prisacariu]{gennari2020finding}
Marcelo Gennari~do Nascimento, Theo~W Costain, and Victor~Adrian Prisacariu.
\newblock Finding non-uniform quantization schemes using multi-task gaussian
  processes.
\newblock In \emph{European Conference on Computer Vision}, pages 383--398.
  Springer, 2020.

\bibitem[Groh et~al.(2018)Groh, Wieschollek, and Lensch]{flexconv}
Fabian Groh, Patrick Wieschollek, and Hendrik P.~A. Lensch.
\newblock Flex-convolution (million-scale point-cloud learning beyond
  grid-worlds).
\newblock In \emph{Asian Conference on Computer Vision (ACCV)}, Dezember 2018.

\bibitem[Han et~al.(2015)Han, Pool, Tran, and Dally]{han2015learning}
Song Han, Jeff Pool, John Tran, and William Dally.
\newblock Learning both weights and connections for efficient neural network.
\newblock \emph{Advances in neural information processing systems}, 28, 2015.

\bibitem[Han et~al.(2016)Han, Mao, and Dally]{han2015deep_compression}
Song Han, Huizi Mao, and William~J Dally.
\newblock Deep compression: Compressing deep neural networks with pruning,
  trained quantization and huffman coding.
\newblock \emph{International Conference on Learning Representations (ICLR)},
  2016.

\bibitem[He et~al.(2016)He, Zhang, Ren, and Sun]{resnet}
Kaiming He, Xiangyu Zhang, Shaoqing Ren, and Jian Sun.
\newblock Deep residual learning for image recognition.
\newblock In \emph{Proceedings of the IEEE conference on computer vision and
  pattern recognition}, pages 770--778, 2016.

\bibitem[He et~al.(2017)He, Zhang, and Sun]{he2017channel}
Yihui He, Xiangyu Zhang, and Jian Sun.
\newblock Channel pruning for accelerating very deep neural networks.
\newblock In \emph{Proceedings of the IEEE international conference on computer
  vision}, pages 1389--1397, 2017.

\bibitem[Hinton et~al.(2015)Hinton, Vinyals, and Dean]{hinton2015distil}
Geoffrey Hinton, Oriol Vinyals, and Jeffrey Dean.
\newblock Distilling the knowledge in a neural network.
\newblock In \emph{NIPS Deep Learning and Representation Learning Workshop},
  2015.

\bibitem[Howard et~al.(2017)Howard, Zhu, Chen, Kalenichenko, Wang, Weyand,
  Andreetto, and Adam]{mobilenets}
Andrew~G Howard, Menglong Zhu, Bo~Chen, Dmitry Kalenichenko, Weijun Wang,
  Tobias Weyand, Marco Andreetto, and Hartwig Adam.
\newblock Mobilenets: Efficient convolutional neural networks for mobile vision
  applications.
\newblock \emph{arXiv preprint arXiv:1704.04861}, 2017.

\bibitem[Huang et~al.(2017)Huang, Liu, Van Der~Maaten, and
  Weinberger]{huang2017densely}
Gao Huang, Zhuang Liu, Laurens Van Der~Maaten, and Kilian~Q Weinberger.
\newblock Densely connected convolutional networks.
\newblock In \emph{Proceedings of the IEEE conference on computer vision and
  pattern recognition}, pages 4700--4708, 2017.

\bibitem[Iandola et~al.(2016)Iandola, Han, Moskewicz, Ashraf, Dally, and
  Keutzer]{squeezenet}
Forrest~N Iandola, Song Han, Matthew~W Moskewicz, Khalid Ashraf, William~J
  Dally, and Kurt Keutzer.
\newblock Squeezenet: Alexnet-level accuracy with 50x fewer parameters and< 0.5
  mb model size.
\newblock \emph{arXiv preprint arXiv:1602.07360}, 2016.

\bibitem[Jacobsen et~al.(2016)Jacobsen, Van~Gemert, Lou, and Smeulders]{SRF}
Jorn-Henrik Jacobsen, Jan Van~Gemert, Zhongyu Lou, and Arnold~WM Smeulders.
\newblock Structured receptive fields in cnns.
\newblock In \emph{Proceedings of the IEEE Conference on Computer Vision and
  Pattern Recognition}, pages 2610--2619, 2016.

\bibitem[Jaderberg et~al.(2014)Jaderberg, Vedaldi, and
  Zisserman]{jaderberg2014speeding}
Max Jaderberg, Andrea Vedaldi, and Andrew Zisserman.
\newblock Speeding up convolutional neural networks with low rank expansions.
\newblock In \emph{Proceedings of the British Machine Vision Conference. BMVA
  Press}, 2014.

\bibitem[Krizhevsky et~al.(2009)Krizhevsky, Hinton, et~al.]{cifar}
Alex Krizhevsky, Geoffrey Hinton, et~al.
\newblock Learning multiple layers of features from tiny images.
\newblock 2009.

\bibitem[LeCun et~al.(1989)LeCun, Denker, and Solla]{lecun1989optimal}
Yann LeCun, John Denker, and Sara Solla.
\newblock Optimal brain damage.
\newblock \emph{Advances in neural information processing systems}, 2, 1989.

\bibitem[Liu et~al.(2022)Liu, Mao, Wu, Feichtenhofer, Darrell, and
  Xie]{convnext}
Zhuang Liu, Hanzi Mao, Chao-Yuan Wu, Christoph Feichtenhofer, Trevor Darrell,
  and Saining Xie.
\newblock A convnet for the 2020s.
\newblock In \emph{Proceedings of the IEEE/CVF Conference on Computer Vision
  and Pattern Recognition (CVPR)}, pages 11976--11986, June 2022.

\bibitem[Luo et~al.(2018)Luo, Zhang, Zhou, Xie, Wu, and Lin]{luo2018thinet}
Jian-Hao Luo, Hao Zhang, Hong-Yu Zhou, Chen-Wei Xie, Jianxin Wu, and Weiyao
  Lin.
\newblock Thinet: pruning cnn filters for a thinner net.
\newblock \emph{IEEE transactions on pattern analysis and machine
  intelligence}, 41\penalty0 (10):\penalty0 2525--2538, 2018.

\bibitem[Malach et~al.(2020)Malach, Yehudai, Shalev-Schwartz, and
  Shamir]{malach2020proving}
Eran Malach, Gilad Yehudai, Shai Shalev-Schwartz, and Ohad Shamir.
\newblock Proving the lottery ticket hypothesis: Pruning is all you need.
\newblock In \emph{International Conference on Machine Learning}, pages
  6682--6691. PMLR, 2020.

\bibitem[Mirzadeh et~al.(2020)Mirzadeh, Farajtabar, Li, Levine, Matsukawa, and
  Ghasemzadeh]{mirzadeh2020improved}
Seyed~Iman Mirzadeh, Mehrdad Farajtabar, Ang Li, Nir Levine, Akihiro Matsukawa,
  and Hassan Ghasemzadeh.
\newblock Improved knowledge distillation via teacher assistant.
\newblock In \emph{Proceedings of the AAAI conference on artificial
  intelligence}, pages 5191--5198, 2020.

\bibitem[Nascimento et~al.(2019)Nascimento, Fawcett, and
  Prisacariu]{nascimento2019dsconv}
Marcelo Gennari~do Nascimento, Roger Fawcett, and Victor~Adrian Prisacariu.
\newblock Dsconv: efficient convolution operator.
\newblock In \emph{Proceedings of the IEEE/CVF International Conference on
  Computer Vision}, pages 5148--5157, 2019.

\bibitem[Norouzzadeh et~al.(2018)Norouzzadeh, Nguyen, Kosmala, Swanson, Palmer,
  Packer, and Clune]{conservation}
Mohammad~Sadegh Norouzzadeh, Anh Nguyen, Margaret Kosmala, Alexandra Swanson,
  Meredith~S Palmer, Craig Packer, and Jeff Clune.
\newblock Automatically identifying, counting, and describing wild animals in
  camera-trap images with deep learning.
\newblock \emph{Proceedings of the National Academy of Sciences}, 115\penalty0
  (25):\penalty0 E5716--E5725, 2018.

\bibitem[Romero et~al.(2015)Romero, Ballas, Kahou, Chassang, Gatta, and
  Bengio]{romero2015fitnets}
Adriana Romero, Nicolas Ballas, Samira~Ebrahimi Kahou, Antoine Chassang, Carlo
  Gatta, and Yoshua Bengio.
\newblock Fitnets: Hints for thin deep nets.
\newblock In Yoshua Bengio and Yann LeCun, editors, \emph{3rd International
  Conference on Learning Representations, {ICLR} 2015, San Diego, CA, USA, May
  7-9, 2015, Conference Track Proceedings}, 2015.

\bibitem[Ronneberger et~al.(2015)Ronneberger, Fischer, and Brox]{unet}
Olaf Ronneberger, Philipp Fischer, and Thomas Brox.
\newblock U-net: Convolutional networks for biomedical image segmentation.
\newblock In \emph{International Conference on Medical image computing and
  computer-assisted intervention}, pages 234--241. Springer, 2015.

\bibitem[Sainath et~al.(2013)Sainath, Kingsbury, Sindhwani, Arisoy, and
  Ramabhadran]{sainath2013low}
Tara~N Sainath, Brian Kingsbury, Vikas Sindhwani, Ebru Arisoy, and Bhuvana
  Ramabhadran.
\newblock Low-rank matrix factorization for deep neural network training with
  high-dimensional output targets.
\newblock In \emph{2013 IEEE international conference on acoustics, speech and
  signal processing}, pages 6655--6659. IEEE, 2013.

\bibitem[Saldanha et~al.(2021)Saldanha, Pintea, van Gemert, and Tomen]{FracSRF}
Nikhil Saldanha, Silvia~L Pintea, Jan~C van Gemert, and Nergis Tomen.
\newblock Frequency learning for structured cnn filters with gaussian
  fractional derivatives.
\newblock In \emph{BMVC}, 2021.

\bibitem[Simonovsky and Komodakis(2017)]{edgeconv}
Martin Simonovsky and Nikos Komodakis.
\newblock Dynamic edge-conditioned filters in convolutional neural networks on
  graphs.
\newblock In \emph{Proceedings of the IEEE conference on computer vision and
  pattern recognition}, pages 3693--3702, 2017.

\bibitem[Simonyan and Zisserman(2015)]{vgg}
Karen Simonyan and Andrew Zisserman.
\newblock Very deep convolutional networks for large-scale image recognition.
\newblock In \emph{International Conference on Learning Representations}, 2015.

\bibitem[Song et~al.(2018)Song, Liu, and Wang]{song2018computation}
Zhourui Song, Zhenyu Liu, and Dongsheng Wang.
\newblock Computation error analysis of block floating point arithmetic
  oriented convolution neural network accelerator design.
\newblock In \emph{Proceedings of the AAAI Conference on Artificial
  Intelligence}, volume~32, 2018.

\bibitem[Trefethen(2019)]{TrefethenLloydN2019Ataa}
Lloyd~N Trefethen.
\newblock \emph{Approximation theory and approximation practice [electronic
  resource]}.
\newblock Other titles in applied mathematics. SIAM, Philadelphia, extended
  edition. edition, 2019.
\newblock ISBN 9781611975949 (ebook).

\bibitem[Wang et~al.(2019)Wang, Liu, Lin, Lin, and Han]{wang2019haq}
Kuan Wang, Zhijian Liu, Yujun Lin, Ji~Lin, and Song Han.
\newblock Haq: Hardware-aware automated quantization with mixed precision.
\newblock In \emph{Proceedings of the IEEE/CVF Conference on Computer Vision
  and Pattern Recognition}, pages 8612--8620, 2019.

\bibitem[Zamora et~al.(2021)Zamora, Vargas, Rhodes, Nachman, and
  Sundararajan]{Fractional}
Julio Zamora, Jesus A.~Cruz Vargas, Anthony Rhodes, Lama Nachman, and Narayan
  Sundararajan.
\newblock Convolutional filter approximation using fractional calculus.
\newblock In \emph{Proceedings of the IEEE/CVF International Conference on
  Computer Vision (ICCV) Workshops}, pages 383--392, October 2021.

\bibitem[Zhang et~al.(2019)Zhang, Song, Gao, Chen, Bao, and Ma]{zhang2019your}
Linfeng Zhang, Jiebo Song, Anni Gao, Jingwei Chen, Chenglong Bao, and Kaisheng
  Ma.
\newblock Be your own teacher: Improve the performance of convolutional neural
  networks via self distillation.
\newblock In \emph{Proceedings of the IEEE/CVF International Conference on
  Computer Vision}, pages 3713--3722, 2019.

\bibitem[Zhang et~al.(2015)Zhang, Zou, He, and Sun]{zhang2015accelerating}
Xiangyu Zhang, Jianhua Zou, Kaiming He, and Jian Sun.
\newblock Accelerating very deep convolutional networks for classification and
  detection.
\newblock \emph{IEEE transactions on pattern analysis and machine
  intelligence}, 38\penalty0 (10):\penalty0 1943--1955, 2015.

\bibitem[Zhang et~al.(2018)Zhang, Zhou, Lin, and Sun]{shufflenet}
Xiangyu Zhang, Xinyu Zhou, Mengxiao Lin, and Jian Sun.
\newblock Shufflenet: An extremely efficient convolutional neural network for
  mobile devices.
\newblock In \emph{Proceedings of the IEEE conference on computer vision and
  pattern recognition}, pages 6848--6856, 2018.

\bibitem[Zhou et~al.(2016)Zhou, Wu, Ni, Zhou, Wen, and Zou]{zhou2016dorefa}
Shuchang Zhou, Yuxin Wu, Zekun Ni, Xinyu Zhou, He~Wen, and Yuheng Zou.
\newblock Dorefa-net: Training low bitwidth convolutional neural networks with
  low bitwidth gradients.
\newblock \emph{arXiv preprint arXiv:1606.06160}, 2016.

\bibitem[Zhou et~al.(2021)Zhou, Hu, Han, Wang, and Duan]{zhou2021high}
Yue Zhou, Xiaofang Hu, Jiaqi Han, Lidan Wang, and Shukai Duan.
\newblock High frequency patterns play a key role in the generation of
  adversarial examples.
\newblock \emph{Neurocomputing}, 459:\penalty0 131--141, 2021.

\end{thebibliography}

\newpage
\FloatBarrier
\appendix
\section{Datasets}
We use two common datasets in our experiments: CIFAR-10\cite{cifar} and ImageNet\cite{imagenet}.

CIFAR-10 consists 50,000 training images and 10,000 validation images, distributed across 10 categories.
During fine-tuning, images are randomly cropped to $32\times32$, randomly horizontally flipped and normalised.
During validation/testing images are normalised identically to training.

The much larger ImageNet dataset consists of 1.28M training images and 50,000 validation images across 1000 categories.
During fine-tuning, images are randomly cropped to $224\times224$, randomly horizontally flipped and normalised.
At test time, images are center cropped to the same size, and normalised identically to training.

% Following other works, we report our results for both datasets using the validation split.

\section{Training Schedules}
For ResNet-20, ResNet-32, and ResNet-18 we fine-tune using an SGD optimiser with the learning rate set to \num{1e-4}, a momentum of 0.9, and weight decay of \num{5e-4}. We reduce the learning rate by a factor of 10 after 3 iterations, although this had marginal impact on the final validation error.

For ConvNeXt-T, we fine-tune with with the same optimiser as above, but with the learning rate set to \num{5e-5}.
We also do not use any of the data augmentations from the paper\cite{convnext}, except for Stochastic Depth\cite{huang2017densely}.

\section{Quantisation}
For our quantisation we make use of the BFP quantisation scheme from \cite{nascimento2019dsconv,song2018computation,drumond2018training}, with a block size of 1.
We quantize both weights and activations.
In the backwards pass, we use the straight through estimator\cite{bengio2013estimating} to calculate the gradients.
% We found in our early experiments that using blocks larger than 1 over the input channel dimensions significantly impaired our method as the different harmonics tend to have significant differences in scale.
We use 7 bits for the exponent in all configurations, and quantize all weights and activations to the same number of bits.

\section{Further Kernel Visualisations}
\begin{figure}[thpb]
    \centering
    \includegraphics[width=0.24\linewidth]{images/visfig/orig_stages.0.0.dwconv.weight_23_0.eps}
    \includegraphics[width=0.24\linewidth]{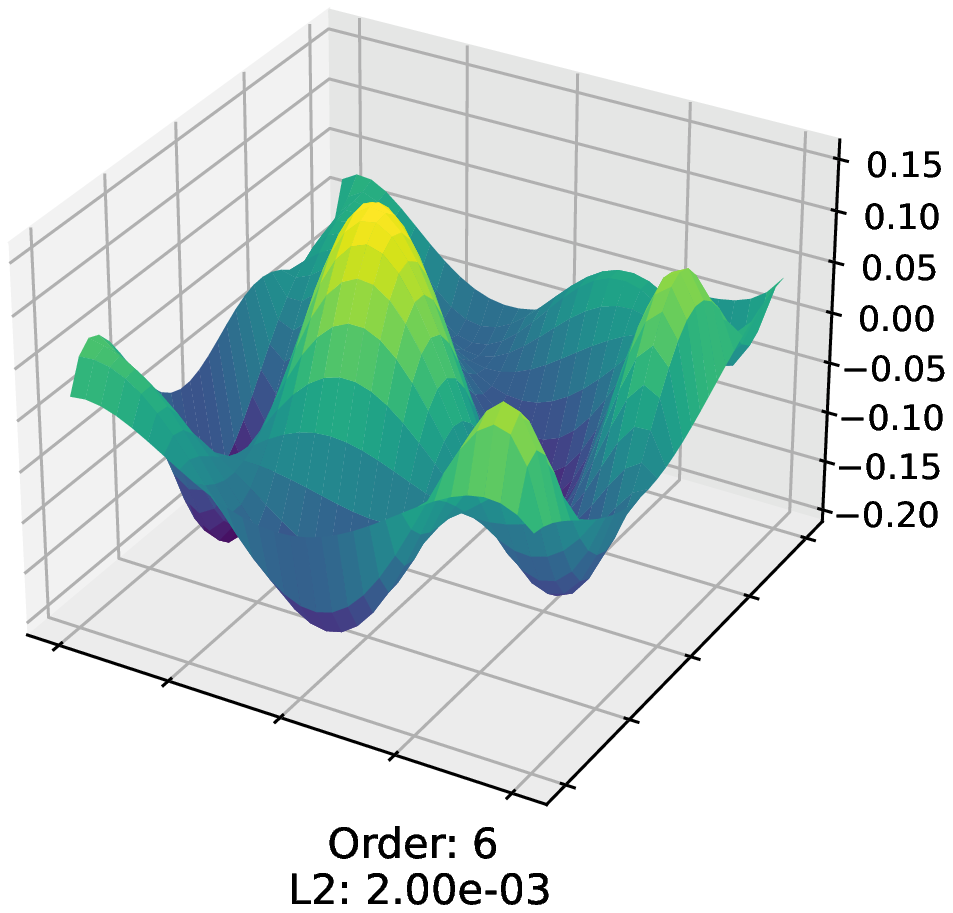}
    \includegraphics[width=0.24\linewidth]{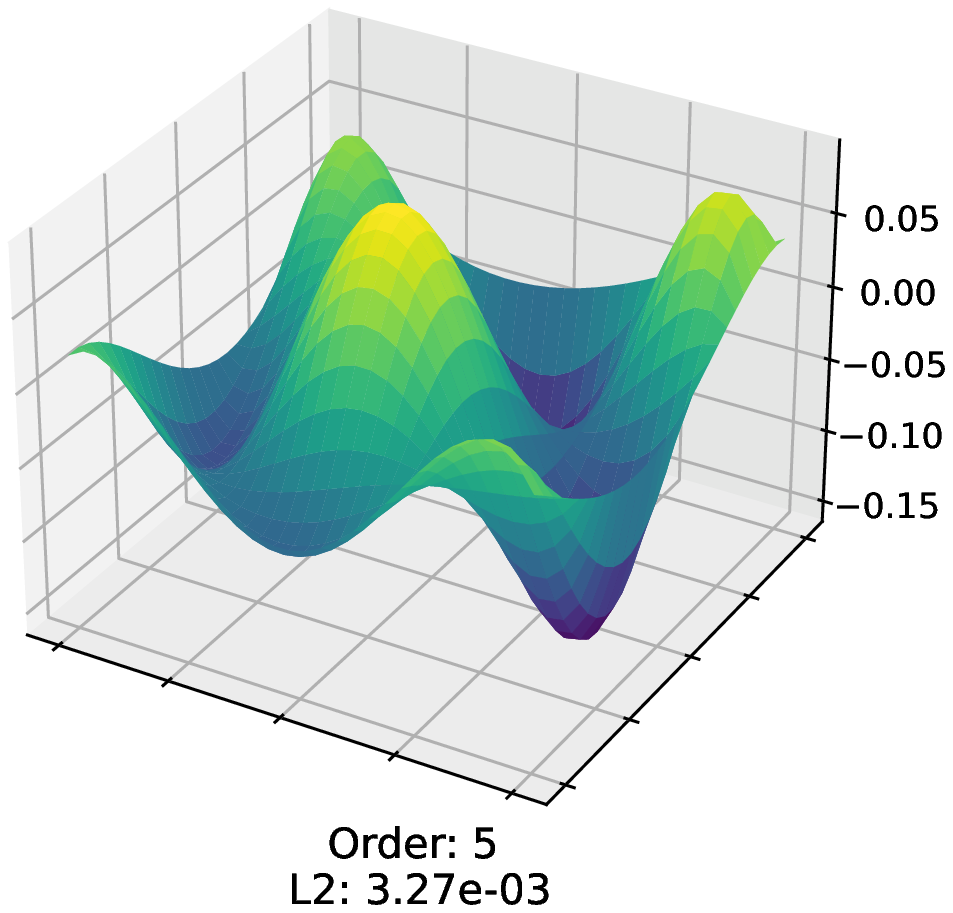}
    \includegraphics[width=0.24\linewidth]{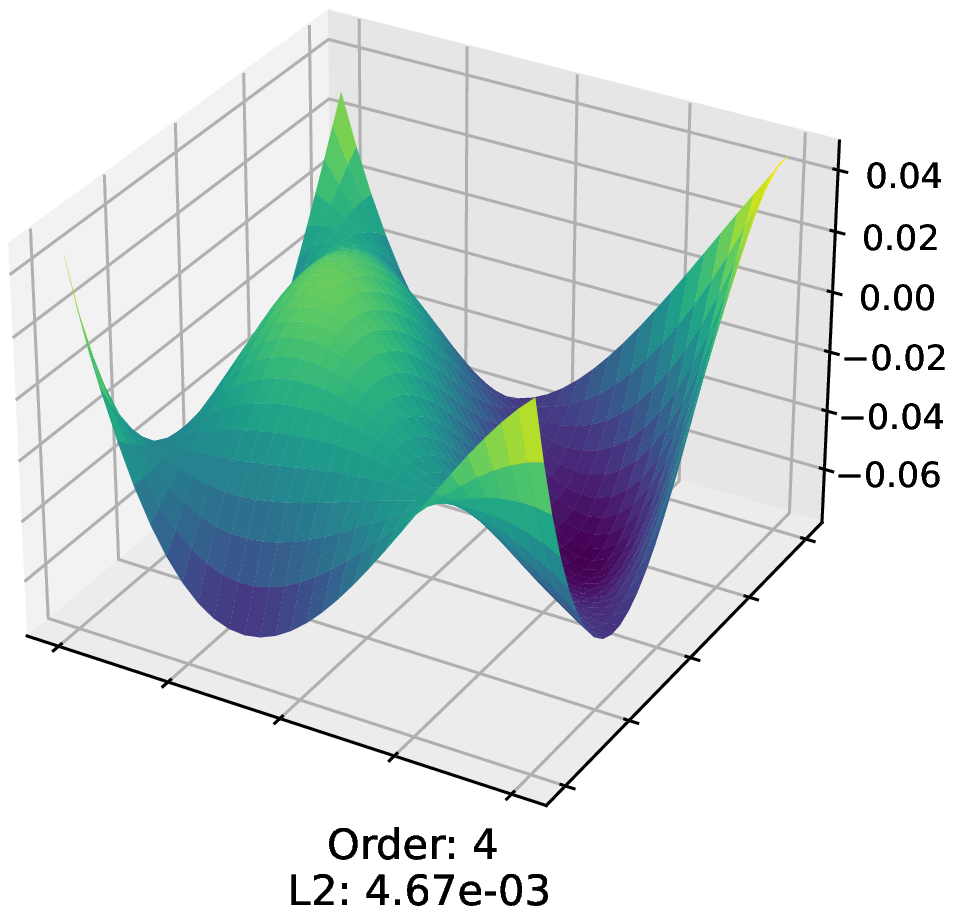}

    \includegraphics[width=0.24\linewidth]{images/visfig/orig_conv1.weight_53_0.eps}
    \includegraphics[width=0.24\linewidth]{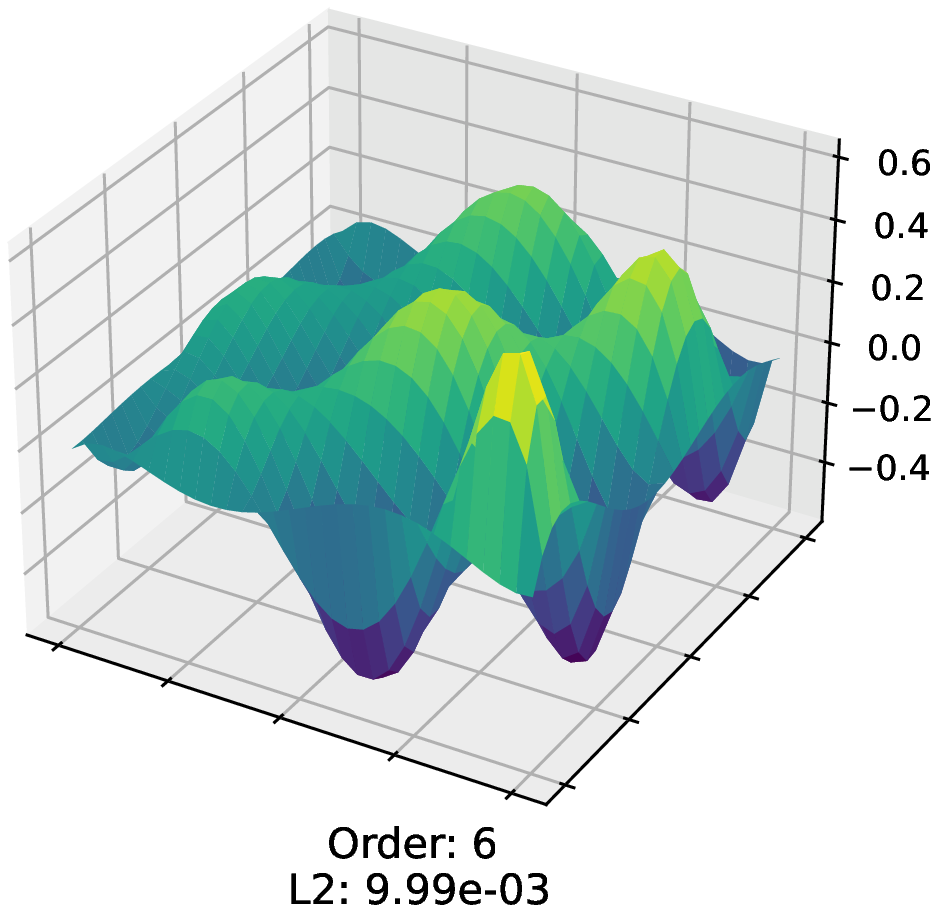}
    \includegraphics[width=0.24\linewidth]{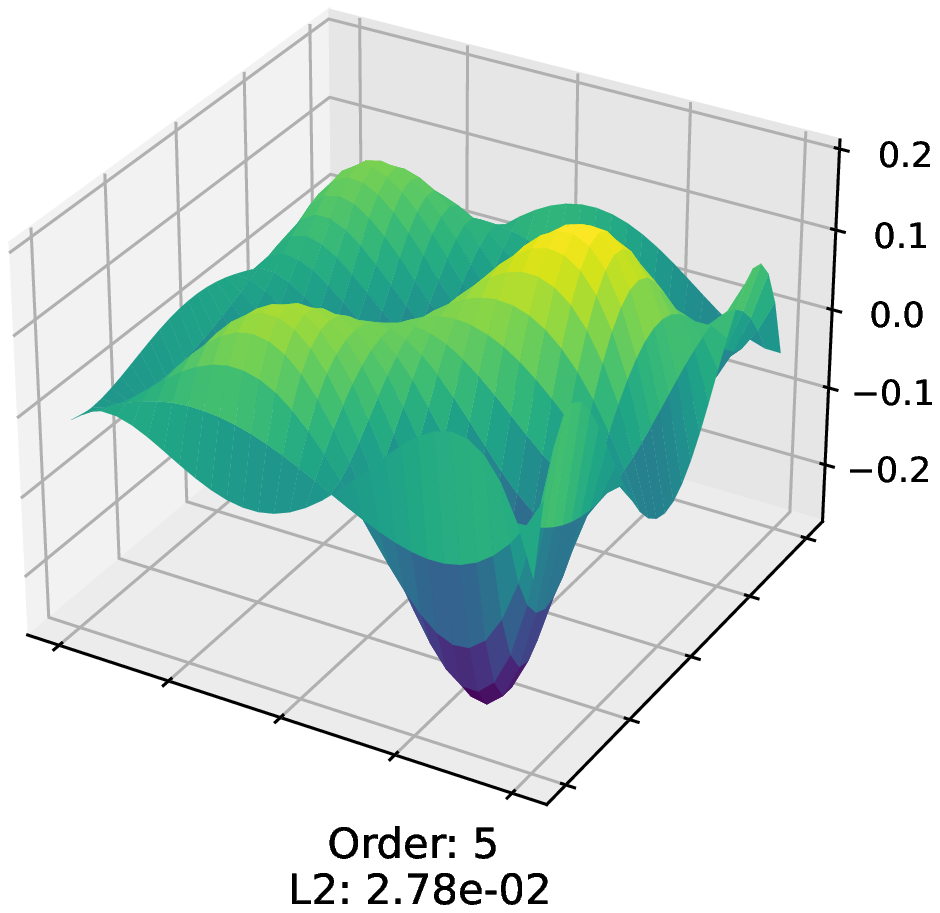}
    \includegraphics[width=0.24\linewidth]{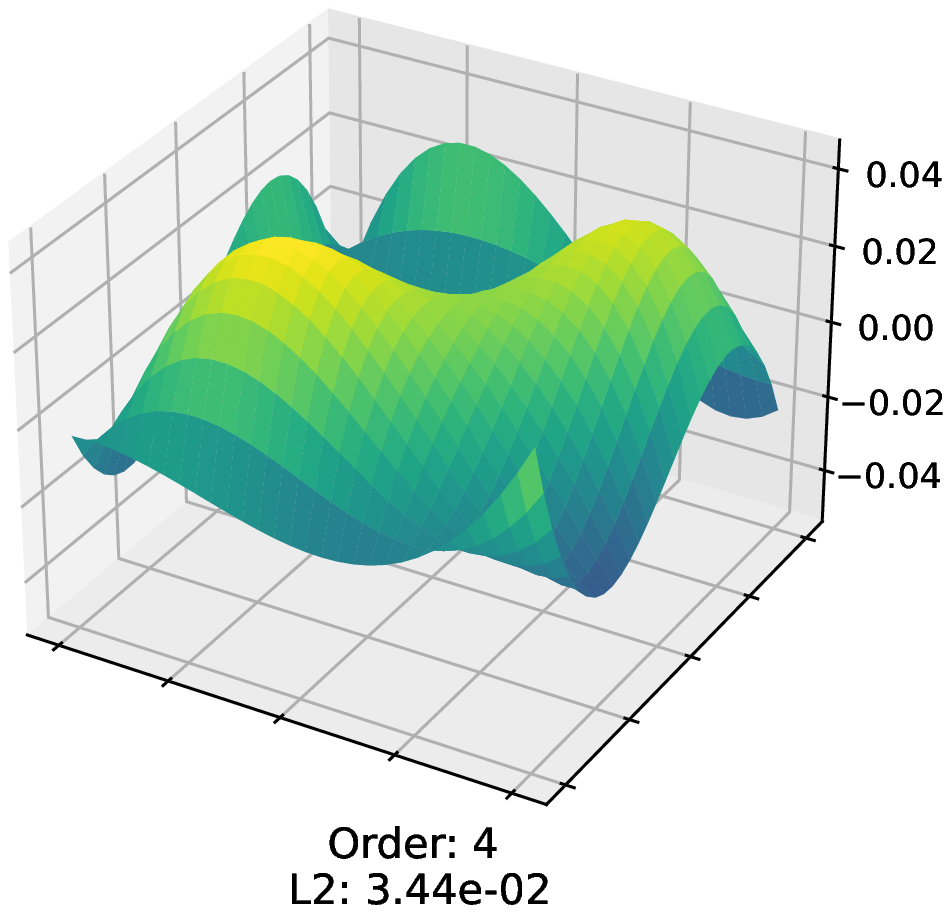}

    \includegraphics[width=0.24\linewidth]{images/visfig/orig_conv1.weight_23_2.eps}
    \includegraphics[width=0.24\linewidth]{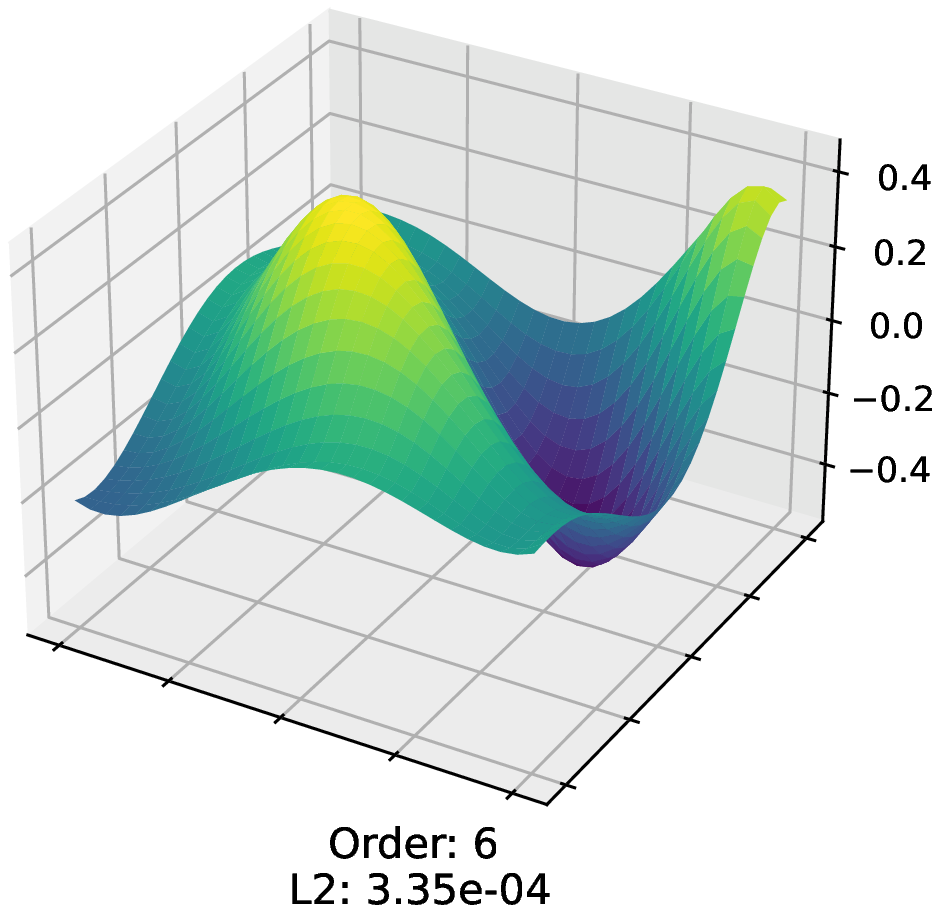}
    \includegraphics[width=0.24\linewidth]{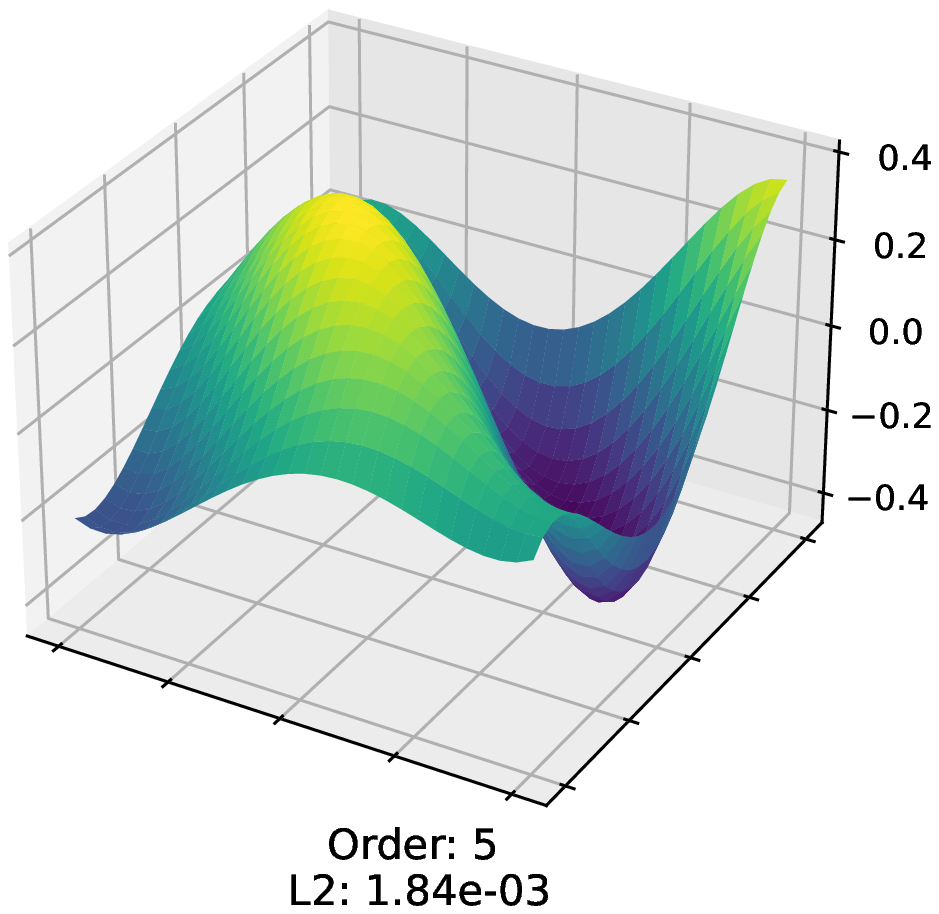}
    \includegraphics[width=0.24\linewidth]{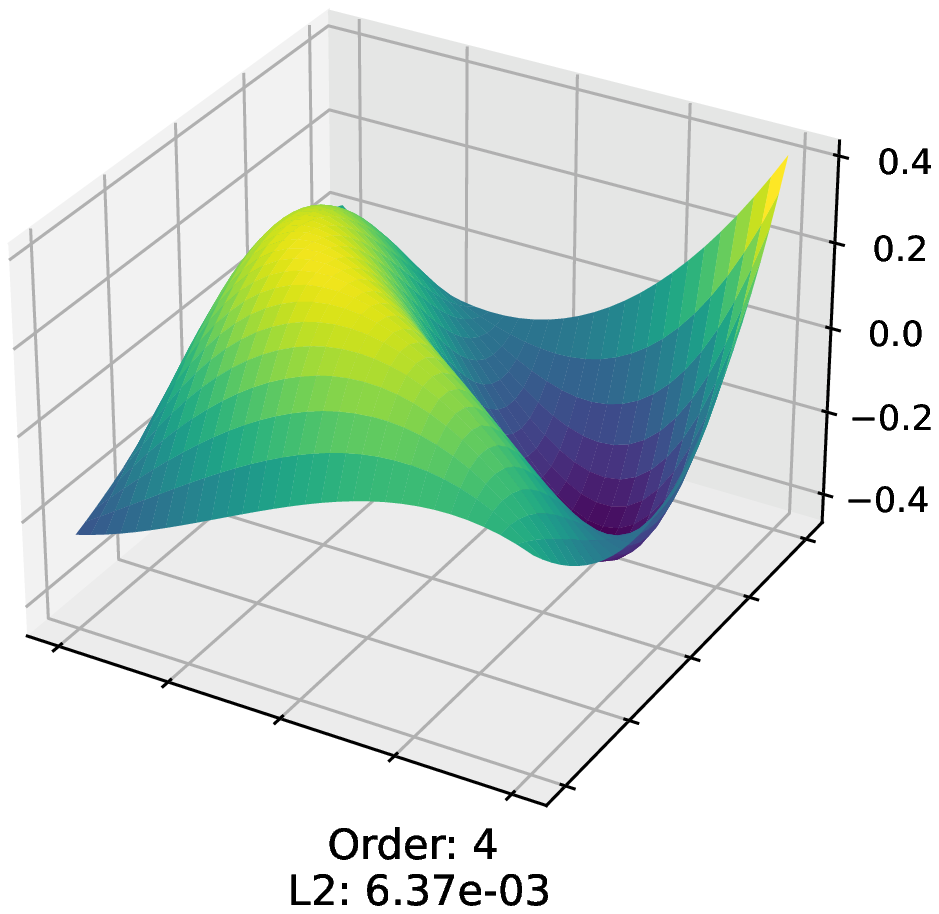}

    \caption{Filters from the ConvNeXt and ResNet networks compressed using our method.
        Left: Original Kernel, Right (in columns): Compressed version of this kernel using ChebConv with 6, 5, and 4 orders/harmonics respectively.}
    \label{fig:chebvis}
\end{figure}

\begin{figure}[tpb]
    \centering
    \includegraphics[width=0.24\linewidth]{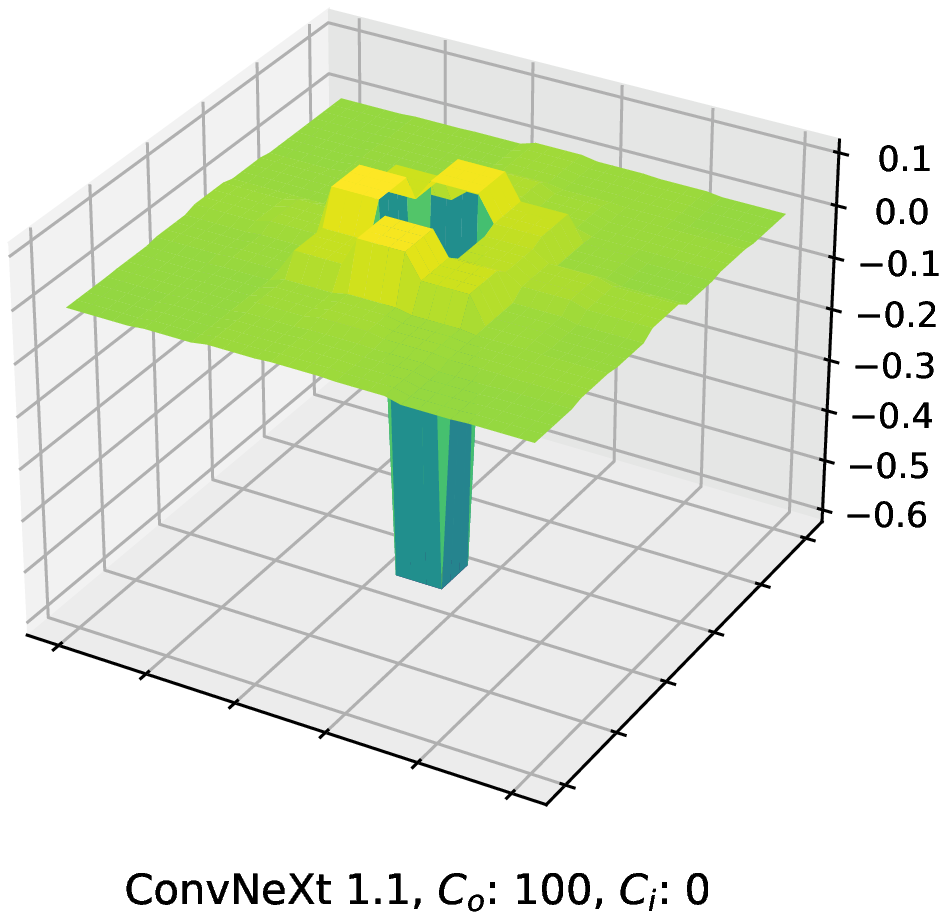}
    \includegraphics[width=0.24\linewidth]{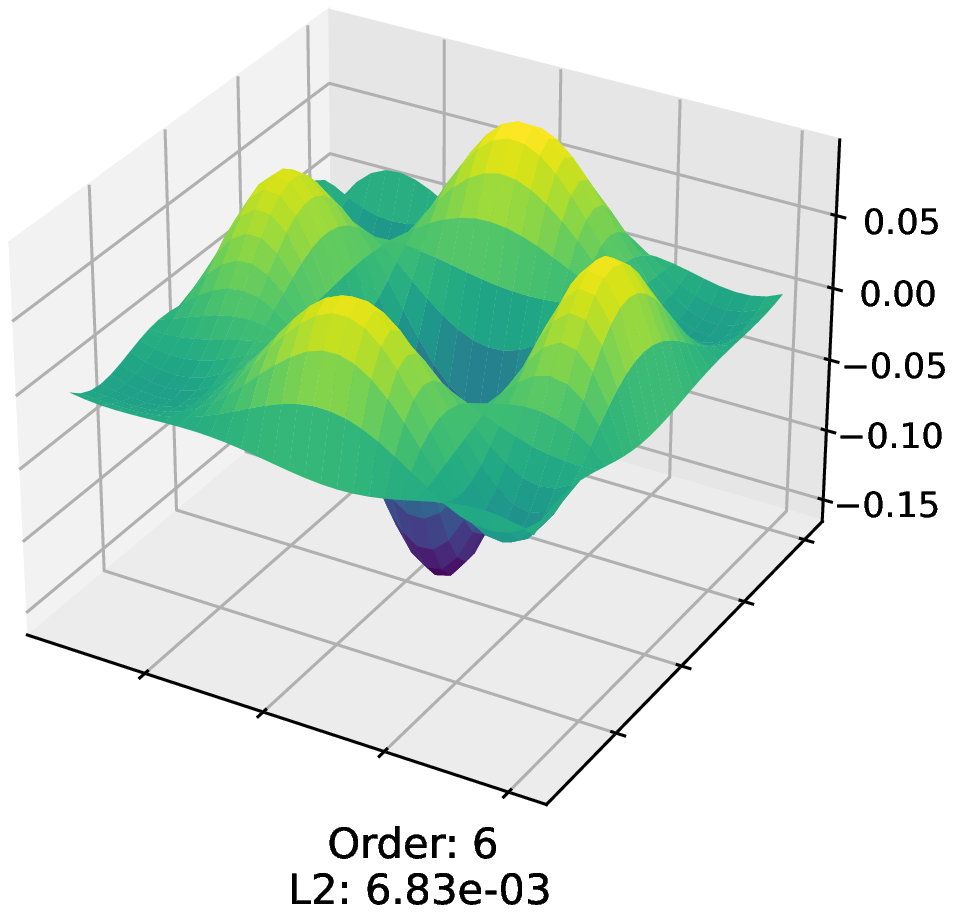}
    \includegraphics[width=0.24\linewidth]{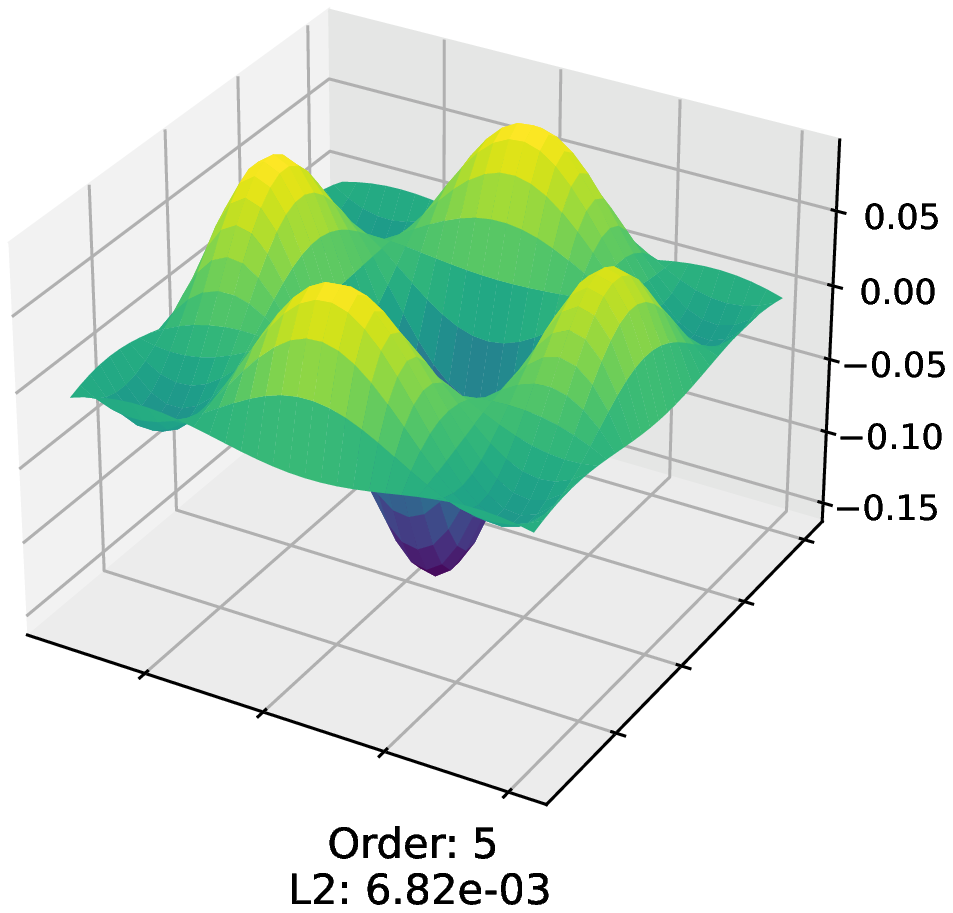}
    \includegraphics[width=0.24\linewidth]{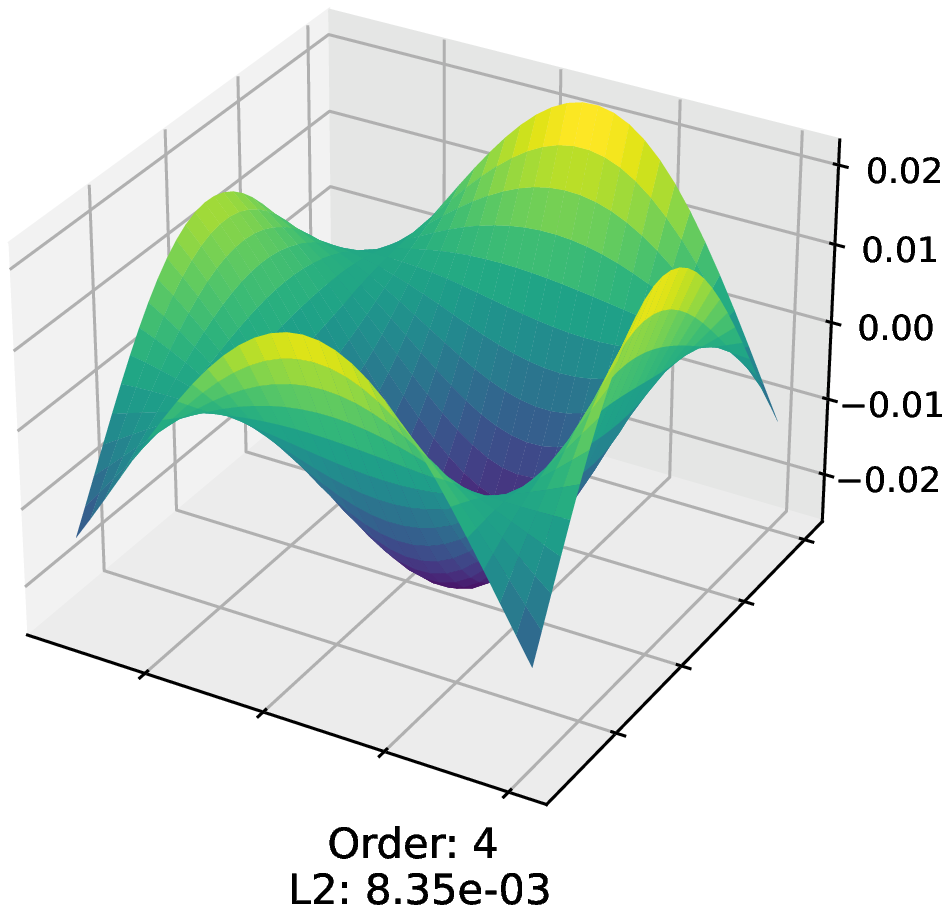}

    \includegraphics[width=0.24\linewidth]{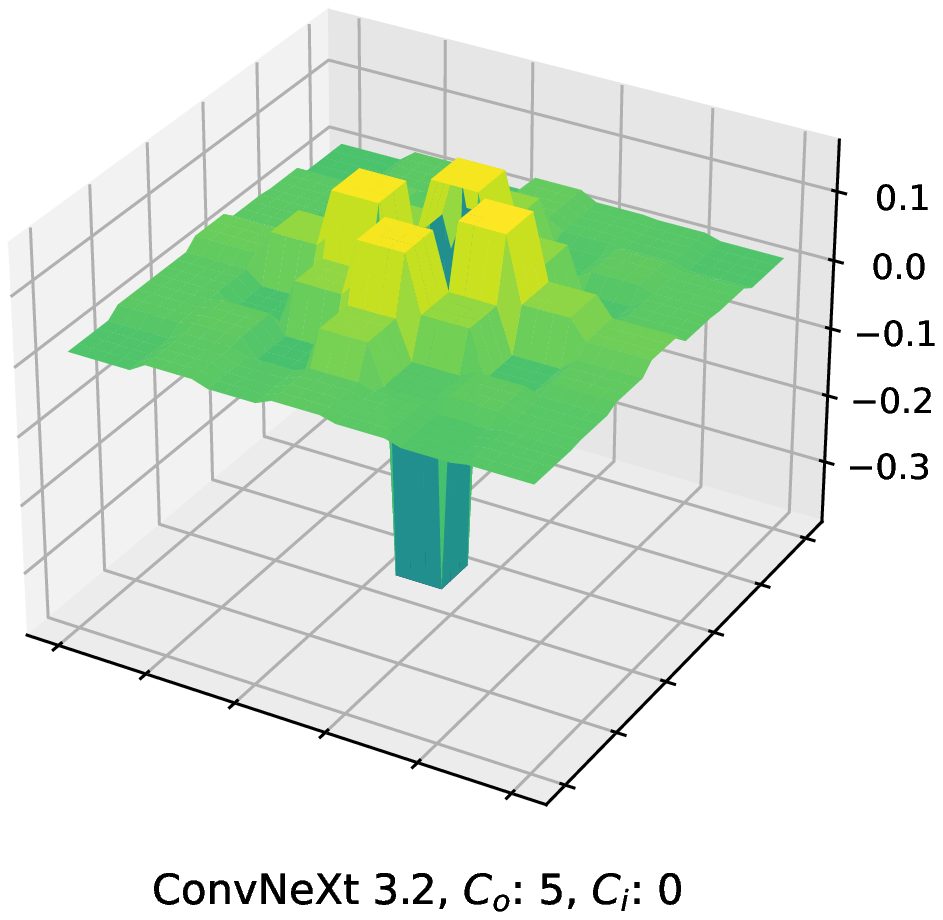}
    \includegraphics[width=0.24\linewidth]{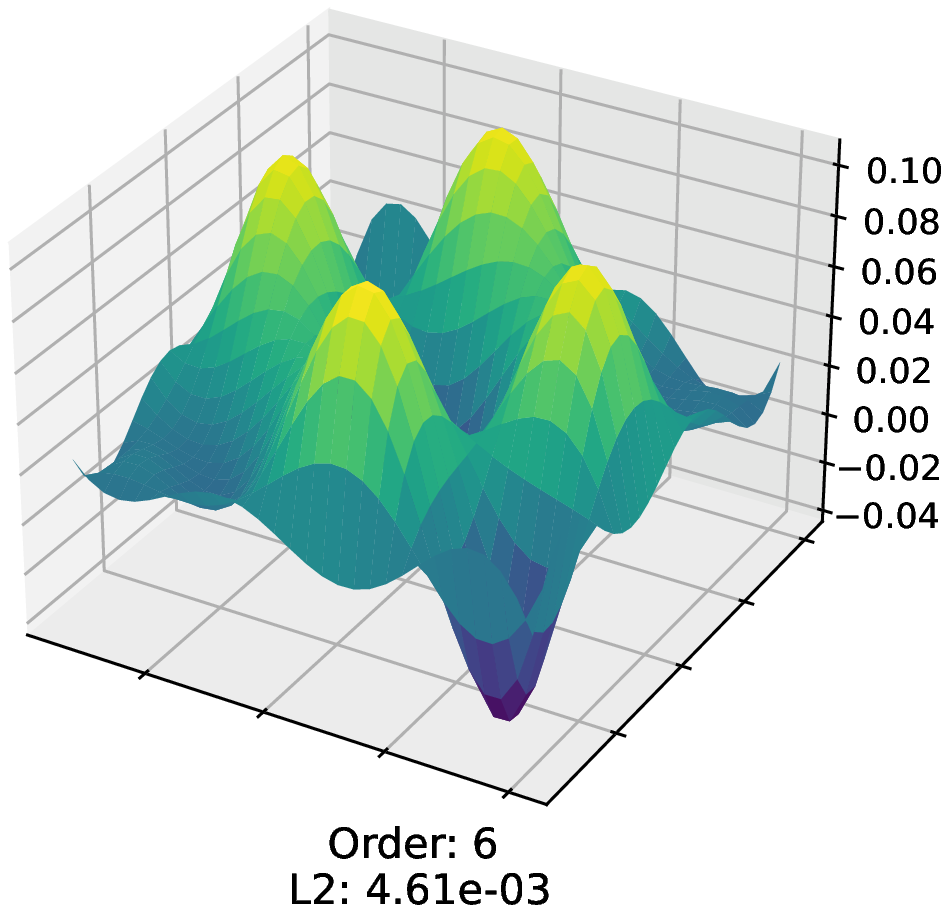}
    \includegraphics[width=0.24\linewidth]{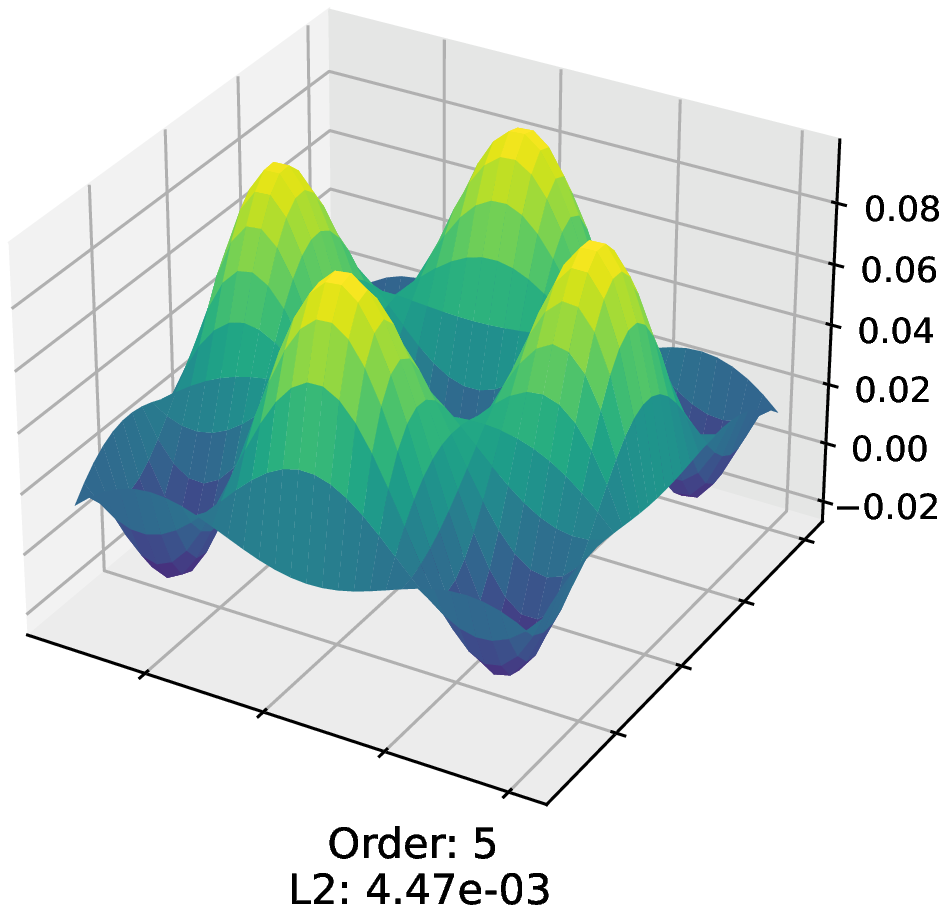}
    \includegraphics[width=0.24\linewidth]{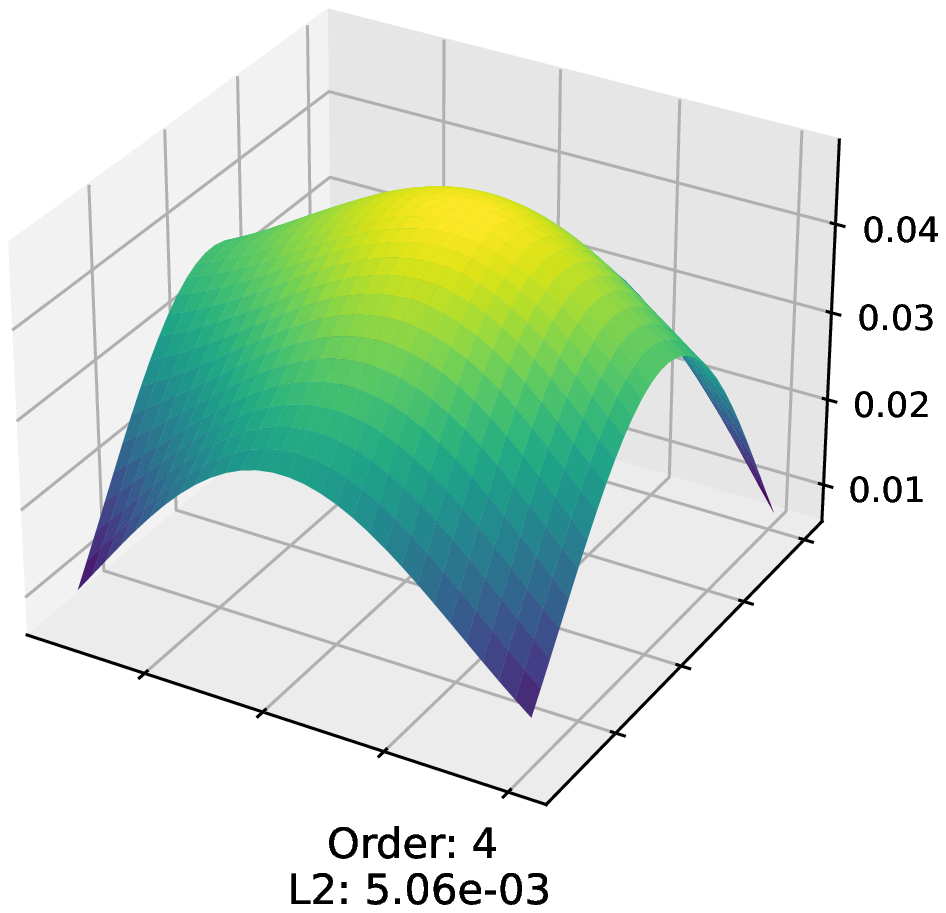}

    \includegraphics[width=0.24\linewidth]{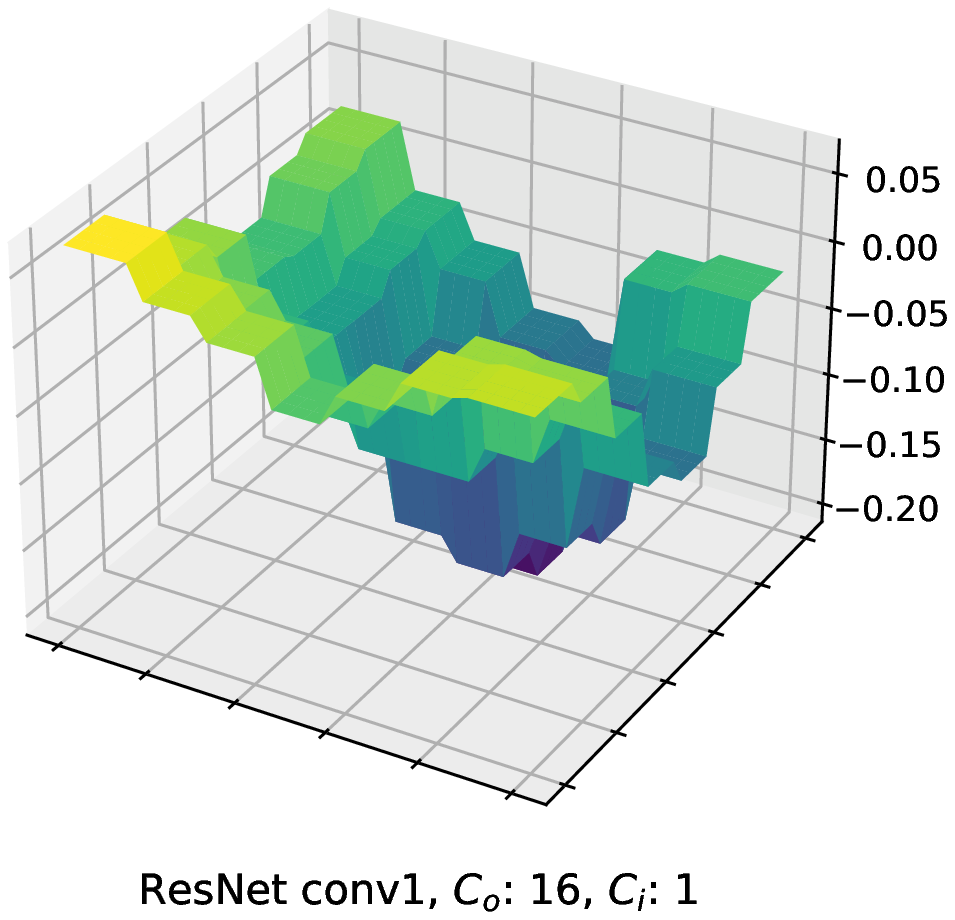}
    \includegraphics[width=0.24\linewidth]{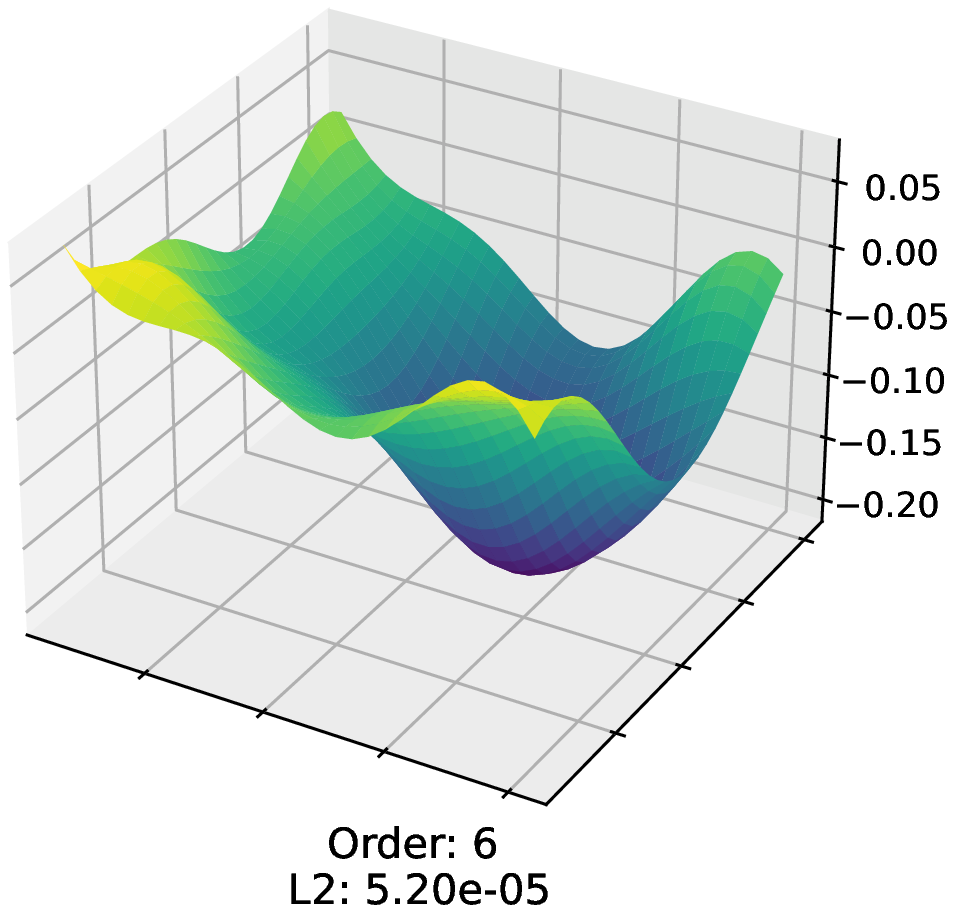}
    \includegraphics[width=0.24\linewidth]{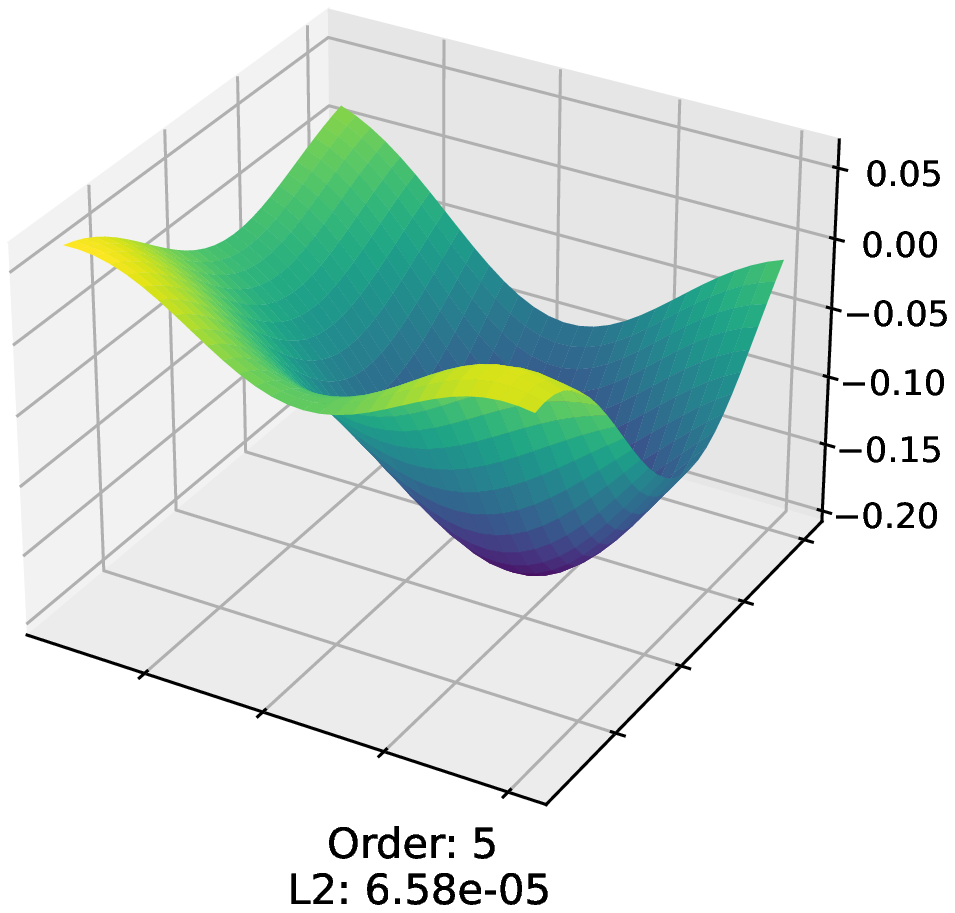}
    \includegraphics[width=0.24\linewidth]{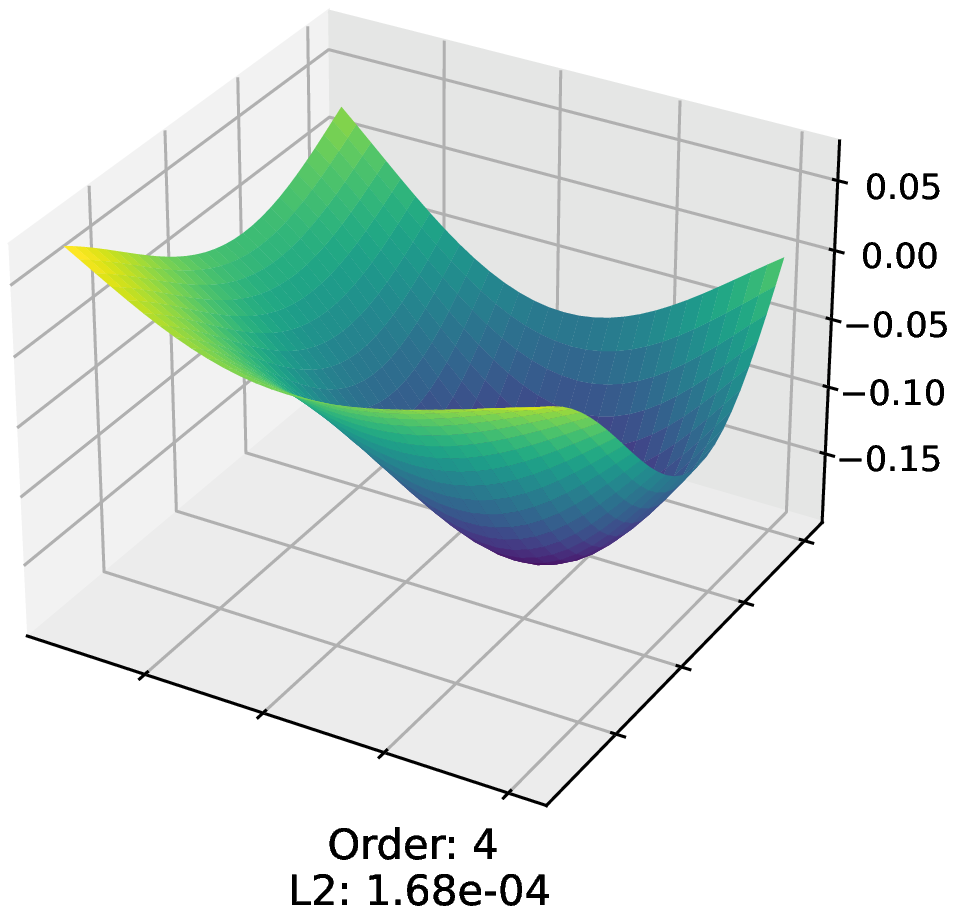}

    \includegraphics[width=0.24\linewidth]{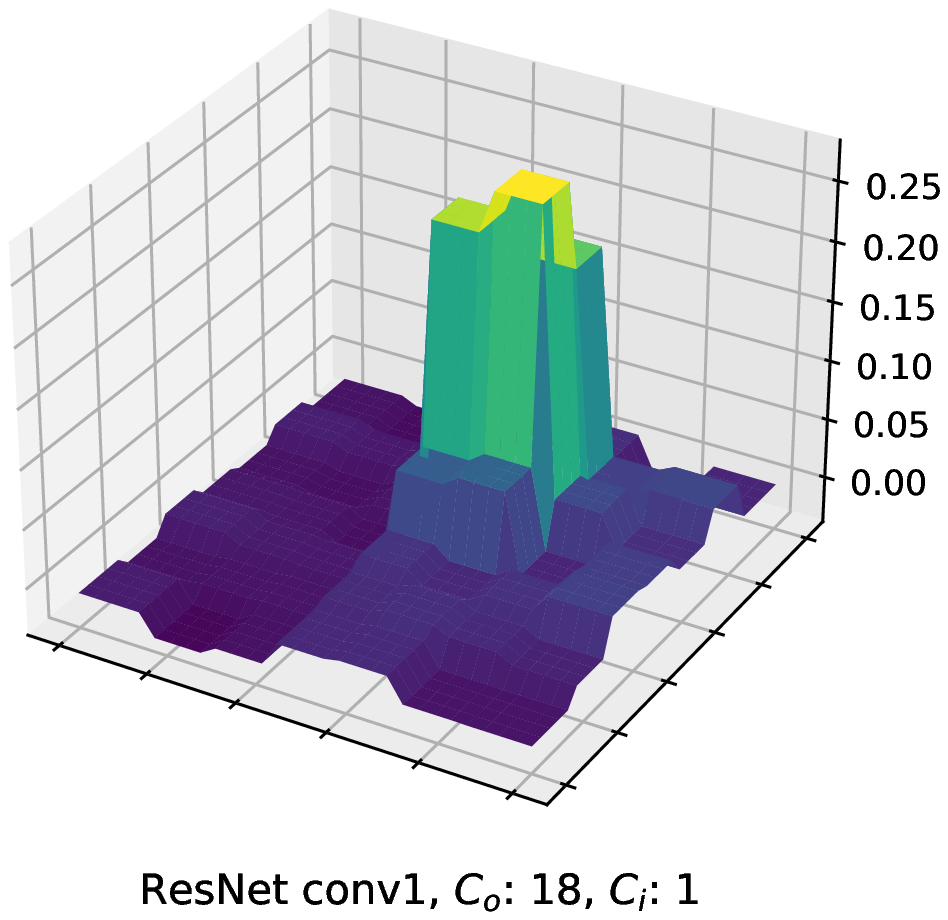}
    \includegraphics[width=0.24\linewidth]{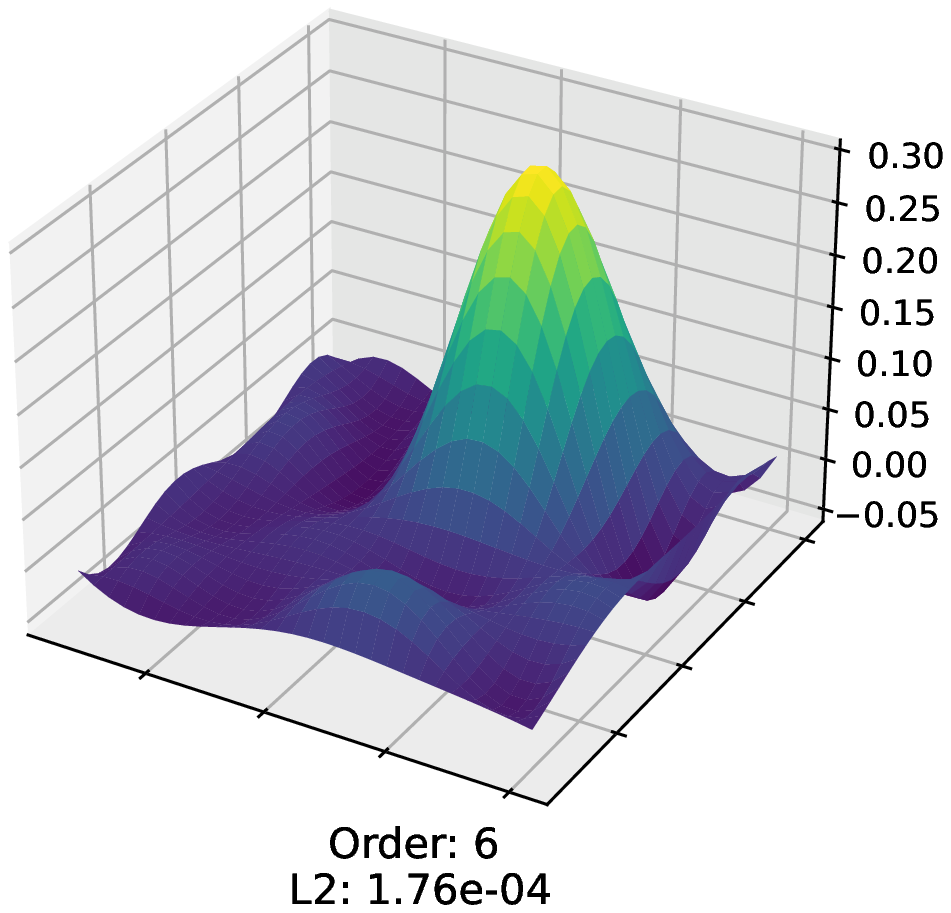}
    \includegraphics[width=0.24\linewidth]{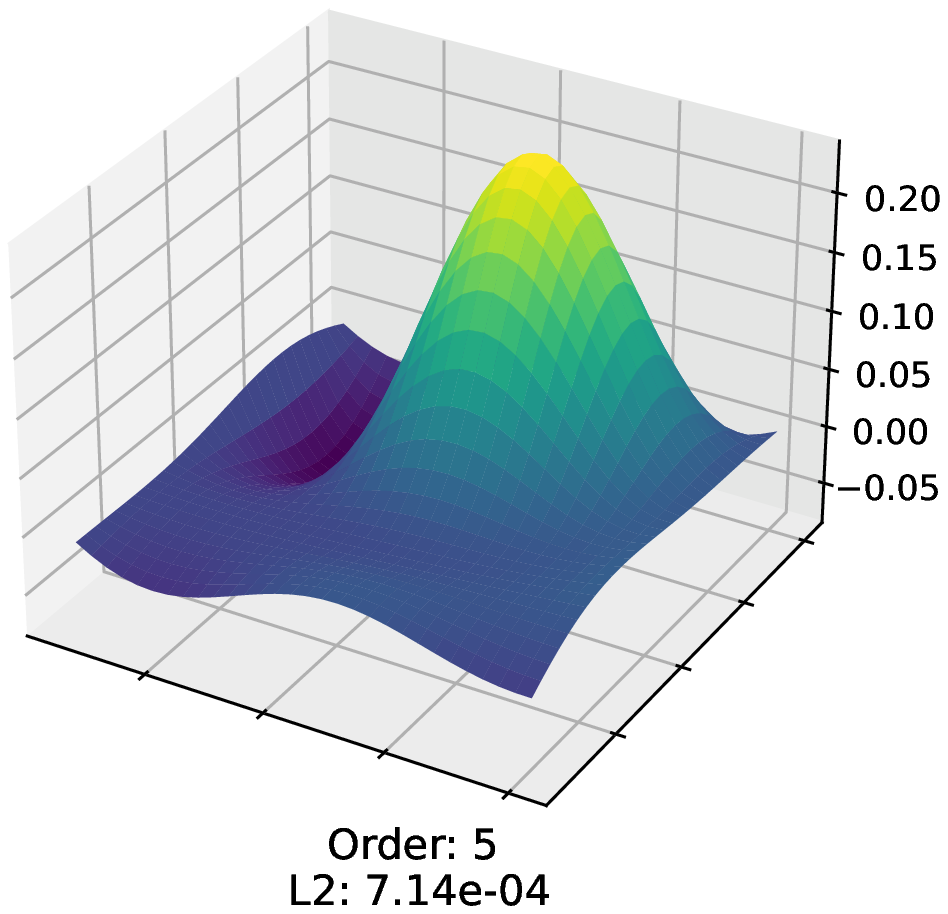}
    \includegraphics[width=0.24\linewidth]{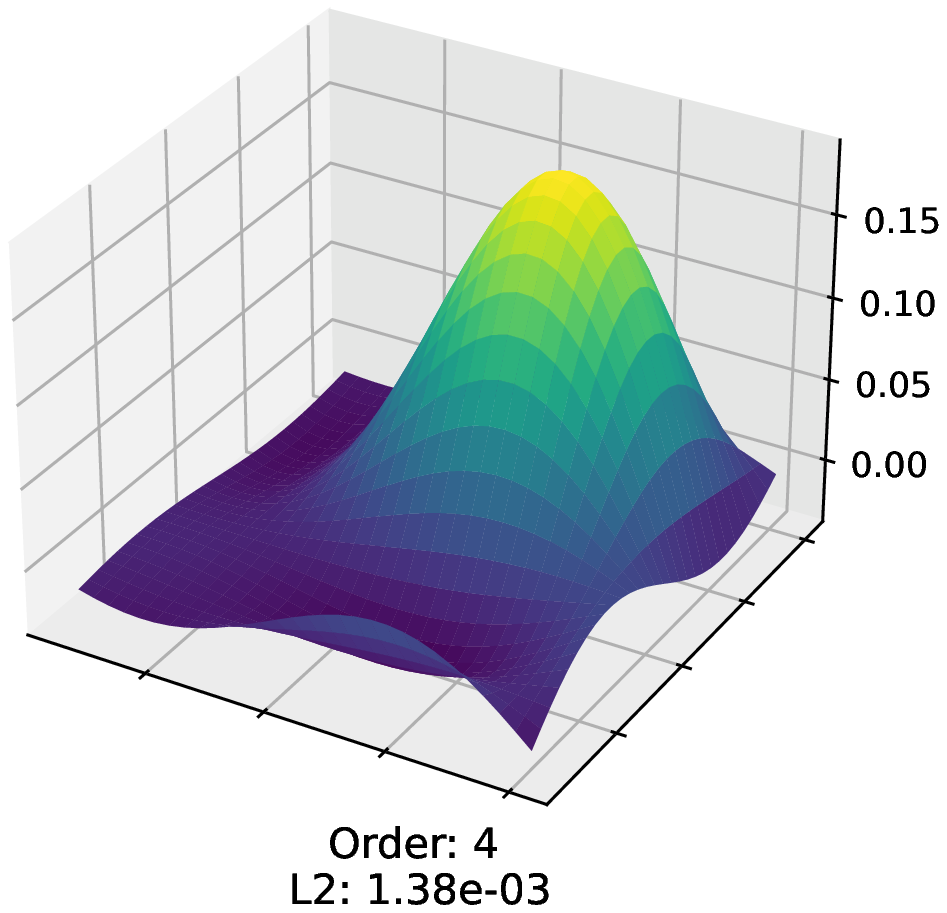}

    \includegraphics[width=0.24\linewidth]{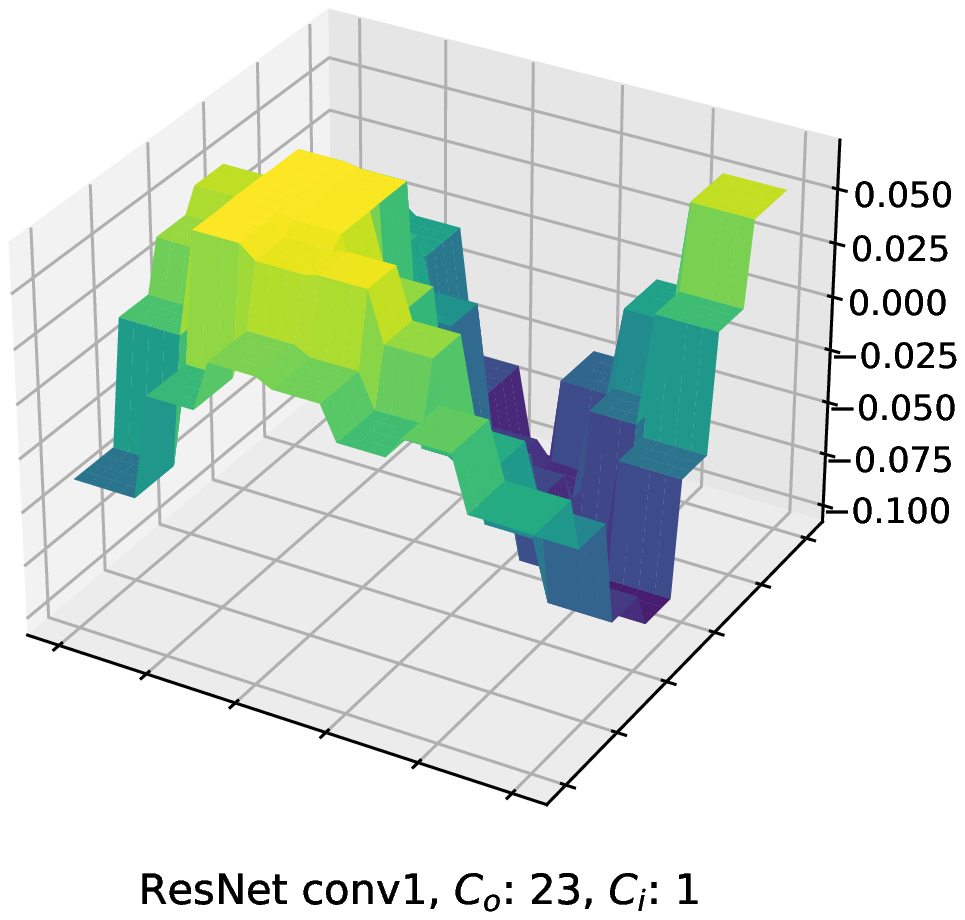}
    \includegraphics[width=0.24\linewidth]{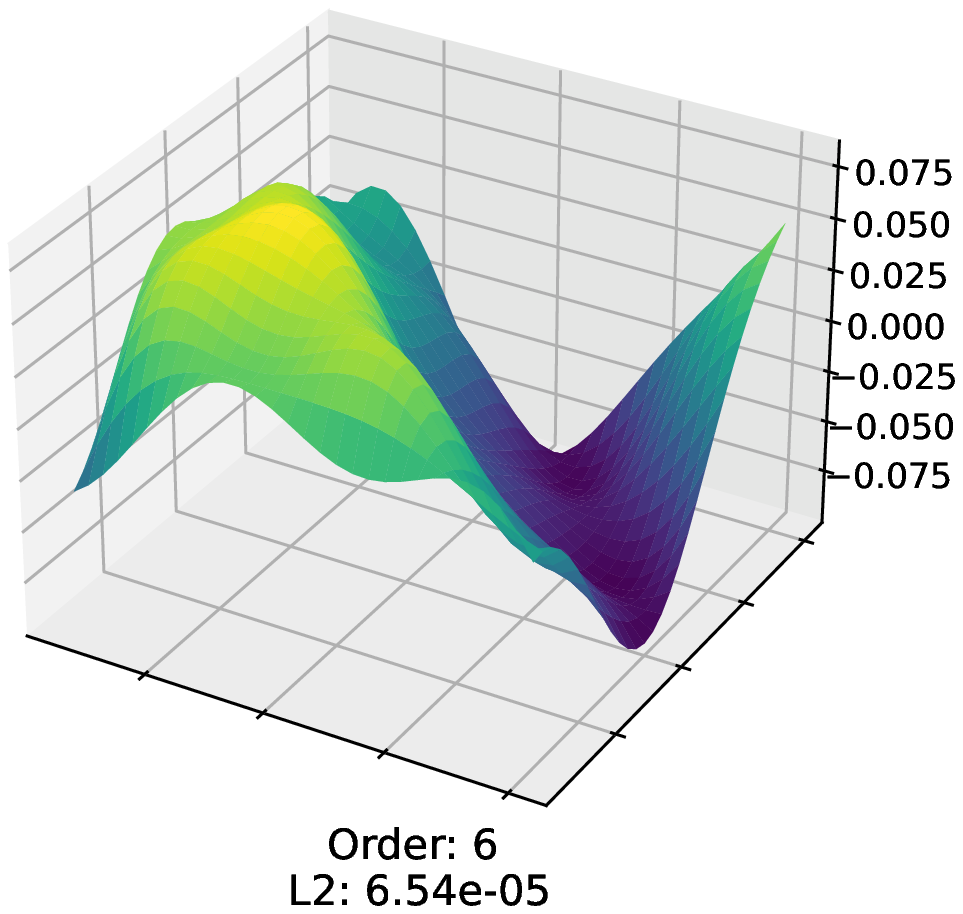}
    \includegraphics[width=0.24\linewidth]{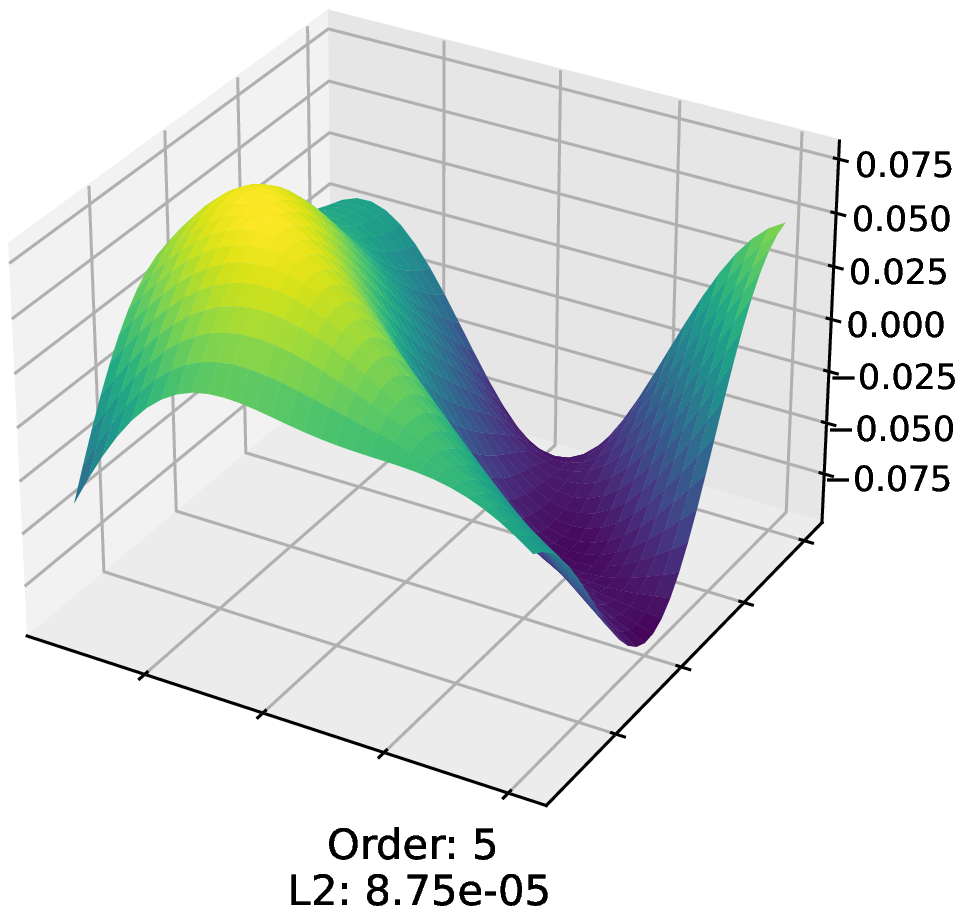}
    \includegraphics[width=0.24\linewidth]{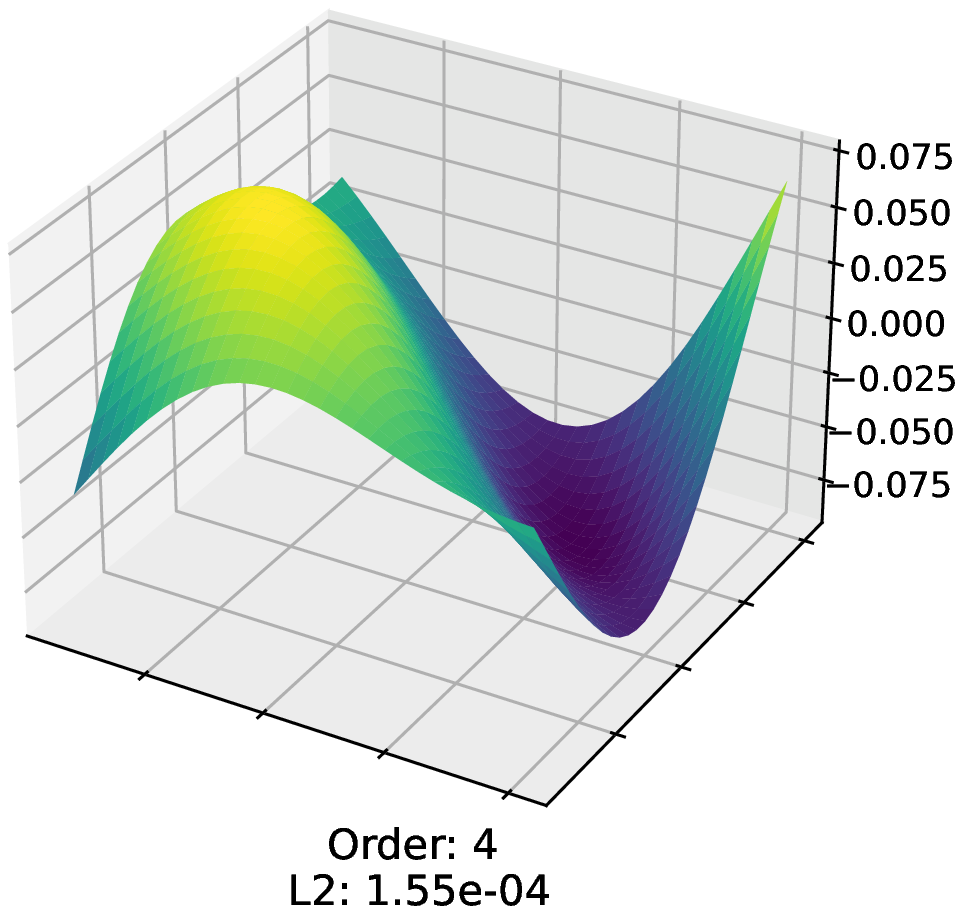}

    \caption{Filters from the ConvNeXt and ResNet networks compressed using our method.
        Left: Original Kernel, Right (in columns): Compressed version of this kernel using CosConv with 6, 5, and 4 orders/harmonics respectively.}
    \label{fig:extracosvis}
\end{figure}

\begin{figure}[tbp]
    \centering
    \includegraphics[width=0.24\linewidth]{images/visfig/orig_stages.1.1.dwconv.weight_100_0.eps}
    \includegraphics[width=0.24\linewidth]{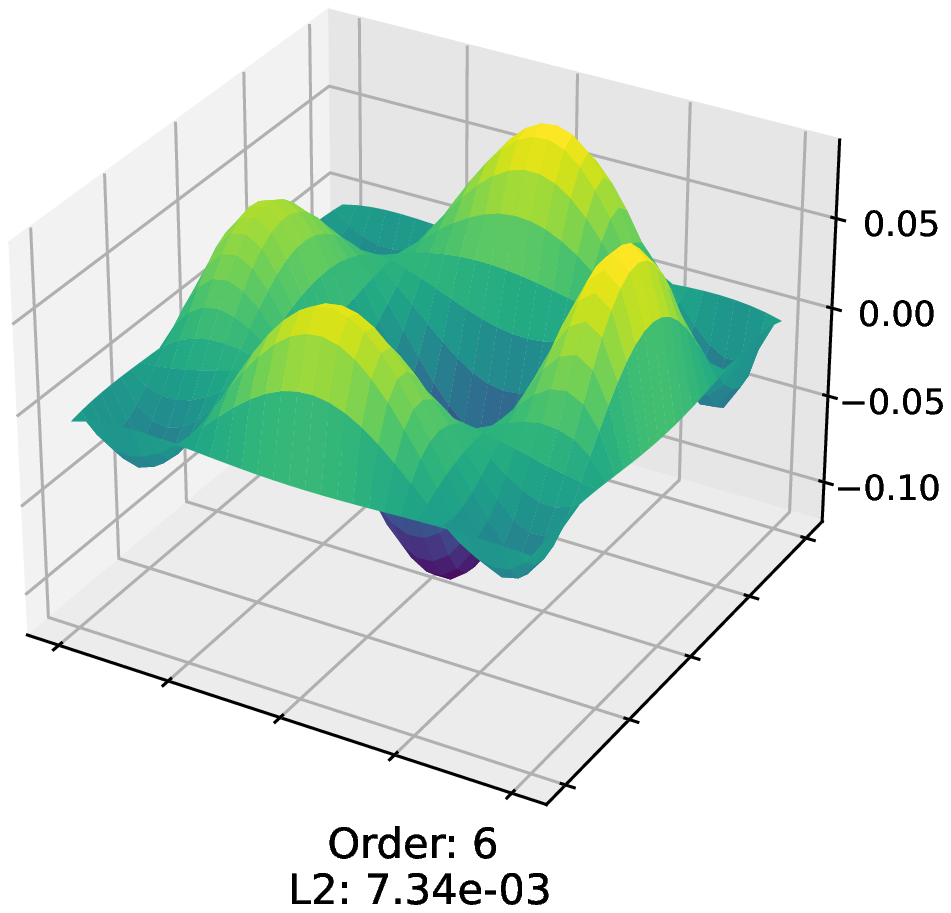}
    \includegraphics[width=0.24\linewidth]{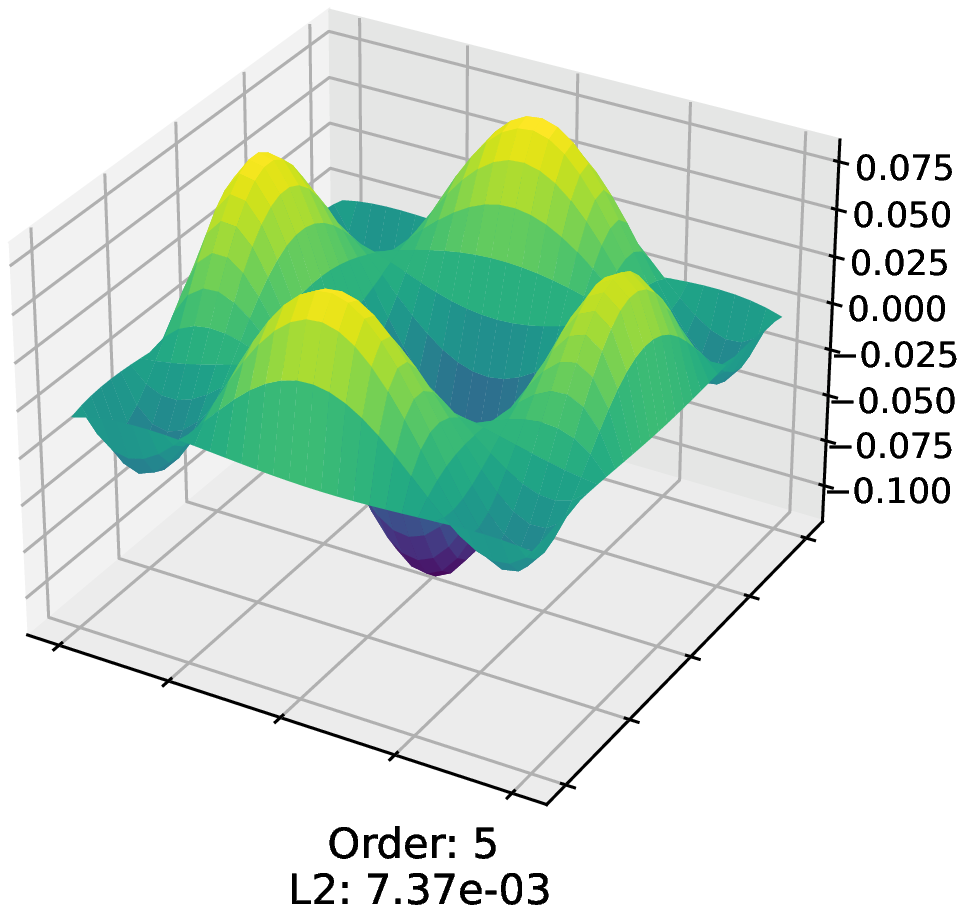}
    \includegraphics[width=0.24\linewidth]{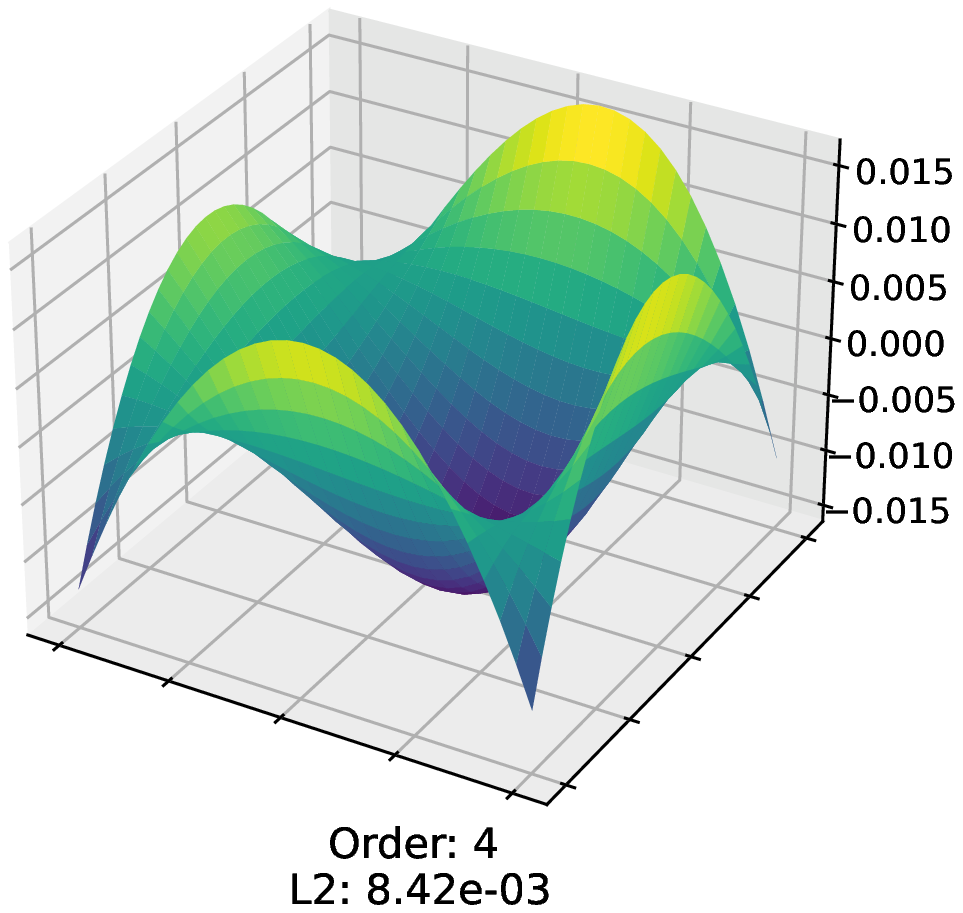}

    \includegraphics[width=0.24\linewidth]{images/visfig/orig_stages.3.2.dwconv.weight_5_0.eps}
    \includegraphics[width=0.24\linewidth]{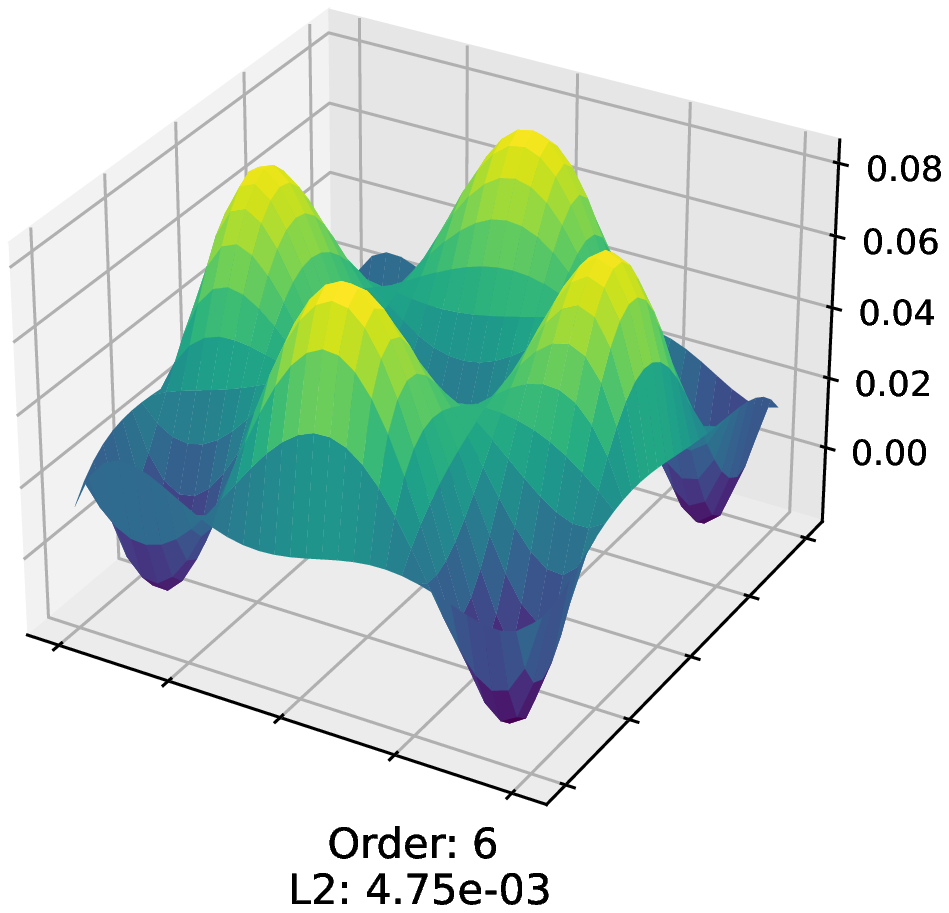}
    \includegraphics[width=0.24\linewidth]{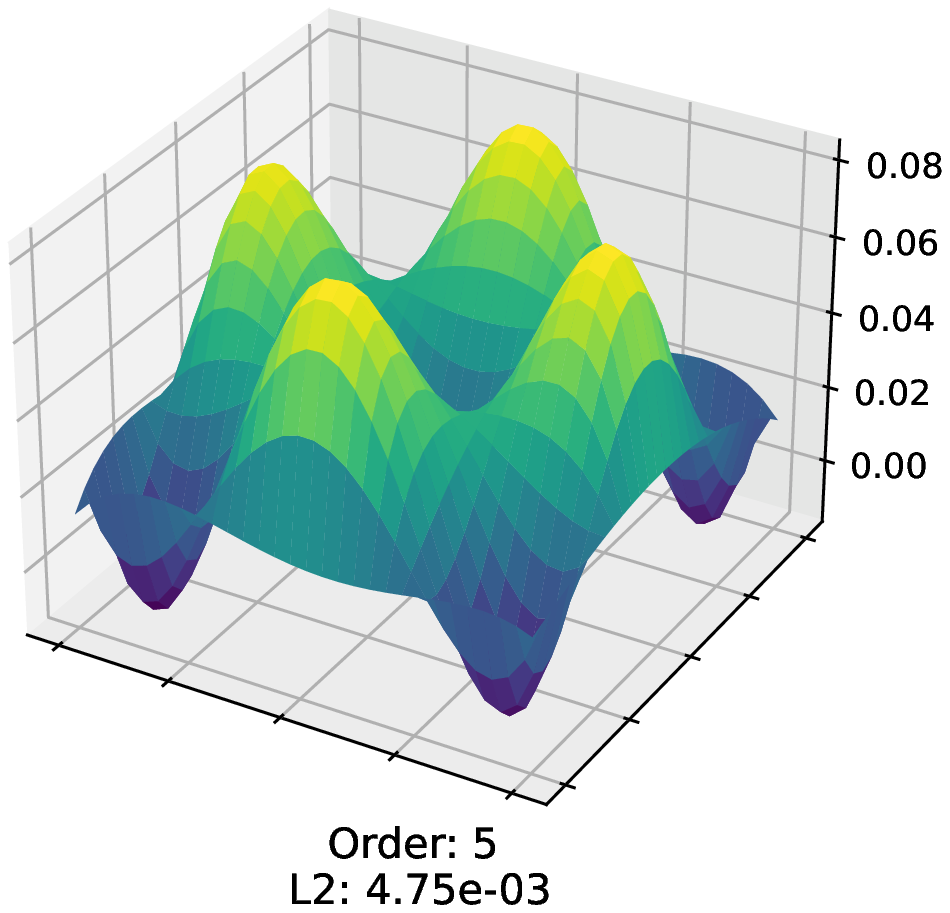}
    \includegraphics[width=0.24\linewidth]{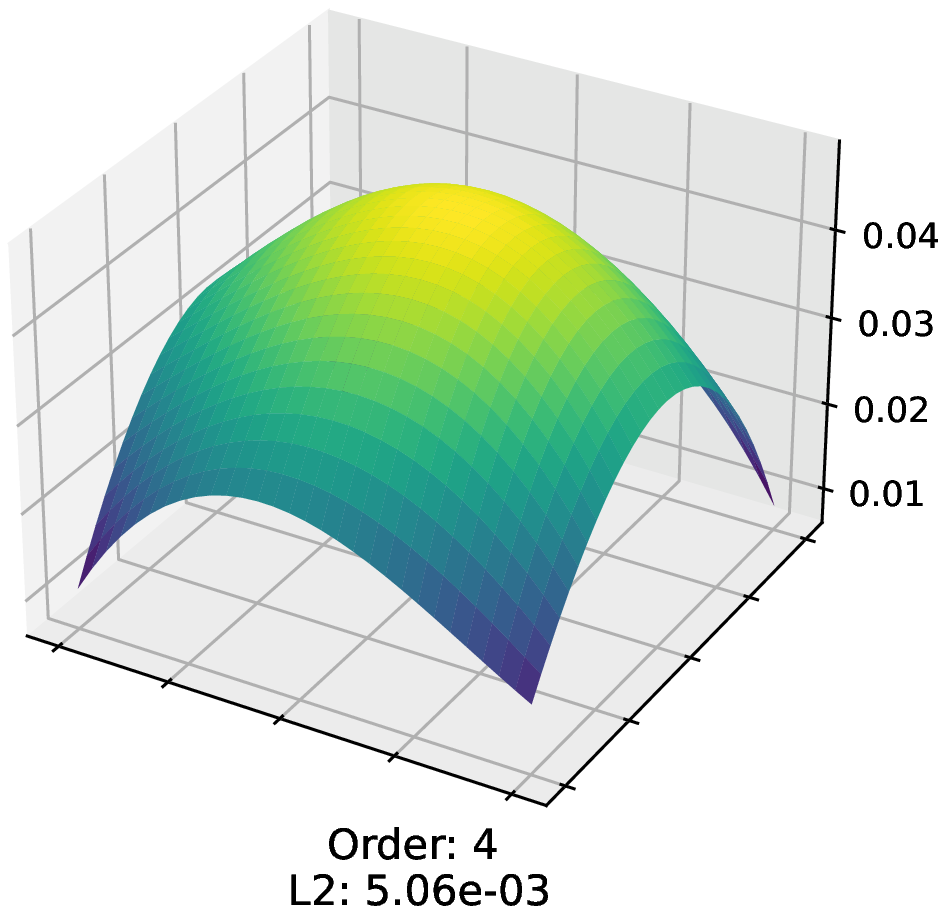}

    \includegraphics[width=0.24\linewidth]{images/visfig/orig_conv1.weight_16_1.eps}
    \includegraphics[width=0.24\linewidth]{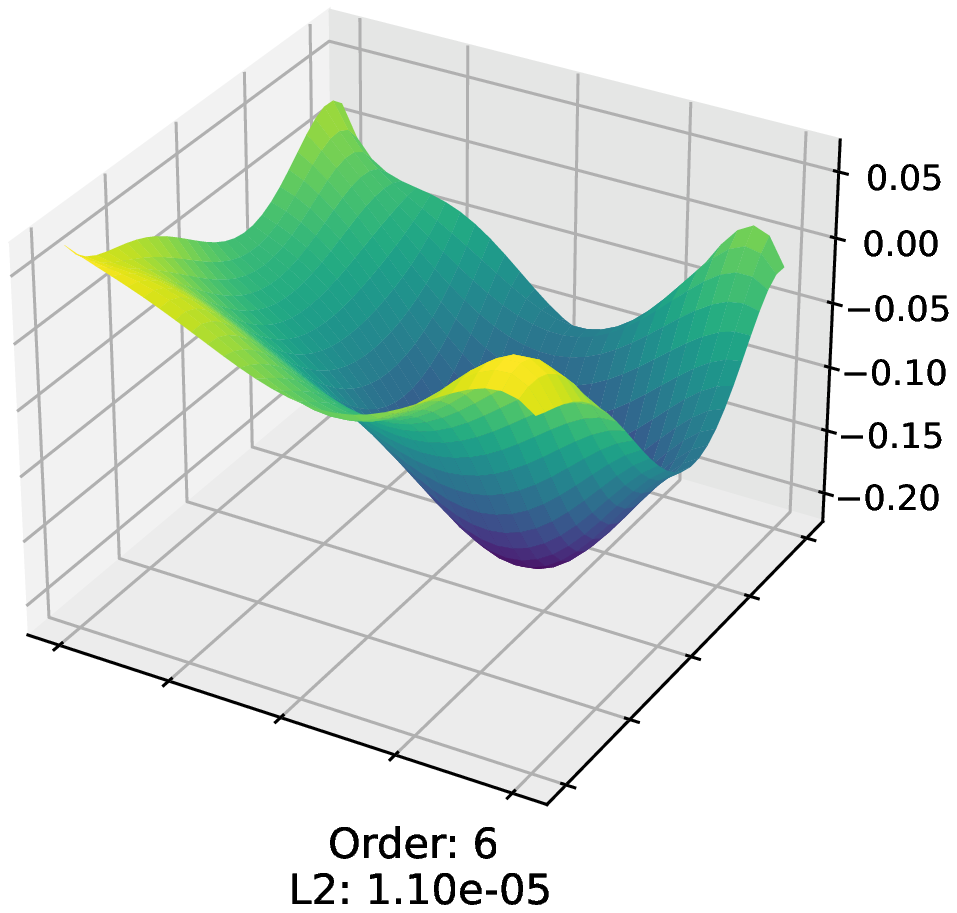}
    \includegraphics[width=0.24\linewidth]{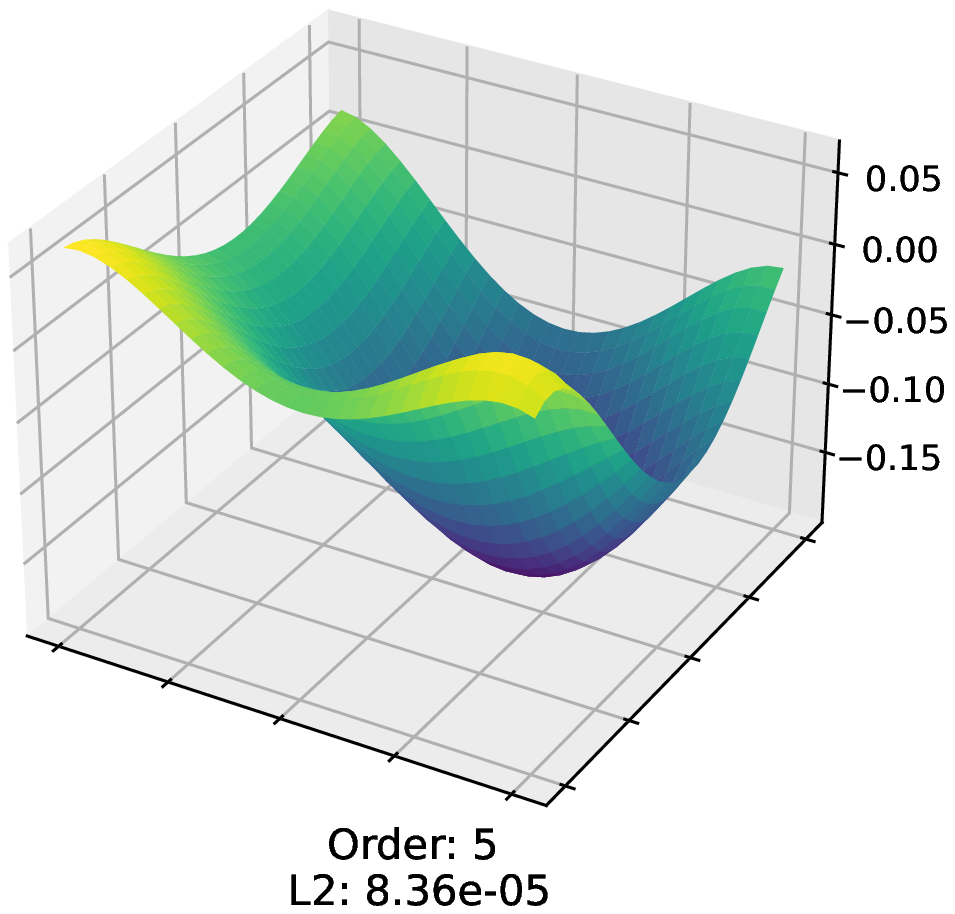}
    \includegraphics[width=0.24\linewidth]{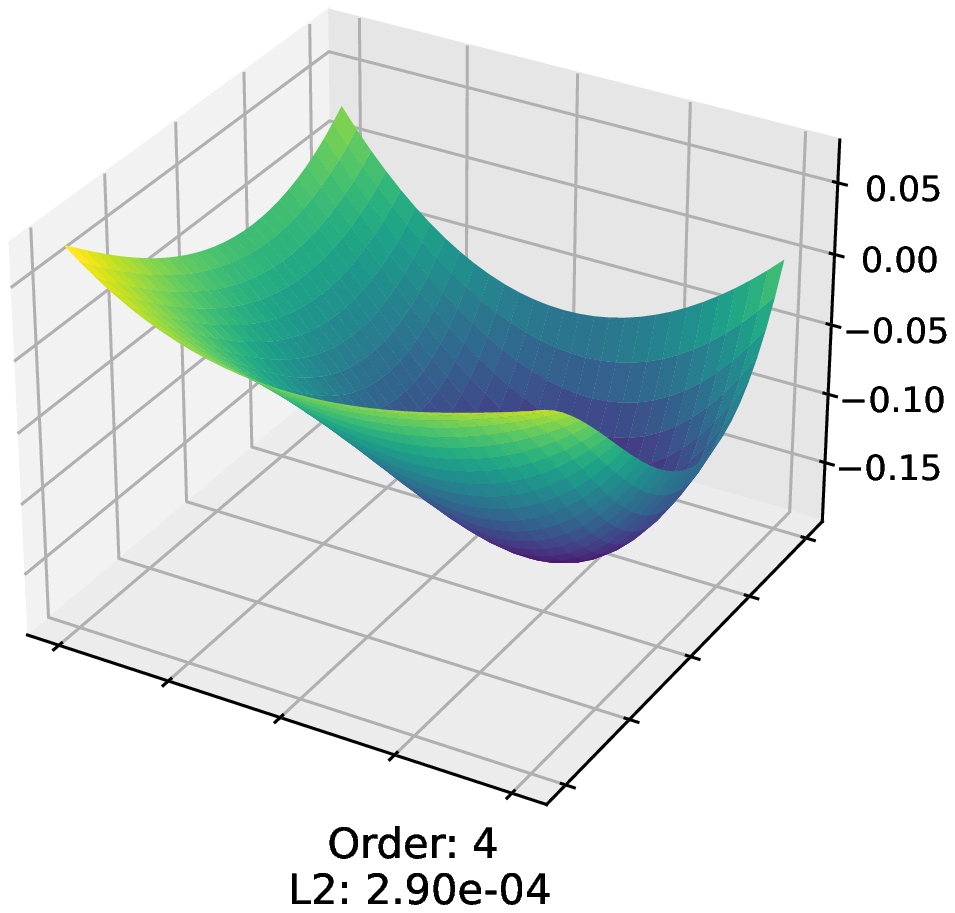}

    \includegraphics[width=0.24\linewidth]{images/visfig/orig_conv1.weight_18_1.eps}
    \includegraphics[width=0.24\linewidth]{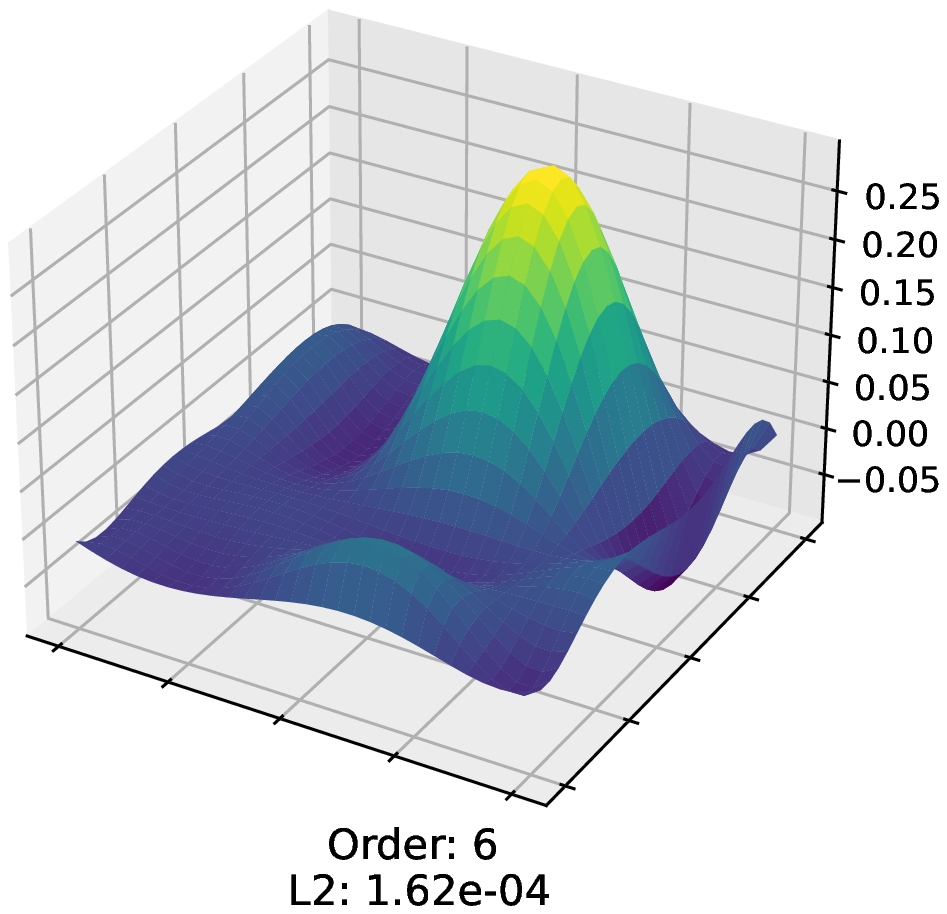}
    \includegraphics[width=0.24\linewidth]{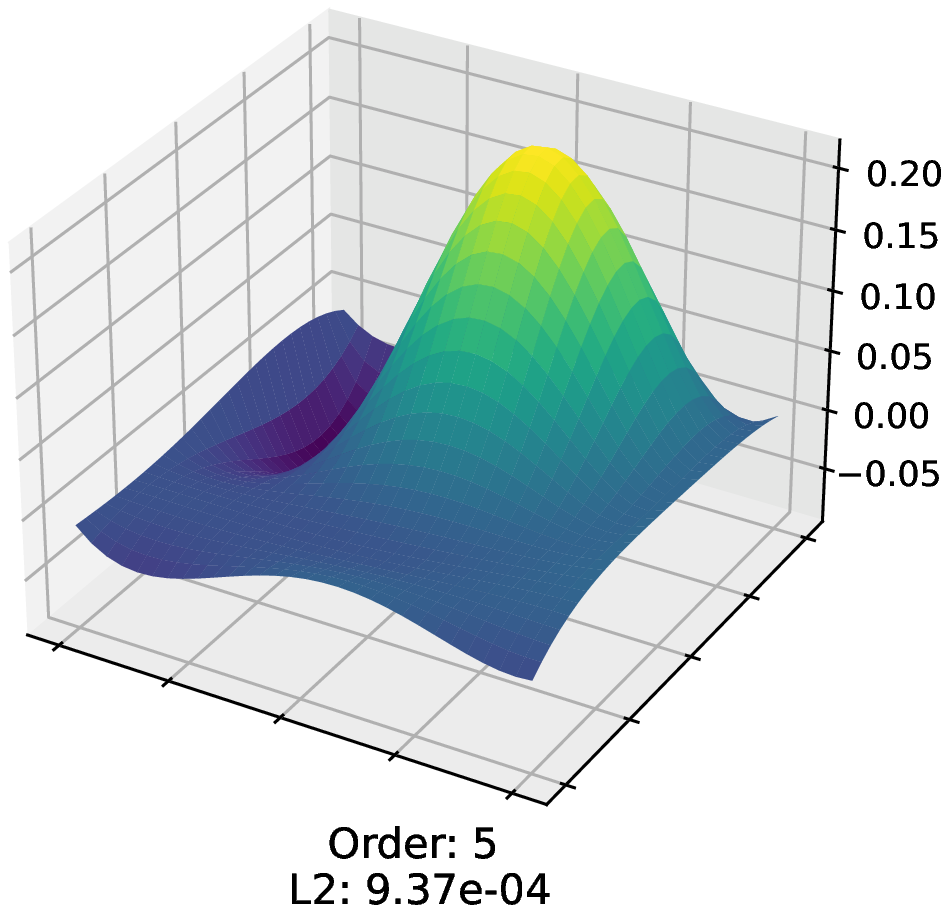}
    \includegraphics[width=0.24\linewidth]{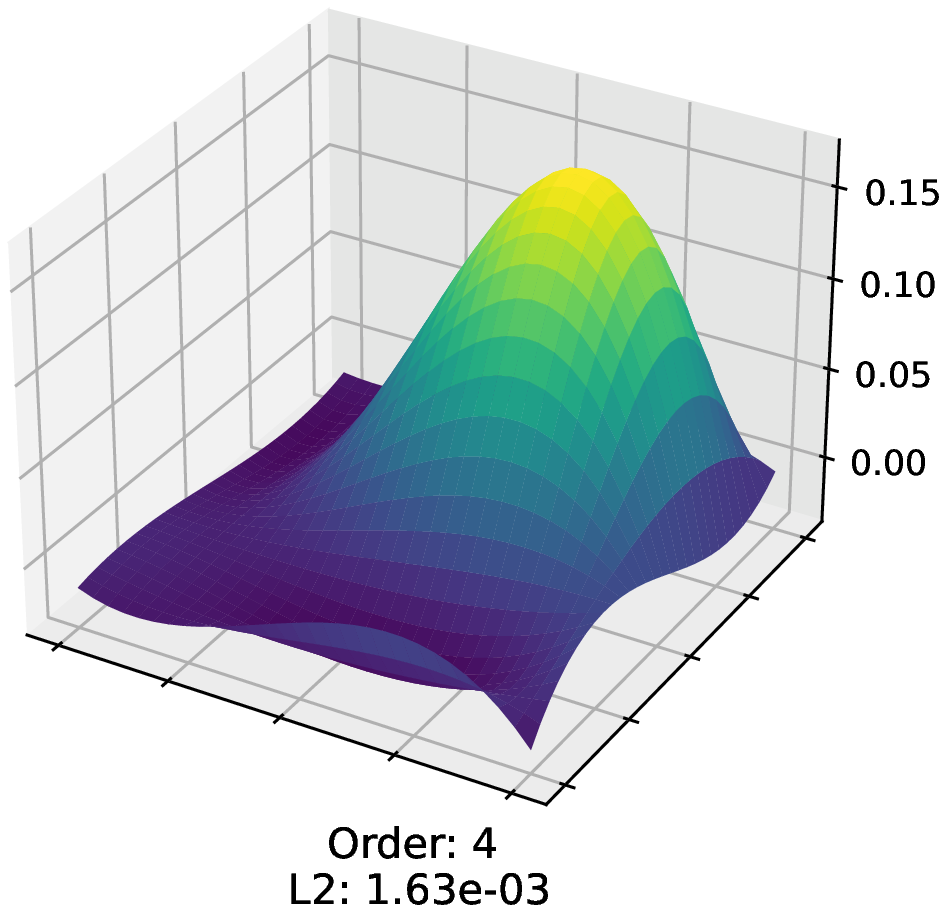}

    \includegraphics[width=0.24\linewidth]{images/visfig/orig_conv1.weight_23_1.eps}
    \includegraphics[width=0.24\linewidth]{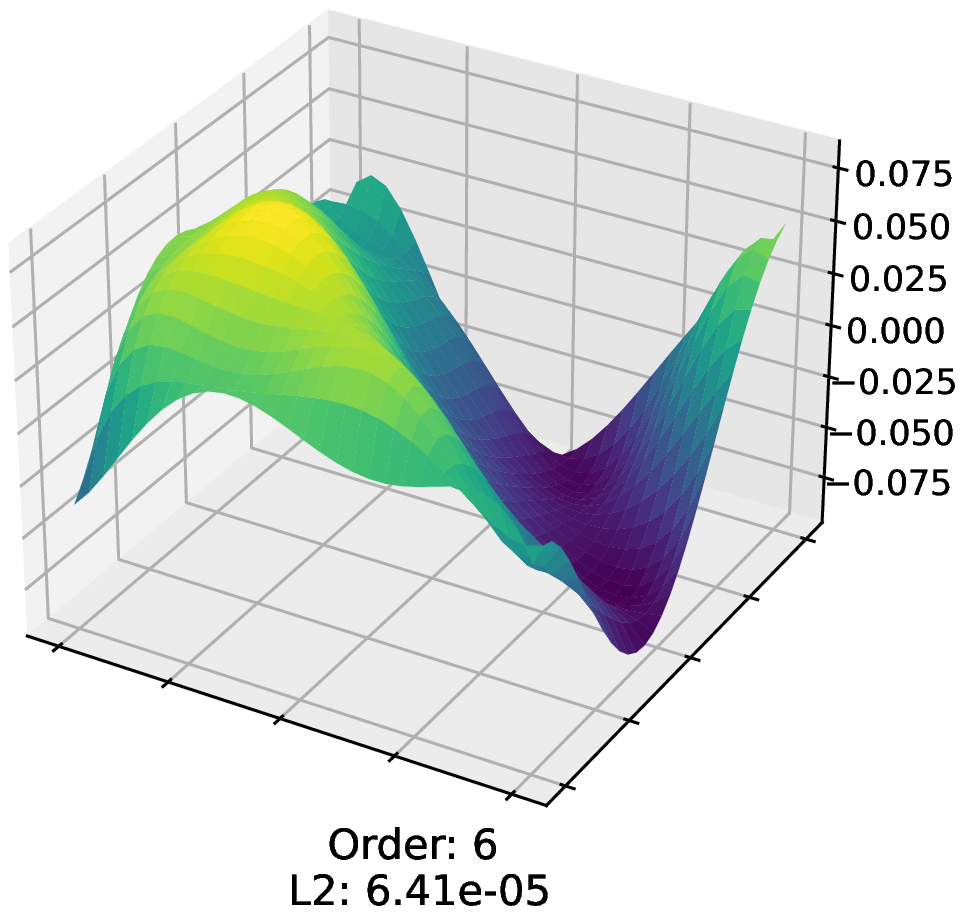}
    \includegraphics[width=0.24\linewidth]{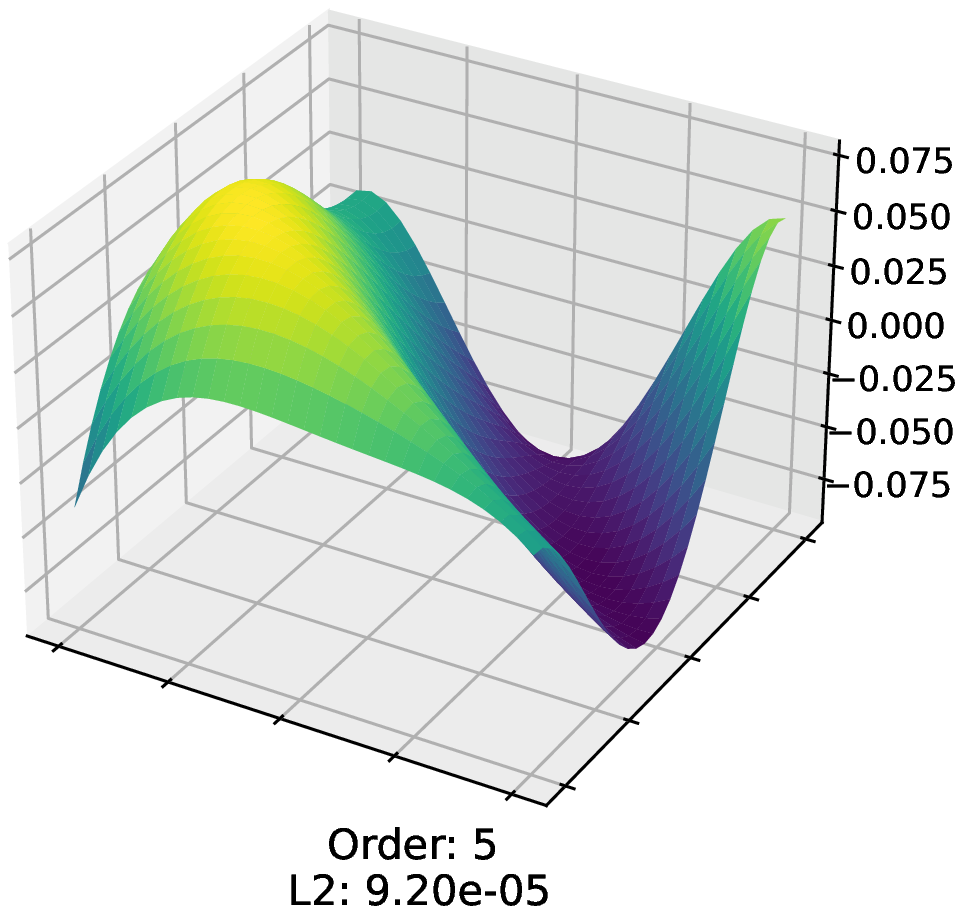}
    \includegraphics[width=0.24\linewidth]{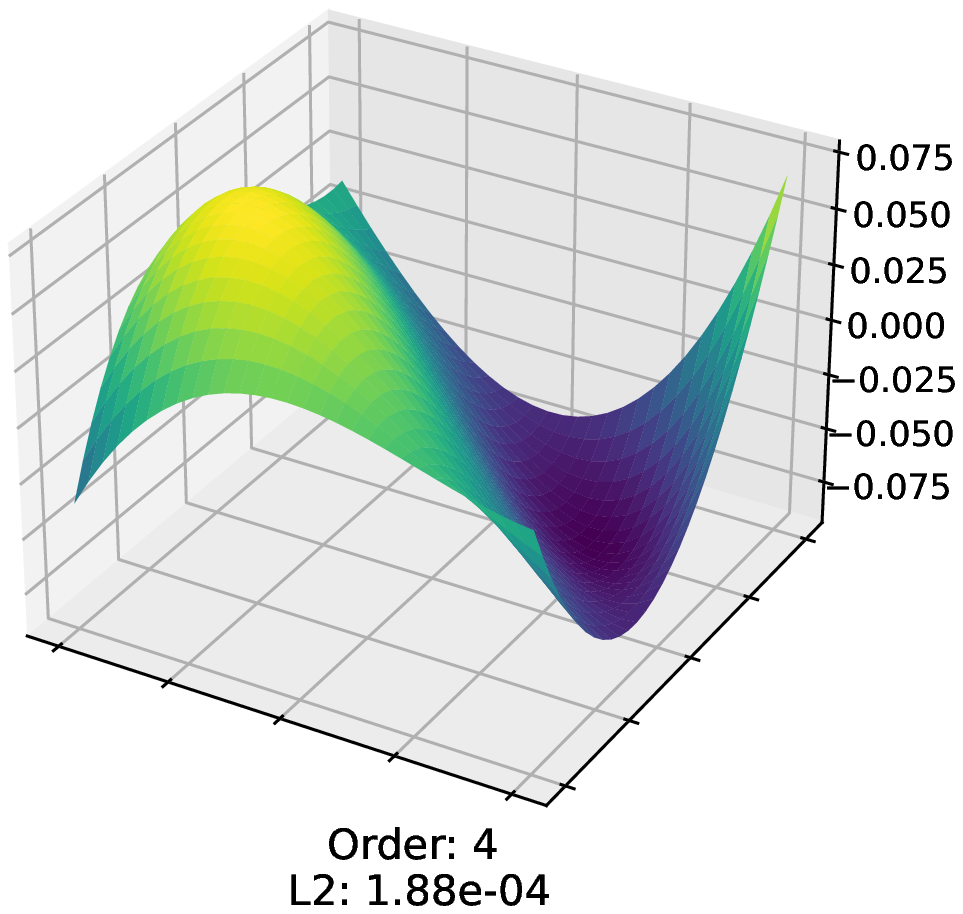}

    \caption{Filters from the ConvNeXt and ResNet networks compressed using our method.
        Left: Original Kernel, Right (in columns): Compressed version of this kernel using ChebConv with 6, 5, and 4 orders/harmonics respectively.}
    \label{fig:extrachebvis}
\end{figure}

\Cref{fig:chebvis}, \cref{fig:extracosvis}, and \cref{fig:extrachebvis} show more visualisations of kernel functions compressed using our method.
\Cref{fig:chebvis} shows the same kernels as Figure 2. in the main paper, but using ChebConv rather than CosConv.
The remaining two figures show visualisations of further kernels using CosConv (\cref{fig:extracosvis}) and ChebConv (\cref{fig:extrachebvis})

\end{document}